\documentclass[10pt,journal,compsoc]{IEEEtran}
\ifCLASSOPTIONcompsoc
\usepackage[nocompress]{cite}
\else
\usepackage{cite}
\fi

\usepackage[utf8]{inputenc}
\usepackage{longtable}
\usepackage{graphicx}
\usepackage{color}

\usepackage[cmex10]{amsmath}

\usepackage{array}

\usepackage{mdwmath}
\usepackage{mdwtab}
\usepackage{subfig}
\usepackage{url}
\usepackage{bm}

\usepackage{soul}
\usepackage{multirow}
\usepackage{booktabs}
\usepackage{amssymb}
\usepackage{lscape}

\usepackage{rotating}

\usepackage{tikz}
\usepackage{pifont}

\usepackage{nicefrac}

\usepackage{algorithm}
\usepackage{algpseudocode}

\usepackage{pstricks}

\usepackage{wrapfig}

\usepackage[colorlinks, linkcolor=blue, urlcolor=blue]{hyperref}

\usepackage{xr-hyper}
\makeatletter
\newcommand*{\addFileDependency}[1]{
	\typeout{(#1)}
	\@addtofilelist{#1}
	\IfFileExists{#1}{}{\typeout{No file #1.}}
}\makeatother

\newrgbcolor{guancolor}{0.0 0.0 0.0}
\newrgbcolor{hamidcolor}{0.0 0.0 0.0}
\newcommand{\guan}[1]{{\guancolor{#1}}}
\newcommand{\hamid}[1]{{\hamidcolor{#1}}}

\newrgbcolor{guanwangCol}{0.0, 0.0, 0.0}
\newcommand{\guanwang}[1]{{\guanwangCol{#1}}} 

\newrgbcolor{guancolorRRR}{0.0, 0.0, 0.0}
\newcommand{\guanwangRRR}[1]{{\guancolorRRR{#1}}} 

\newcommand{\real}{ \mathbb{R}}
\newcommand{\rplus}{ \mathbb{R}^+}

\newcommand{\radiusfunc}{r}

\newcommand{\rthree}{ \real^3}

\newcommand{\argmin}{\arg\min}
\newcommand{\ie}{\emph{i.e., }}
\newcommand{\eg}{\emph{e.g., }}
\newcommand{\etal}{\emph{et al.}}
\newcommand{\noi}{\noindent}

\newcommand{\esrvfspace}{\Omega}			

\newcommand{\ltwo}{\mathbb{L}^{2}}

\newcommand{\inner}[2]{\langle #1,#2 \rangle}
\newcommand{\innerd}[2]{\langle \langle #1,#2 \rangle \rangle}

\newcommand{\srvf}{q}

\newcommand{\branch}{\beta}		
\newcommand{\parameter}{s}
\newcommand{\curvefunc}{f}
\newcommand{\thickness}{r}

\newcommand{\proot}{\bm{\beta}}	

\newcommand{\preshapecurves}{\mathcal{A}_{\curvefunc}}
\newcommand{\deformationpath}{\alpha}
\newcommand{\branchone}{\branch_1}
\newcommand{\branchtwo}{\branch_2}
\newcommand{\timeparam}{t}
\newcommand{\srvfmap}{\mathcal{Q}}
\newcommand{\diffeo}{\gamma}
\newcommand{\diffeos}{\Gamma}
\newcommand{\rotation}{O}
\newcommand{\rotations}{SO(3)}
\newcommand{\preshapesrvfs}{\mathcal{A}_q}
\newcommand{\preshapesrvts}{\mathcal{C}_{\srvft}}

\newcommand{\preshape}{\mathcal{C}}
\newcommand{\diffeotree}{\boldsymbol{\gamma}}
\newcommand{\permutetree}{\boldsymbol{\sigma}}
\newcommand{\permute}{\sigma}
\newcommand{\diffeotrees}{\boldsymbol{\Gamma}}
\newcommand{\permutetreespace}{\textbf{S}}
\newcommand{\treeorder}{n}
\newcommand{\treedeformationpath}{\boldsymbol{\alpha}}
\newcommand{\meantree}{\boldsymbol{\mu}}
\newcommand{\meansrvft}{\boldsymbol{\mu}_{\srvf}}

\newcommand{\srvft}{\bm{\srvf}}
\newcommand{\preshapetrees}{\mathcal{C}_{\proot}}

\newcommand{\covMatrix}{K}          
\newcommand{\diagCovMatrix}{\mathcal{\covMatrix}}
\def\argmin{\mathop{\rm argmin}}

\graphicspath{{Figures/}{Figures/GeodesicsBotanicalTrees/}
{Figures/GeodesicsNeuroTrees/}{Figures/SymmetryAnalysisBotanicalTrees/}{Figures/SymmetryAnalysisNeuroTrees/}{Figures/GeodesicsAblationStudy/}{Figures/MeanNeuroTrees/}{Figures/MeanBotanTrees/}{Figures/ModesRandSamplesNeuroTrees/}{Figures/ModesRandSamplesBotanTrees/}{Figures/ComparisonGeodBotanicalTrees/}{Figures/GeodesicsWeightStudy/}
{Figures/QEDdrawbacks/}{Figures/GeodesicCorrMaps/}{Figures/Geodesics10timesRun/}{Figures/Geodesics10timesRun_v2/}}

\DeclareGraphicsExtensions{.pdf,.eps}

\begin{document}
\bstctlcite{IEEEexample:BSTcontrol}

\title{Elastic Shape Analysis of Tree-like 3D Objects using Extended SRVF Representation}
\author{Guan Wang, Hamid Laga, Anuj Srivastava~\IEEEmembership{Fellow,~IEEE}
	
	\IEEEcompsocitemizethanks{\IEEEcompsocthanksitem Guan Wang is with Yangtze Delta Region Institute (Huzhou), University of Electronic Science and Technology of China, Huzhou, 313000, China. Email: wangguan12621@gmail.com.
		\IEEEcompsocthanksitem  Hamid Laga is with the School of Information Technology (Murdoch University), the Centre of Biosecurity and One Health (Harry Butler Institute, Murdoch University), and the Centre for Healthy Ageing (Health Futures Institute, Murdoch University). Email: H.Laga@murdoch.edu.au
		\IEEEcompsocthanksitem Anuj Srivastava is with the Department of Statistics, Florida State University. Email: anuj@stat.fsu.edu
	}
	\thanks{Manuscript received Oct. 15, 2021; revised Sep., 2023.}
}

\markboth{Elastic Shape Analysis of Tree-like 3D Objects using Extended SRVF Representation}%
{Guan \MakeLowercase{\etalnospace}: Elastic Shape Analysis of Tree-like 3D Objects using Extended SRVF Representation}

\IEEEtitleabstractindextext{
	\begin{abstract}
		How can one analyze detailed 3D biological objects, such as neuronal and botanical trees, that exhibit complex geometrical and topological variation? In this paper, we develop a novel mathematical framework for representing, comparing, and computing geodesic deformations between the shapes of such tree-like 3D objects. A hierarchical organization of subtrees characterizes these objects -- each subtree has a main branch with some side branches attached -- and one needs to match these structures across objects for meaningful comparisons. We propose a novel representation that extends the Square-Root Velocity Function (SRVF), initially developed for Euclidean curves, to tree-shaped 3D objects. We then define a new metric that quantifies the bending, stretching, and branch sliding needed to deform one tree-shaped object into the other. Compared to the current metrics such as the Quotient Euclidean Distance (QED) and the Tree Edit Distance (TED), the proposed representation and metric capture the full elasticity of the branches (\ie bending and stretching) as well as the topological variations (\ie branch death/birth and sliding). It completely avoids the shrinkage that results from the edge collapse and node split operations of the QED and TED metrics. We demonstrate the utility of this framework in comparing, matching, and computing geodesics between biological objects such as neuronal and botanical trees. We also demonstrate its application to various shape analysis tasks such as \textbf{(i)} symmetry analysis and symmetrization of tree-shaped 3D objects, \textbf{(ii)} computing summary statistics (means and modes of variations) of populations of tree-shaped 3D objects, \textbf{(iii)} fitting parametric probability distributions to such populations, and \textbf{(iv)} finally synthesizing novel tree-shaped 3D objects through random sampling from estimated probability distributions.
	\end{abstract}
	
}

\maketitle


\IEEEdisplaynotcompsoctitleabstractindextext

%
\IEEEpeerreviewmaketitle

\section{Introduction}
\label{sec:introduction}
\IEEEPARstart{T}{ree-like} 3D objects are ubiquitous, especially in delivery systems such as vascular systems, airway trees, blood vessels in the eye, plant roots and shoots, and neuronal structures in the brain. Analyzing and understanding the 3D structural variability of such tree-like biological objects are of great interest from multiple scientific perspectives. Shapes of objects both constrain and enable their functionalities in larger biological systems. Thus, a mathematical characterization of shapes can help provide insights into objects' functional roles in such biological phenomena as genesis, growth, and disease. For instance, studies of changes in the 3D structure of plant roots could improve their water and nutrient uptake efficiency. Several papers have also shown that neuronal morphology, \ie their types, geometry, and topology, and their connections are the key to discerning how neurons integrate information and subsequently explaining brain activity and its functions~\cite{jiang2015principles,spruston2008pyramidal,lin2018modelling}. Also, statistical analysis of neuron morphology is crucial to understanding brain functionality and characterizing cognitive health. Alterations in neuron morphology are not only due to a normal aging~\cite{kabaso2009electrotonic} but can also be the consequence of a pathology, \eg senile dementia~\cite{chan1989alterations} and Alzheimer disease~\cite{west1994differences}.

\begin{figure}[t]
	\center
	
	\includegraphics[width=0.3\textwidth, trim={0cm 0cm  0cm 2cm},clip]{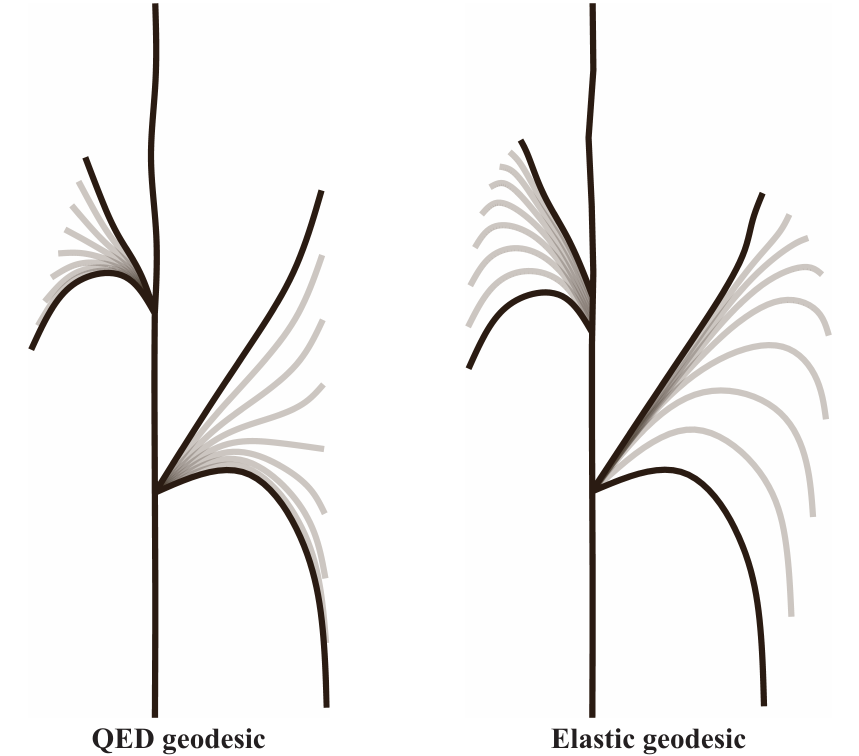}\\
	
	\small{(a) Geodesic shrinkage between trees that significantly bend. }\\
	
	\includegraphics[width=0.45\textwidth, trim={0 0cm  0 0cm},clip]{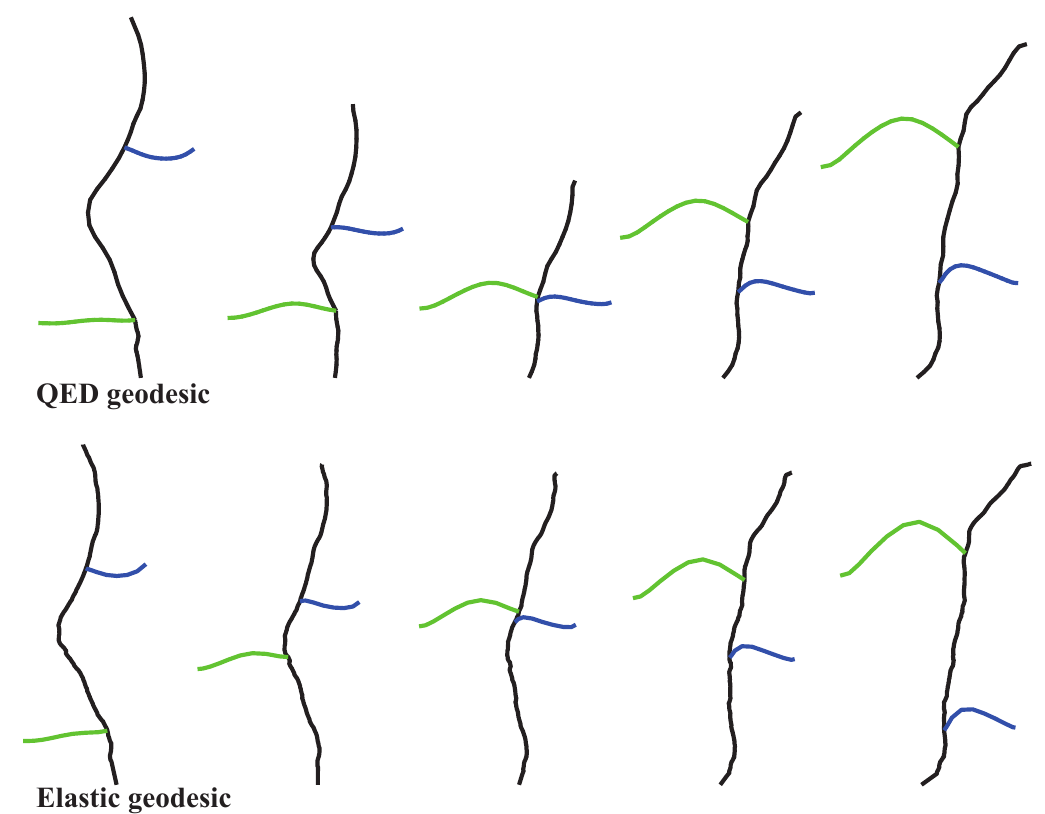}\\
	\small{(b) Geodesic shrinkage when trees exhibit different topologies.\\Branch correspondences are color-coded.}\\ 
	\caption{\guan{Examples of geodesics computed using the  QED metric and with the proposed framework. Observe that when using the QED metric, \textbf{(a)} the branches shrink along the geodesic in the presence of large bending, and (b) the trees along the geodesic shrink in the presence of topological differences.}}
	\label{fig:shrinkage_problem}
\end{figure}

Existing techniques for modeling shape variability are mostly limited to objects with fixed topology, \ie objects that only bend and stretch~\cite{blanz1999A,kurtek2013landmark,laga2014landmark,laga2017numerical,jermyn2017elastic}. Tree-shaped objects, however, are more challenging because they exhibit variability in \textbf{(1)} their geometry, in terms of the shape of the individual branches, \eg axons and dendrites in neuronal structures, which can bend and stretch, and \textbf{(2)} topology, in terms of the structural relationships between those branches. One consequence of variable geometry and topology is that finding an optimal registration, \ie putting in correspondence parts (points, curves, and branches) across such objects, becomes a challenging problem.  \hamid{Recent works that explored this problem, \eg~\cite{feragen2010geometries,feragen2011means,feragen2013toward,feragen2013tree-shapce},  are limited to simple trees that are composed of a few branches due to their computational complexity and the NP-hard nature of the correspondence problem. Wang \etal~\cite{wang2018the,wang2018statistical} extended these frameworks to handle complex botanical trees by pre-computing the correspondences. \guan{As} such, correspondences and geodesics are computed separately, using different optimality criteria, which can be sub-optimal. Both works use metrics such as the Tree Edit Distance (TED) or the Quotient Euclidean Distance (QED), where topological changes are modeled by edge collapse and node split operations. Consequently, geodesics exhibit significant shrinkage, especially between trees that significantly bend (see Fig.~\ref{fig:shrinkage_problem}-(a)) and between trees with large topological differences  (see Fig.~\ref{fig:shrinkage_problem}-(b)).}

Duncan \etal~\cite{duncan2018statistical} addressed this problem by representing 2D tree-shaped objects using curves that meet at the bifurcation points. While this representation is efficient, it is limited to simple 2D neuronal trees, \ie trees composed of one main branch and some lateral branches. In this paper, we generalize the framework of Duncan \etal~\cite{duncan2018statistical} to complex and arbitrary tree-shaped 3D objects. We propose a novel recursive representation where a tree-shaped object is characterized by a hierarchical organization of subtrees, \ie some side branches attached to the main branch. The task of comparing and registering two individual branches requires an optimal bending and stretching of one branch to align it onto the other. The task of comparing and registering two subtrees, each composed of one main branch and multiple side branches, is complex. It requires: \textbf{(i)} optimally sliding branches along the main branch to capture the topological differences between the two subtrees, and \textbf{(ii)} bending and stretching the branches to align them. By defining a metric that quantifies these deformations, \ie bending, stretching, and branch sliding, one can specify geodesic deformations between two subtrees as the shortest path under this metric, after factoring out shape-preserving transformations such as translation, scale, rotation, and re-parameterization. \hamid{(Note that parameterization defines correspondence~\cite{srivastava2011shape,kurtek2013landmark}.)} The geodesic length between two tree-shaped objects is then a measure of dissimilarity between the two objects. Defining this metric recursively across the entire hierarchy of the tree-shaped objects allows us to analyze complex tree-shaped 3D objects.  

In this paper, we develop this mathematical framework for deriving tools that:
\begin{itemize}
	\item Compute correspondences and geodesic deformations between tree-like 3D objects even when undergoing significant bending, stretching, and complex topological deformations.
	\item Compute statistical atlases, \ie means and principal modes of variation, of collections of tree-like 3D objects.
	\item Characterize the geometric and structural variability within a collection using probability distributions.
	\item Synthesize 3D tree-like structures \guan{using random} sampling.
\end{itemize}

\noi We demonstrate these tools using datasets such as 3D botanical trees and neuronal structures\guan{~\cite{ascoli2007neuromorpho,suo2012protocadherin,chen2014morphological}}.




The remainder of this paper is organized as follows; After reviewing the related work in Section~\ref{sec:related_work}, we introduce a novel tree-shape space (Section~\ref{sec:representation}) and a metric on this space (Section~\ref{sec:metric}) to quantify bending, stretching, and topological changes in tree-shaped 3D objects. We then develop algorithms for computing one-to-one correspondences and geodesics, \ie optimal deformation paths with respect to the chosen metric, between tree-shaped 3D objects (Section~\ref{sec:correspondences_geodesics}). Using these building blocks, we develop computational tools for computing statistical summaries of collections of tree-shaped 3D objects (Section~\ref{sec:statistics}) and for synthesizing novel tree-shaped 3D models (Section~\ref{sec:synthesis}). Finally, we present the results in Section~\ref{sec:results} and conclude in Section~\ref{sec:conclusion}.

\section{Related work}
\label{sec:related_work}



Statistical shape analysis has been studied extensively by the computer vision, computer graphics, and statistics communities. There are two subproblems, which are essential for statistical shape analysis: {\bf registration} and {\bf optimal deformation}. Registration is the problem of finding a one-to-one matching of points across objects, \ie deciding which point on one object matches which point on the other. Optimal deformation, or geodesics, is the problem of finding an optimal continuous sequence of shapes starting from one shape and ending at the other. The optimality is measured with respect to a physically motivated metric.    

There is a large body of literature that investigates these problems. Many papers are restricted to 3D objects that can be studied by bending and stretching. The idea is to treat shapes as points in a shape space equipped with a proper metric that measures the amount of bending and stretching that the shapes undergo. Equipping the space with a proper metric allows for comparing objects based on their shapes, computing geodesics, and performing statistical analysis, including regressions and shape synthesis.  The seminal work from Kendall's school~\cite{kendall1989survey,dryden:1998,allen2003space,kendall2009shape} represents shapes with point sets that are already registered and focused only on deformations. Other approaches such as medial surfaces~\cite{bouix:2001,gorczowski:2010} and level sets~\cite{osher:2003} either presume registration or solve for it using some independent pre-processing criterion such as \guan{the Minimum Description Length (MDL)}~\cite{davies:2010}. Kilian~\etal~\cite{Kilian:2007} represent surfaces by discrete triangulated meshes and compute geodesic deformation paths between them while assuming that the meshes are registered. Heeren~\etal~\cite{heeren:2012} propose a method for computing geodesic-based deformations of thin shell shapes, with extensions for computing summary statistics in the shell space~\cite{zhang2015shell}, but with known registration.

Some other papers solve for registration using shape descriptors while ignoring deformation; see~\cite{oliver:2011,laga20183d} for a detailed survey on the topic. \hamid{These methods sample (densely or sparsely) key points on the surface of the objects, compute shape descriptors at those locations, and then match the descriptors across the surfaces by minimizing an objective function composed of a data term and a set of regularization terms that impose constraints on the correspondence. However, feature points and shape descriptors at these feature points represent only partial information about the shape of objects. Also, since the mapping from shape to descriptors is often not one-to-one and often not invertible, statistics on descriptors (\eg the mean of two shape descriptors) do not correspond to the statistics on shapes (\eg the mean of the corresponding two shapes).}  


Recent works, on the other hand, tried to address the registration and optimal deformation problems jointly. Central to this problem is the definition of a Riemannian
metric (on the space of parameterized objects) preserved by the action of the relevant transformations, typically reparameterizations, translations, rotations, and perhaps scale. In particular, Srivastava \etal~\cite{srivastava2011shape} introduced a particular elastic metric in conjunction with a representation called the Square-Root Velocity Function (SRVF). It has been used for joint registration and geodesic computation between planar shapes and between curves in $\mathbb{R}^d, d \ge 3$~\cite{laga2014landmark}. It has been later extended to the analysis of parameterized surfaces~\cite{kurtek2011elastic,jermyn2012elastic,kurtek2013landmark,laga2017numerical}. These methods, however, are limited to 3D models with fixed topology. They cannot capture and model topological variabilities such as those present in plant roots, botanical trees, and neuronal structures.


Closer to this paper are the techniques based on tree statistics. The seminal work of Billera \etal~\cite{billera2011geometry}  proposed the notion of continuous tree-space and its associated tools for computing summary statistics. Some variants of this idea were developed for the statistical analysis of tree-structured data, \eg~\cite{owen2011A,aydn2009a}. However, these works only consider the topological structure of trees and ignore the geometric attributes of edges, limiting their usage. As a result, several papers have defined a more general tree-space. Examples include Feragen \etal's framework~\cite{feragen2010geometries,feragen2011means,feragen2013toward,feragen2013tree-shapce}, which proposed a tree-shape space for computing statistics of airway trees, and its extension to complex botanical trees~\cite{wang2018the,wang2018statistical}. Despite their efficiency and accuracy in certain situations, these techniques exhibit three main fundamental limitations. \textbf{First}, they use the Quotient Euclidean Distance (QED) \hamid{where the geodesic between two branches is equivalent to a linear interpolation between them.  As such,  } the QED is not suitable for capturing large elastic deformations, \ie bending and stretching, of the branches. This is illustrated in Fig.~\ref{fig:shrinkage_problem}-(a), which shows an example of a geodesic between two simple trees that are in correspondence but their branches significantly bend. Observe that under the QED metric, the branches unnaturally shrink along the geodesic but not with the proposed elastic geodesic. \textbf{Second}, they represent tree shapes as a \emph{father-child branching} structure, which leads to significant shrinkage along the geodesics between trees that exhibit large topological differences; see Fig.~\ref{fig:shrinkage_problem}-(b). \textcolor{black}{In fact, the QED metric involves node split and edge collapse operations. However,  edge collapse operations result in the unnatural shrinkage of the trees along the geodesic, \ie the trees become small (due to edge collapse) before growing back due to node split. This shrinkage will be significant if the source and target trees significantly differ in topology, since computing the geodesic between them would require a large number of edge collapse and node split operations; please refer also to the third row of Fig.~\ref{fig:comparison_3botanGeods}.} \textbf{Third}, branch-wise correspondences need to be manually specified, especially when dealing with complex tree-like structures.

In contrast to the state-of-the-art, we propose in this paper a statistical framework that is more suitable for analyzing complex tree-shaped 3D objects such as botanical trees, plant roots, and neuronal structures.  It  builds upon and extends  the recent work of Duncan \etal~\cite{duncan2018statistical}, which  showed that \emph{main-side branching} representation is more efficient for capturing topological changes.  To the best of our knowledge, this is the first approach that deals with the statistical modeling of such complex tree-shaped 3D objects. In addition to classical applications such as the classification of tree-shaped 3D objects based on their shape and the computation of correspondences and geodesics, we demonstrate through experiments that this framework can produce reasonable statistical summaries. It also enables the synthesis of tree-shaped 3D objects, either randomly or in a controlled manner, and thus can be used to populate virtual environments and generate simulated data for training deep neural networks\hamid{, \eg for the 3D reconstruction from monocular RGB images~\cite{han2019image}.}




%

\section{Representation}
\label{sec:representation}

\begin{figure}[t]
	\center
	\includegraphics[width=0.2\textwidth]{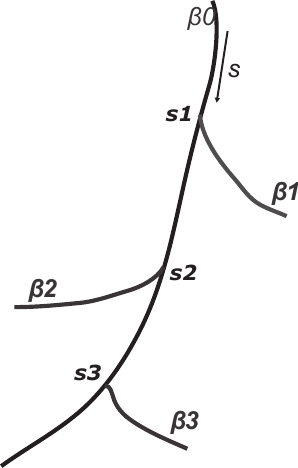}
	\caption{The representation of a simple tree shape.}
	\label{fig:simpleTree}
\end{figure}


The input to our framework is a collection of tree-like 3D objects such as plant roots, botanical trees, or neuronal structures. We first skeletonize each object and convert it into a set of curves, one for each branch. 
\textcolor{black}{The skeleton of each branch can now be seen as a curve of the form $\curvefunc: [0, 1] \to \rthree$ such that $\curvefunc(\parameter) = (x(\parameter), y(\parameter), z(\parameter)\guanwang{)}$.  The parameter $\parameter\in[0,1]$ is expressed as the proportion of arc length along the skeletal curves of $\branch$. Let $\preshapecurves$ be the space of all such curves.}
We then augment the skeletal curves with additional attributes such as the thickness at each skeletal point of the branch it represents. Each branch $\branch$ can then be seen as a continuous curve in $\rthree \times \rplus$ of the form:
\begin{equation}
	\begin{aligned}
		\branch:& \  [0, 1]\rightarrow \rthree\times \rplus,  \\
		&  \branch(\parameter) \equiv (\curvefunc(\parameter), \thickness(\parameter)) = (x(\parameter), y(\parameter), z(\parameter), \thickness(\parameter)).
	\end{aligned}
	\label{eq:curves_and_thickness}	
\end{equation}

\noi Here,  \hamid{$\rplus$ is the set of strictly positive real numbers}, $\curvefunc(\parameter) = (x(\parameter), y(\parameter), z(\parameter))$ is the 3D coordinates of the skeletal curve at $\parameter$,  and $\thickness(\parameter)$ is its thickness. 

We organize the branches into layers, with layer zero representing the main branch. With this setup, a tree-like object can be represented recursively by writing  $\proot= (\branch^0,\{\proot^i, \parameter^i\}_{i=1}^{n})$, where:
\begin{itemize}
	\item $\branch=\branch^0$ is the main branch, \eg the trunk in the case of a botanical tree. 
	\item $\proot^i$ is a subtree attached to $\branch^0$ at the bifurcation point $\branch^0(\parameter^i)$ with $\parameter^i \in [0, 1]$.  %
	
\end{itemize}

\noi This is illustrated in Figure~\ref{fig:simpleTree}. If a subtree  $\proot^i$  further contains subtrees, then that can be represented recursively by repeating this idea at every level.

To make the representation  translation and scale invariant, we first translate each tree so that the start point of its main branch is located at the origin, and then scale the entire tree so that the length of its main branch is one. Note that invariance to scale  is optional as it may not be  required for some applications such as growth analysis. In what follows, we assume that all the objects have been normalized for translation and scale, and thus they are elements of a pre-shape space\footnote{A pre-shape space of curves, surfaces, or trees is the space of all such curves, surfaces, or trees before removing or factoring out rotations and reparameterizations, which are shape-preserving transformations.} of trees, hereinafter referred to as pre-tree shape space and denoted by  $\preshapetrees$. It can be recursively defined  as follows.
\begin{itemize}
	\item Let $\preshapetrees = \preshapetrees^0$ be the pre-tree shape space of all trees that only have the main branch, a single curve. We will call such structures a level-0 tree. It can be shown that 
	\begin{equation}
		\preshapetrees^0 = \preshapecurves \times \mathcal{T}, 
	\end{equation}
	\textcolor{black}{where $\mathcal{T}$ is the space of functions of the form $[0, 1] \to \rplus$, and represents the thickness of the branch at each point along its skeleton.} Note that $\preshapetrees^0$ is in fact the space of all curves defined by Eqn.~\eqref{eq:curves_and_thickness}.  
	
	\item Let now $\preshapetrees^k, k\ge1$ be the pre-tree shape space of all trees that can have up to $k$ levels. This space can be recursively defined as follows;
	\begin{equation}
		\preshapetrees^k = \preshapetrees^0 \times  {\bigsqcup_{n=0}^{\infty}} \left(  \preshapetrees^{k-1} \times [0, 1]  \right)^n. 
		\label{eq:tree_preshape_space}
	\end{equation}
	This means that a level-$k$ tree can have $n$ structures \guan{attached to} the main branch, each structure being a level-$(k-1)$ tree, at specified attachment points. 
\end{itemize}


\section{Metric for tree-shapes}
\label{sec:metric}

In order to perform statistical shape analysis and modeling of the geometry and topology of a collection of tree shapes, we need to define a distance metric in the shape space of trees  and a mechanism for computing correspondences and geodesics between elements in that space. \textcolor{black}{The metric needs to quantify shape-changing deformations, \ie bending, stretching, and topological changes, while being invariant to shape-preserving transformations, \ie translations, global scaling, and rotations. }

\subsection{The elastic metric for the shape of branches}
\label{sec:elastic_metric_branches}

We propose to use an elastic metric that captures the bending and stretching of the branches.    Consider a path $\deformationpath_{\branchone \to \branchtwo}: [0, 1] \to \preshapetrees$, which deforms $\branchone$ onto  $\branchtwo$.  In other words, $\deformationpath(0) = \branchone, \deformationpath(1) = \branchtwo$, and $\forall t \in [0, 1], \deformationpath(t) \in \preshapetrees$.  Its length $L[\deformationpath]$ is given by:
\begin{equation}
	L[\deformationpath] = \int_0^1 \innerd{ \dot{ \deformationpath  } (\timeparam) }{ \dot{ \deformationpath}(\timeparam)} ^{\frac{1}{2}} d\timeparam,
	\label{eq:general_metric}
\end{equation}

\noi where  $ \dot{ \deformationpath  } = \frac{\partial \deformationpath(\timeparam)}{ \partial \timeparam}$ is the velocity vector, and  $ \innerd{ \cdot}{ \cdot}$ is a certain metric. 
Instead of using the $\ltwo$ metric as in~\cite{feragen2013toward,wang2018the}, we use a metric that quantifies the bending and stretching of the branches.

Physically, bending can be quantified by measuring changes in the orientation of the tangent vectors to the skeletal curve $\curvefunc$ of $\branch$ along the deformation path. Stretching can be decomposed into two components; the first one is related to the elongation of the skeletal curve $\curvefunc$, which can be quantified by looking at how the magnitude of the tangent vector to $\curvefunc$  at every point $\parameter$ changes along the deformation path. The second component is related to changes in the thickness of the branch and can be measured by looking at how the thickness $\radiusfunc$ at each point $\parameter$ varies along the deformation path.  Let us write:
\begin{equation}
	\curvefunc'(\parameter) = \frac{\partial \curvefunc (\parameter) }{\partial \parameter} =  \theta(\parameter) e^{\phi(\parameter)},
\end{equation}
where $\theta(\parameter)$ is the unit tangent vector to $\curvefunc$ at $\parameter$, and $e^{\phi(\parameter)}$  is the magnitude of the tangent vector to $\curvefunc$ at $\parameter$. (We also refer to $e^{\phi(\parameter)}$ as the speed.) In other words:
\begin{equation}
	e^{\phi(\parameter)} = \left\| \curvefunc' (\parameter) \right\| = \left\| \frac{\partial \curvefunc (\parameter)}{\partial s} \right\|, \text{ and } \theta(\parameter) = \frac{{\curvefunc}' (\parameter)}{\| {\curvefunc}' (\parameter) \|}.
\end{equation}

\noi One can then define an elastic metric as the weighted sum of the changes of $\phi$, $\theta$, and $\radiusfunc$ along the path $\deformationpath$. In other words:
\begin{eqnarray}
	\innerd{\dot{\beta}}{\dot{\beta}}  & \equiv&   a \int \inner{ \dot{\theta}(\parameter) }{ \dot{\theta}(\parameter)  }    e^{\phi(\parameter)}  d \parameter +  \nonumber \\
	& 	      &   b \int \dot{\phi}(\parameter)^2  e^{\phi(\parameter)}  d \parameter +   c \int \dot{\radiusfunc}(\parameter)^2  e^{\phi(\parameter)}  d \parameter. 
	\label{eq:full_elastic_metric}
\end{eqnarray}

\noi Here, $\dot g$ denotes the derivative of the function $g$ with respect to  $\timeparam$. The first term of Eqn.~\eqref{eq:full_elastic_metric} quantifies bending by measuring changes in the orientation of the tangent vector to the skeletal curve. The second and third terms quantify stretching.  The first two terms are equivalent to the elastic metric between curves in $\rthree$, which has been introduced by Srivastava \etal~\cite{srivastava2011shape}.   While the full elastic metric of Eqn.~\eqref{eq:full_elastic_metric} is the most appropriate for analyzing the shape of branches, it is computationally very expensive.  To reduce the computation time, Srivastava \etal~\cite{srivastava2011shape} introduced the Square Root Velocity Function (SRVF) of a curve $\curvefunc$ as a mapping $\srvfmap$ of the form:
\begin{equation}
	\label{eq:SRVF_representation}
	\srvfmap(\curvefunc)(\parameter)= \left\{
	\begin{array}{@{}l@{}l@{ }}
		\frac{{\curvefunc}'(\parameter)}{\sqrt{\parallel{\curvefunc}'(\parameter)\parallel}}  & \text{ if } {\curvefunc}'(\parameter)  \text{ exists and } \| {\curvefunc}'(\parameter) \| \neq 0,  \\
		0 & \text{ otherwise.}
	\end{array}
	\right.
\end{equation}

\noi More importantly, Srivastava \etal~\cite{srivastava2011shape}  showed that the $\ltwo$ metric in the space of SRVFs is equivalent  to the elastic metric of Eqn.~\eqref{eq:full_elastic_metric} when setting $a = 1, b = \frac{1}{4}$, and $c=0$. In our case, we define the Extended SRVF (ESRVF) of a branch $\branch = (\curvefunc, \radiusfunc)$ as the pair $\srvf = \text{SRVF}( [\curvefunc,  c\radiusfunc])$, with $c$ is the weight of the third term of Eqn.~\eqref{eq:full_elastic_metric}.  We can easily show that the $\ltwo$ metric in the space of ESRVFs,  hereinafter denoted by $\esrvfspace$,  is equivalent to the elastic metric of Eqn.~\eqref{eq:full_elastic_metric} when setting $a = c = 1, b = \frac{1}{4}$.    This is very important since, instead of working with the complex elastic metric, we can represent the shape of each branch  using its ESRVF and use the associated $\ltwo$ metric to capture the bending and stretching of branches. 

The  ESRVF representation inherits all the properties of the SRVF. \textcolor{black}{It is invariant to translations since it is defined in terms of derivatives. Also,} the action of the reparameterization group on the ESRVF is by isometry. In other words, $\forall \diffeo\in \diffeos, \|\srvf_1 - \srvf_2\| = \|\srvf_1 \circ \diffeo - \srvf_2 \circ \diffeo\|$.  \guan{Here $\diffeos$ is the space of all diffeomorphisms of the domain $[0, 1]$ to itself.} With this,  \guan{aligning} two branches $\branch_1$ and $\branch_2$ becomes the problem of finding the optimal rotation $\rotation^* \in \rotations$ and reparametrization $\diffeo^* \in \diffeos$ such that:
\begin{equation}
	(\rotation^*, \diffeo^* ) = \argmin_{\rotation \in \rotations, \diffeo \in \diffeos} \|\srvf_1  - \rotation (\srvf_2, \diffeo)\|^2.
	\label{eq:srvf_metric}
\end{equation}

\noi Here,  $(\srvf_2, \diffeo) = \text{ESRVF}(\branchtwo \circ \diffeo)$. \hamid{Thus, $\rotation (\srvf_2, \diffeo)$ is the process of applying the rotation $\rotation$ to $(\srvf_2, \diffeo) = \text{ESRVF}(\branchtwo \circ \diffeo)$.} The distance between the two branches is then defined as $ \|\srvf_1  - \rotation^* (\srvf_2, \diffeo^*)\|$. \textcolor{black}{Thus, the invariance to rotations is achieved through the optimization  over $\rotations$ in Eqn.~\eqref{eq:srvf_metric}.}

With this setup, each tree shape $\proot= (\branch^0,\{\proot^i, \parameter^i\}_{i=1}^{n})$   will be represented with the collection of the ESRVFs of its branches. We refer to this recursive representation of the form $\srvft= (\srvf^0,\{\srvft^i, \parameter^i\}_{i=1}^{n})$ as the  Square Root Velocity Function Tree (SRVFT). The pre-tree shape space in which $\srvft$ resides can now be recursively defined as follows;
\begin{itemize}
	\item Let $\preshapesrvts = \preshapesrvts^0$ be the pre-tree shape space of all SRVFTs  that have one single branch. One can  show that 
	\begin{equation}
		\preshapesrvts^0 = \preshapesrvfs \times \mathcal{T},
	\end{equation}
	\textcolor{black}{where $\preshapesrvfs = \text{SRVF}(\preshapecurves)$, \ie the image of  $\preshapecurves$ using the SRVF map, and $\mathcal{T}$ is the space of functions of the form $[0, 1] \to \rplus$ and which represent the thickness of the branch at each point along its skeleton.} 
	
	\item Let now $\preshapesrvts^k, k\ge1$ be the pre-tree shape space of all SRVFTs that can have up to $k$ levels. This space can be recursively defined as follows;
	\begin{equation}
		\preshapesrvts^k = \preshapesrvts^0 \times {\bigsqcup_{n=0}^{\infty}} \left(  \preshapesrvts^{k-1} \times [0, 1]  \right)^n. 
	\end{equation}
\end{itemize}

\noi This representation has many nice properties.  For instance, 
\begin{itemize}
	\item The mapping from  $\preshapetrees$  to $\preshapesrvts$   is surjective.
	\item The mapping, however,  is not injective, but two curves are mapped to the same SRVF space if and only if they are translates of each other. 
	\item The inverse mapping has a \textcolor{black}{closed} analytical form. In other words, for a given pair $(\srvf, \radiusfunc)$,  one can find, analytically and up to  a translation,  a branch $\branch$ whose SRVF is $(\srvf, \radiusfunc)$.
\end{itemize}

\noi As a consequence of the last property, and since the locations of the subtrees is encoded in the SRVFT representation, the inverse mapping, \ie the mapping from $\preshapesrvts$  to $\preshapetrees$, can be found in an analytical form, up to a global translation.

\subsection{The elastic metric for comparing tree-shapes}
\label{sec:metric_fixed_topology}

The next step is to define a metric in the space of SRVFTs and a mechanism for computing correspondences and geodesics between points in this space. We do this by generalizing the metric proposed by Duncan \etal~\cite{duncan2018statistical} for the analysis of simple trees, composed of the main branch and a set of side branches, to complex 3D trees of arbitrary levels of hierarchy. Let $\proot_1$ and $\proot_2$ be two tree shapes represented with their SRVFTs $\srvft_1$ and $\srvft_2$ in $\preshapesrvts$. Let us first assume that the two trees have the same number of branches and are in branch-wise correspondence. We define the distance between   $\proot_1$ and $\proot_2$ recursively as follows;
\begin{equation}
	\label{eq:init_distanceTwoTrees}
	\begin{aligned}
		d^2_{\preshapesrvts}(\srvft_{1}, \srvft_{2}) =& \lambda_m \parallel\srvf^0_{1} - \srvf^0_{2} \parallel^2 +  
		\lambda_s \sum_{i=1}^{n} d^2_{\preshapesrvts}(\srvft^i_{1}, \srvft^i_{2})  + \\
		& \lambda_p \sum_{i=1}^{n} ({s}^{i}_{1} - {s}^{i}_{2})^2.
	\end{aligned}
\end{equation}

\noi The first term of Eqn.~\eqref{eq:init_distanceTwoTrees} measures the amount of bending and stretching needed to align one main branch to another. The second term, which is computed recursively using Eqn.~\eqref{eq:init_distanceTwoTrees}, is the dissimilarity between two subtrees $\srvft^{i}_{1}$ and $\srvft^{i}_{2}$ attached, respectively, to the bifurcation points ${s}^{i}_{1}$ and ${s}^{i}_{2}$. Note that when $\srvft_{1}$ and $\srvft_{2}$ are null trees, then we set $d_\preshape(\srvft_{1}, \srvft_{2}) = 0$. The third term is the distance between the locations of the corresponding main branches of the subtrees $\srvft^i_{1}$ and $\srvft^i_{2}$. The parameters $\bm{\lambda} = (\lambda_m, \lambda_s, \lambda_p)$ control the relative cost of deforming the main branch, deforming the subtrees connected to the main branch, and moving the positions of the subtrees connected to the main branch.

\noi Section~\ref{sec:ablation} shows the effects of these parameters with practical examples.

\subsection{Invariant metric}
\label{sec:invariant_metric}
A proper metric for comparing the shape of tree-like objects should be invariant to shape-preserving transformations. The translation and scale have already been factored out in a pre-processing step. Now, we consider rotations, reparameterization of the branches, and permutations of the orders of the lateral subtrees attached to a branch. Let: 
\begin{itemize}
	\item $\rotation \in \rotations$ be a global rotation applied to the entire tree, 
	
	\item $\diffeotree = (\diffeo^0, \{\diffeotree^{i}\}_{i=1}^{n})$ be the reparameterization of the main branch and its subtrees.  Specifically,  $\diffeo^0 \in \diffeos$ is a diffeomorphism that applies to the main branch of $\srvft$ and $\diffeotree^{i}$ the reparameterization, defined recursively, of the $i-$th subtree. 
	
	\item $\permutetree = (\permute^0, \{ \permutetree^{i}\}_{i=1}^{n_k})$ such that $\permute^0 \in \permutetree$ is the   permutation of the orders of the lateral subtrees on their corresponding main branch $\srvf^0_k$ (here, $\permutetree$ is the space of such permutations), and $\permutetree^{i}$ defines recursively these permutations  for the $i-$th subtree.
\end{itemize}

\noi Let $(\srvft, \rotation,  \diffeotree, \permutetree) $ denote the result after applying these three transformations to $\srvft$. Since these transformations are shape-preserving, $\srvft$ and $(\srvft, \rotation,  \diffeotree, \permutetree) $ have the same shape. They thus are equivalent under the action of global rotations, branch reparameterizations, and permutations of the indices of the lateral subtrees of the different branches. \hamid{We treat $(\rotation,  \diffeotree, \permutetree) $ as a shape-preserving nuisance action and define a quotient space on the closures of orbits under this action. We then define the rotation, reparameterization, and index permutation-invariant distance between two tree shapes $\srvft_1$ and $\srvft_2$ as the distance between their orbits, which we denote by $[\srvft_1]$ and $[\srvft_2]$. It is defined as 
	\begin{equation}
		\begin{aligned}
			d^2\left([\srvft_1], [\srvft_2]\right) = \inf_{
				\tilde{\srvft}_1 \in [\srvft_1], 						\tilde{\srvft}_2 \in [\srvft_2]
			}
			\{ d^2_{\preshapesrvts}\left(\tilde{\srvft}_1, \tilde{\srvft}_2\right)\}.
		\end{aligned}
	\end{equation}
	
	\noi In other words, to compute the geodesic distance between two equivalence classes, or two orbits in the quotient space, $[\srvft_1]$ and $[\srvft_2]$, one can select two representative instances $\srvft_1 \in [\srvft_1]$  and $\srvft_2 \in [\srvft_2]$, and then optimize over all possible rotations, branch reparameterizations, and branch order permutations of the two instances. From the computational point of view, we fix $\srvft_1$ and optimize over all possible rotations, branch reparameterizations, and branch order permutations of $\srvft_2$. In other words,} 
\begin{equation}
	\label{eq:invariantdistance}
	\begin{aligned}
		d^2\left(\srvft_1, \srvft_2\right) = \inf_{\tiny{\begin{tabular}{c} 
					$\rotation \in \rotations$ \\
					$\diffeotree \in \diffeotrees$ \\
					$\permutetree \in \permutetreespace$
				\end{tabular}
		}}  d^2_{\preshapesrvts}\left(\srvft_1, (\srvft_2,  \rotation, \diffeotree, \permutetree)\right),
		%
	\end{aligned}
\end{equation}
\noi where
\begin{equation}
	\label{eq:invariant_metric}
	\begin{aligned}
		d^2_{\preshapesrvts}\left(\srvft_1, (\srvft_2,  \rotation, \diffeotree, \permutetree)\right)  =&  \lambda_m \parallel\srvf^0_{1} - \rotation(\srvf^0_{2}, \diffeo_0) \parallel^2  +  \\
		& \lambda_s \sum_{i=1}^{n} d^2 \left( \srvft^i_{1}, \srvft^{\permutetree(i)}_{2}   \right) + 	\\
		& \lambda_p \sum_{i=1}^{n} \left({s}^{i}_{1} - {s}^{\permutetree(i)}_{2}\right)^2.
	\end{aligned}
\end{equation}

\noi The optimal registration of $\srvft_2$ onto $\srvft_1$ can then be found by solving the following optimization problem:
\begin{equation}
	\label{eq:registration}
	(\tilde\rotation,   \tilde\diffeotree, \tilde\permutetree) = \argmin_{\rotation, \diffeotree,\permutetree} d^2_{\preshapesrvts}\left(\srvft_1, \srvft_2, \rotation,   \diffeotree, \permutetree \right).
\end{equation}

\noi The challenge now is how to optimize simultaneously over all these  transformation spaces. This will be discussed in Section~\ref{sec:correspondences_geodesics}.

%
%

\subsection{Trees with different numbers of side branches}
\label{sec:metric_variable_topology}

The formulation presented in Sections~\ref{sec:metric_fixed_topology} and~\ref{sec:invariant_metric} assumes that the trees have the same number of branches and the same topology. Thus, a one-to-one branch-wise correspondence exists. However, this is not the case in practice. Two tree models rarely have the same number of bifurcation points and branches. As a result, these trees lie on disjoint subspaces, making it difficult to compare them and compute geodesics. One way to overcome this issue is by adding null branches, \ie branches of length zero. 

First, we define the order of a tree $\proot$, and subsequently of $\srvft$, by taking the maximum of the number of bifurcation points on  its branches. Let $\treeorder_i, i=1, 2$ be the order of the trees $\srvft_1$ and $\srvft_2$, and let $\treeorder = \max(\treeorder_1, \treeorder_2)$. Then, we add to each branch $l$ of the $i-$th tree, $\treeorder - \treeorder_i^l$ null branches. (Here, $ \treeorder_i^l$ is the order of branch $l$. ) By doing so, all  $k$-level trees become elements of the same pre-tree shape space.   This facilitates their comparison, and subsequently putting them in correspondence, and computing geodesic paths. \hamid{The location of each additional null branch, \ie the value of its  parameter $s$, is initialized to the $s$ value of its initial corresponding branch. These correspondences will then get updated during the optimization over the branch-wise correspondences.}


\section{Correspondences and geodesics}
\label{sec:correspondences_geodesics}

With this formulation, the metric in the pre-tree shape space $\preshapesrvts$ is a weighted sum of $\ltwo$ distances. This  has significant practical benefits. In particular, instead of working in the original space of 3D objects, which is nonlinear and equipped with a complex metric, one can map the input tree-like objects into the SRVFT space, use the simple metric of Eqn.~\eqref{eq:invariantdistance} to compute correspondences, geodesics, and statistics using standard tools from vector calculus, and finally map the results back to the original space of trees for visualization. Since the SRVFT mapping is one-to-one and onto (up to global translation), the inverse mapping exists, is unique, and more importantly, has a closed analytical form. This invertibility property is very important in practice. 

Given two tree-shaped objects $\proot_1$ and $\proot_2$ represented with their SRVFTs $\srvft_1$ and $\srvft_2$,  we obtain one-to-one correspondences by solving the optimization problem of Eqn.~\eqref{eq:registration}. This is done by alternating the optimisation over $\rotations$, $\diffeos$, and $\permutetreespace$, \ie
\begin{itemize}
	\item Optimize over $\rotation \in \rotations$, assuming fixed parameterizations and permutations, which is straightforward to achieve using Procrustes analysis~\cite{kendall1989survey}. Then,
	\item Optimize over $\diffeotree \in \diffeotrees$ and 	$\permutetree \in \permutetreespace$  while assuming a fixed rotation. 
\end{itemize}

\noi In this section,  we discuss the process of optimizing over  $\diffeotree \in \diffeotrees$ and $\permutetree \in \permutetreespace$. We first consider the simple case of trees with two levels of hierarchy, \ie trees that only have one main branch and multiple side, or lateral, branches (Section~\ref{sec:simple_trees}). We then show how this method can be generalized to complex tree-shaped 3D objects of arbitrary branching structures (Section~\ref{sec:complex_trees}). Algorithms~\ref{alg:alignment_procedure} and~\ref{alg:reparamPermute} summarize the overall procedures.

\subsection{Correspondence between simple tree shapes}
\label{sec:simple_trees}

Assume the simple case where the two trees are composed of only one main branch and some lateral branches. In this case $\diffeotree = (\diffeo^0, \{\diffeo^i\}_{i=1}^n )$ and $\permutetree = (\permute_0 )$. \hamid{Since the first term of Eqn.~\eqref{eq:invariant_metric}, which measures the distance between the two main branches, is independent of the other two terms, 
	we first find, using the approach of Srivastava \etal~\cite{srivastava2011shape}, the optimal re-parameterization $\diffeo^0$ that elastically aligns the main branch of $\srvft_2$ onto the main branch of $\srvft_1$.} Next, we formulate the joint problem of \textbf{(1)} finding which side branch of $\srvft_2$ is matched to which branch of $\srvft_1$ (\ie using the third term of Eqn.~\eqref{eq:invariant_metric}) and \textbf{(2)} elastically registering  the corresponding branches (\ie using the second term of Eqn.~\eqref{eq:invariant_metric}), as a linear assignment problem. That is, we build a pairwise distance matrix $E$ of size $n_1\times n_2$ where 
\begin{equation}
	E_{ij} = \lambda_s \inf_{\rotation, \diffeo} \| \srvf_1^i - \rotation(\srvf_2^j, \diffeo) \|^2 + \lambda_p \left(s^i_1 - s_2^j\right)^2.
\end{equation}

\noi We  then apply the  Kuhn-Munkres algorithm, also known as the Hungarian Algorithm~\cite{kuhn1955hungarian}, to compute the optimal matching $\permute_0$. Next, for each matching branches $\srvf_1^i$ and $\srvf_2^{\permute_0(i)}$, we compute the reparameterization $\diffeo_i$ as
\begin{equation}
	(\rotation_i,\diffeo_i) = \argmin_{\rotation, \diffeo} \| \srvf_1^i - \rotation(\srvf_2^{\permute_0(i)}, \diffeo) \|^2
\end{equation}

\noi by using the approach of Srivastava \etal~\cite{srivastava2011shape}. Note that, in contrast to Duncan \etal~\cite{duncan2018statistical}, which builds an $(n_1+n_2)\times(n_1+n_2)$ cost matrix, we solve the linear assignment problem by building a $\max(n_1, n_2)\times\max(n_1, n_2)$ cost matrix. In this way,  we reduce  the worst-case time complexity from $O((n_1+n_2)^3)$ to $O(\max(n_1, n_2)^3)$ and reduce the computation time by a factor of $8$. This is important especially when processing recursively  tree shapes with many layers.

\begin{algorithm}[t]
	\caption{\label{alg:alignment_procedure} Alignment procedure. }
	
	\textbf{Input:}
	\begin{itemize}
		\item $n\_\text{levels} = 2, 3 \text{ or } 4$.
		\item$\srvft^1 $, $\srvft^2$, the SRVFT representations of two tree-shaped objects to be aligned.
		\item$\bm{\lambda} = (\lambda_m, \lambda_s, \lambda_p)$, the  distance weight parameters.
		\item$n_\text{max}$, the  number of iterations.
	\end{itemize}
	
	\textbf{Output:}
	\begin{itemize}
		\item$\rotation$, $\diffeotree$, and $\permutetree$.
		\item$E_\text{final}$, the  distance between $\srvft^1$ and $\srvft^2$.  
	\end{itemize}
	
	\begin{algorithmic}[1]
		\Procedure{AlignTrees}{$\srvft^1$, $\srvft^2$, $n\_levels$}
		\State $(\diffeotree, \permutetree) = \text{ ReparamPermute}(\rotation_\text{id}, \srvft^1, \srvft^2, \bm{\lambda}, n\_\text{levels})$;
		
		\For{$n_\text{max}$ iterations}
		\State $\rotation = \text{ Procurstes}(\srvft^1, (\srvft^2, \diffeotree, \permutetree), \bm{\lambda})$.
		\State $(\diffeotree, \permutetree) = \text{ ReparamPermute}(\rotation, \srvft^1, \srvft^2, \bm{\lambda}, n\_\text{levels})$;
		
		\EndFor
		
		\State $E_\text{final}  =  d_{\preshapesrvts}(\srvft^1, \rotation(\srvft^2, \diffeotree,  \permutetree))$
		\State Return $(\rotation, \diffeotree, \permutetree, E_\text{final} )$.
		\EndProcedure
	\end{algorithmic}
	\label{alg:diantance_refinement}
\end{algorithm}

\guanwang{Note that, when solving the optimization of Eqn.~\eqref{eq:registration}, the relative cost of deforming the main branch, deforming the subtrees connected to the main branch, and moving the positions of the subtrees connected to the main branch is controlled by the parameters  $\bm{\lambda} = (\lambda_m, \lambda_s, \lambda_p)$. The parameter $\lambda_m$ quantifies bending and stretching of the main branch. $\lambda_s$ allows the recursion over the subtrees and quantifies the similarity between subtrees. $\lambda_p$ quantifies branch sliding. In summary,
	\begin{itemize}
		\item High values of $\lambda_p$ and low values of $\lambda_m$ will penalize branch sliding. Thus, the algorithm will try to match side branches that are close to each other. It will prioritize correspondences that will result in bending and stretching over those that will cause changes in topology.
		\item Low values of $\lambda_p$ and  high values of $\lambda_m$  will have the opposite effect.
		\item The parameter $\lambda_s$ controls the contribution, or importance, of the subtrees to the overall distance measure. 
	\end{itemize}
	
	\noi Section~\ref{sec:ablation} shows, with examples, the effect of different values of these parameters.
}

\subsection{Extension to complex tree shapes}
\label{sec:complex_trees}

To handle complex trees, one can recursively run the procedure described in Section~\ref{sec:simple_trees} over the entire hierarchy of the trees. By doing so, however, the computation time would increase exponentially. We approximate this procedure by considering three levels at a time. That is, we first consider the main branch and the subtrees attached to it. However, for each subtree, we only consider its main branch and the side branches attached to it. To align two trees, we first align their main branches, compute the assignment matrix where each entry $E_{ij}$ is the optimal distance between the subtree \guan{$\srvft_1^i$} and \guan{$\srvft_2^j$}. Since these two subtrees have only two levels, $E_{ij}$ can be computed using the procedure described in Section~\ref{sec:simple_trees}. Similar to Section~\ref{sec:simple_trees}, we apply the Hungarian algorithm to find the assignment $\permute_0$ which matches subtrees on \guan{$\srvft_2$} onto subtrees on \guan{$\srvft_1$}. By repeating this procedure recursively on $\srvft_1^i$ and $\srvft_2^{\permute_0(i)}$, we obtain a full matching between the complex trees. Algorithms~\ref{alg:alignment_procedure} and~\ref{alg:reparamPermute} summarize the overall procedure.

\begin{algorithm}[t]
	\caption{\label{alg:reparamPermute} Reparameterization and permutation. }
	\begin{itemize}
		\item $n\_\text{levels} = 2, 3 \text{ or } 4$.
		\item$\srvft^1 $, $\srvft^2$, the SRVFT representations of two tree-shaped objects to be aligned.
		\item$\bm{\lambda} = (\lambda_m, \lambda_s, \lambda_p)$, the  distance weight parameters.
		\item$n_1, n_2$, the number of side branches in $\srvft^1$, respectively, $\srvft^2$.
		\item \textbf{DynamicProgQ} denotes the dynamic programming algorithm, as in Srivastava et al.~\cite{srivastava2011shape}, to compute the optimal reparameterization between two SRVF curves.
		\item \textbf{LAPJV} denotes to solve the \textbf{L}inear Sum \textbf{A}ssignment \textbf{P}roblem using the algorithm of \textbf{J}onker and \textbf{V}olgenant~\cite{jonker1988shortest}.
	\end{itemize}
	
	\textbf{Output:}
	\begin{itemize}
		\item $\diffeotree$ and $\permutetree$.
	\end{itemize}
	
	\begin{algorithmic}[1]
		\Procedure{ReparamPermute}{$\rotation$, $\srvft^1$, $\srvft^2$, $\bm{\lambda}$, $n\_\text{levels}$}
		
		\State $N = \max(n_1,n_2)$.
		\For{$i=1:N, j=1:N$}
		\If{$n\_\text{levels} = 3 \text{ or } 4$ }\vspace{2pt}
		\State $\begin{aligned}
			(\rotation, \diffeotree_{ij}, \permutetree_{ij}, \guan{E_{ij}})  =   \text{ AlignTrees}&(\srvft^1_i, \srvft^2_j, \\
			& n\_\text{levels}-1).
		\end{aligned}$
		\Else
		\State $\diffeo_{ij} = \text{ DynamicProgQ}(\srvf^1_i, \rotation \srvf^2_j)$ 
		\State $E_{ij} = \lambda_m \| \srvf^1_0 - \rotation(\srvf^2_0, \diffeo_{ij}) \|^2 + \lambda_p (s^1_i -s^2_j)^2$. 
		\EndIf
		\EndFor
		
		\State $\guan{\permutetree_0} = \text{ LAPJV}(E)$.
		\State $\diffeo_0 = \text{ DynamicProgQ }(\srvf_0^1, \srvf_0^2)$.
		
		\For{$i=1:N, j=1:N$}
		\If{$n\_\text{levels} = 3 \text{ or } 4$ }
		\State $(\diffeotree_{i}, \permutetree_{i} ) =  \text{ AlignTrees}(\srvft^1_i, \srvft^2_{\permutetree_0(i)}, n\_\text{levels})$.
		\State $\diffeotree_ \leftarrow \diffeotree_ \cup \{\diffeotree_{i} \}$. 
		\State $\permutetree \leftarrow \permutetree \cup \{\permutetree_{i} \}$. 
		\Else
		\State $\diffeotree \leftarrow \diffeotree \cup \diffeo_{i\permutetree_0(i)}$.
		\State $\permutetree \leftarrow \permutetree \cup \{\permutetree_0\}$.
		\EndIf
		\EndFor
		
		\State Return $\diffeotree, \permutetree$.
		\EndProcedure
	\end{algorithmic}
	\label{alg:reparamepermute}
\end{algorithm}

\subsection{Computing geodesics}
\label{sec:geodesics}

Let $\tilde\srvft_2 = (\srvft_2, \tilde\rotation,   \tilde\diffeotree, \tilde\permutetree)$. Since the metric is a weighted Euclidean norm of $\ltwo$ distances,   the geodesic $\tilde\treedeformationpath$ between $\srvft_1$ and $\srvft_2$ is a  straight line that connects $\srvft_1$ to $\tilde\srvft_2$, \ie:
\begin{equation}
	\tilde\treedeformationpath (\timeparam) = (1 - \timeparam) \srvft_1 + \timeparam \tilde\srvft_2, \timeparam \in [0, 1].
\end{equation}

\noi Finally, for visualization, we map $\treedeformationpath (\timeparam),  \timeparam \in [0, 1], $ back to the nonlinear space  of tree shapes $\preshapetrees$ using the inverse SRVF mapping, which has a closed analytical form; see~\cite{srivastava2011shape}.


\section{Statistics on the tree-shape space} 
\label{sec:statistics}

The ability to compute correspondences and geodesics between 3D tree-shaped objects enables a wide range of shape analysis tasks. In this section, we show how these fundamental tools can be used to compute the means and modes of variability of a collection of tree-shaped 3D objects $\{\proot_i, i =1,\dots, m\}$, \eg botanical trees, plant roots, or neuronal structures.  Let  $\{\srvft_i, i =1, 2, \dots, m\}$ be their corresponding SRVFT representations. To compute the mean tree $\meantree$, we first compute the mean SRVFT representation $\meansrvft $ and then map it back to the space of tree shapes. 

Mathematically, the mean  $\meansrvft $ can be regarded as the point in $\preshapesrvts$    that is as close as possible to all the tree shapes in $\{\srvft_i, i =1, \dots, m\}$. The closeness is defined with respect to the metric of Eqn.~\eqref{eq:invariant_metric}. In other words,
\begin{equation}
\begin{aligned}
	\meansrvft = \argmin_{\tiny{\begin{tabular}{c} 
				$\rotation_i \in \rotations$ \\
				$\diffeotree_i \in \diffeotrees$, $\permutetree_i \in \permutetreespace$ \\
				$i=1\dots m$
				
			\end{tabular}
	}}\sum_{i=1}^{m}d^2_{\preshapesrvts}(\srvft, \rotation(\srvft_i, \gamma_i, \permutetree_i)).
\end{aligned}
\label{eq:karcher_mean}
\end{equation}

\noi Solving  Eqn.~\eqref{eq:karcher_mean} involves finding  $\meansrvft $, also known as the Karcher mean, while simultaneously registering  every $\srvft_i$ on $\meansrvft $. This can be efficiently done via a gradient descent approach. That is:
\begin{enumerate}
\item Set $\meansrvft =  \srvft_1$. 
\item \label{step:for}For $i=1:m$
\begin{itemize}
	\item Optimally register  $\srvft_i$ onto $\meansrvft$ using Eqn.~\eqref{eq:registration}. 
	\item Let $(\tilde\rotation_i,   \tilde\diffeotree_i, \tilde\permutetree_i)$ be the solution to Eqn.~\eqref{eq:registration}.
\end{itemize}
\item \label{set:assign} Set $\meansrvft =  \frac{1}{m}\sum_{i=1}^{m}\tilde\rotation_i(\srvft_i,  \tilde\diffeotree_i, \tilde\permutetree_i)$.
\item Repeat steps~\ref{step:for} and~\ref{set:assign} until convergence.
\item Return $\meansrvft$ and $(\tilde\rotation_i,   \tilde\diffeotree_i, \tilde\permutetree_i)_{i=1}^{m}$. 
\end{enumerate}

\noi Finally, the mean tree $\meantree$ is obtained by mapping $\meansrvft$  back to the space $\preshapetrees$.  This mean shape characterizes the primary morphological properties of shapes in the collection $\{\proot_i, i =1, \dots, m\}$. In what follows, let $\srvft_i = \tilde\rotation_i(\srvft_i,  \tilde\diffeotree_i, \tilde\permutetree_i)$ be the SRVFT representation of $\proot_i$ but optimally registered onto  the mean $\meansrvft$. 

Let $\bm{v}_i = \srvft_i - \meantree$.  The covariance matrix is then defined as  $\displaystyle \covMatrix = \frac{1}{m-1}\sum_{i=1}^m\bm{v}_i\bm{v}_i^t.$ Its leading eigenvectors  $\Lambda_i$ define the principal directions of variations. The eigenvalue $\lambda_i$ that corresponds to the $i-$th mode of variation defines the variance along the $i-$th direction.

With this setup, each SRVFT sample $\srvft$ can be modeled as a  linear combination of the $k$ leading eigenvectors:
\begin{equation}
\srvft = \meansrvft + \sum_{i=1}^k a_i\sqrt{\lambda_i} \Lambda_i, a_i \in \real\ ,
\label{eq:eigenprojection}
\end{equation}
\noi Note that with this representation, we are fitting a Gaussian distribution to the input data $\{\srvft_i, i =1, \dots, m\}$.  In general, however, we can fit any arbitrary distribution from the parametric or non-parametric families. \hamid{Note , however, that a Gaussian distribution in the SRVFT space does not necessarily correspond to a Gaussian in the original space of trees (which is nonlinear). In fact, the mapping of distributions across the two spaces remains to be studied and this will be explored in future work.}

\hamid{Note also that the SRVFT space is flat only in the case when the shapes are not normalized for scale. However, the shape (quotient) space is not flat and the minimizer of Eqn.~\eqref{eq:karcher_mean} may not be unique but a set of distinct shapes. The algorithm  above tries to find a minimizer, which we then use to define covariance and PCA. Since the algorithm for computing the mean is a gradient-based search, it may not find the global minimizer. This is a limitation of the current approach and some recent papers have discussed the usage of stochastic searchers for reaching the global minimizer, but at the expense of a significantly higher computation time. As the objective function is bounded by zero from below, and the iterations are monotonic, the algorithm is guaranteed to converge, although it may converge to a local minimum.
}

\section{Generative model of 3D tree-shapes}
\label{sec:synthesis}


We characterize the shape variability of an input population of tree shapes by fitting to the shape population a multivariate Gaussian  of mean  $\meansrvft$ and a \guan{diagonal covariance matrix $\diagCovMatrix$} whose diagonal elements are the eigenvalues $\lambda_i$. A random 3D tree-shape can then be generated by randomly sampling from this multivariate Gaussian. That is, we first randomly sample  $k$ real numbers $a_1, \dots, a_k \sim {\cal N}(0,1)$. The SRVFT $\srvft $ of a random tree-shape $\proot$ is then given by Eqn.~\eqref{eq:eigenprojection}, and the random tree-shape $\proot$ is obtained by mapping $\srvft $ back to the space of 3D tree-shapes.   In this paper, we only consider the $k$-leading eigenvalues such that $k \le m$ and $\frac{\sum_{i=1}^{k}  \lambda_i}{\sum_{i=1}^{m} \lambda_i}>0.99$.  To ensure that the synthesized tree-shapes are plausible, one can restrict $a_i$  to be within a certain range. For example, by setting  $a_i \in [-1, 1]$, the generated random trees are restricted to be within one standard deviation around the mean tree.

\begin{figure*}[!ht]
\center
\includegraphics[width=\textwidth, trim={0 2cm  0 5cm},clip]{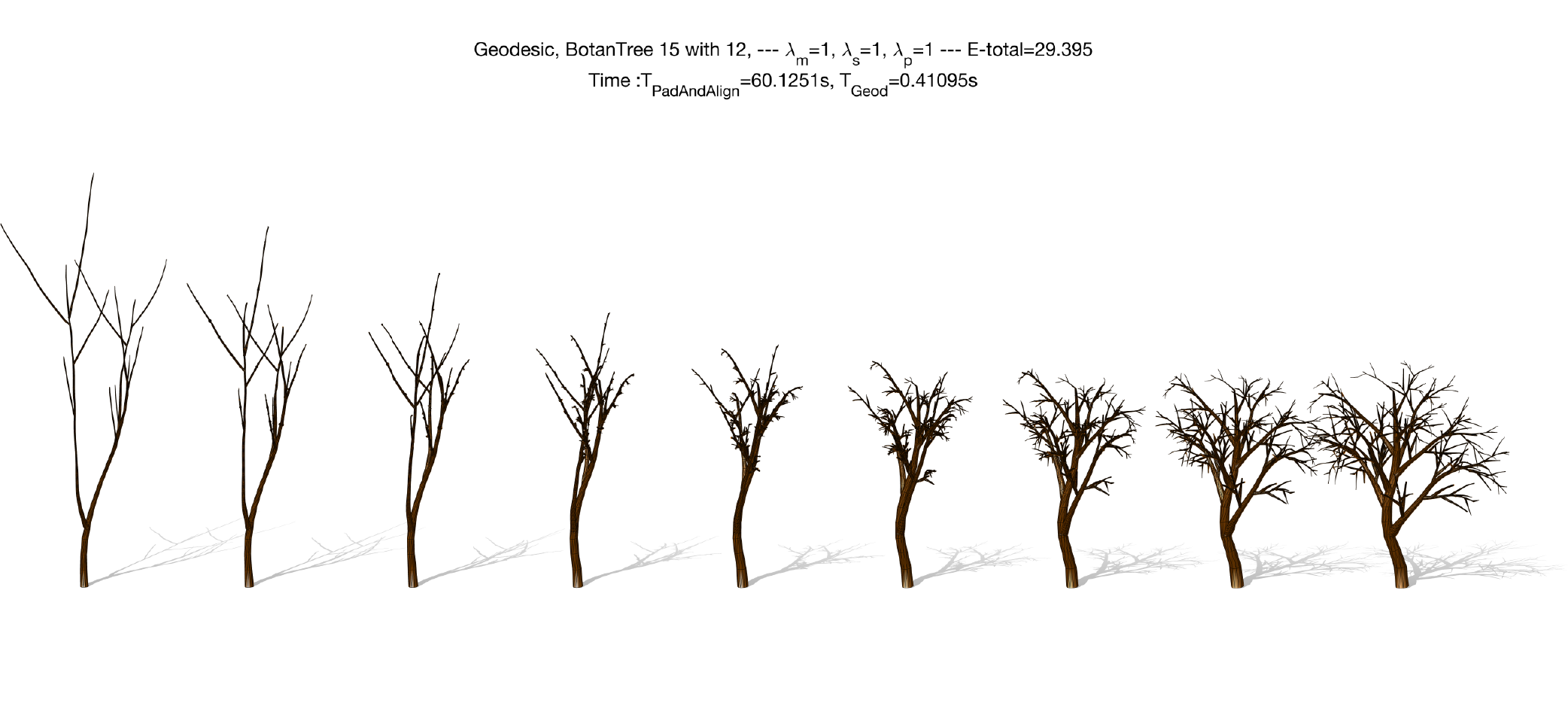}\\
\small{(a) Geodesic length $d_s = 29.3$. \guan{Computation time: $60.1$s for computing the correspondences and  $0.4$s for the geodesics.}}
\includegraphics[width=\textwidth, trim={0 11cm  0 10cm},clip]{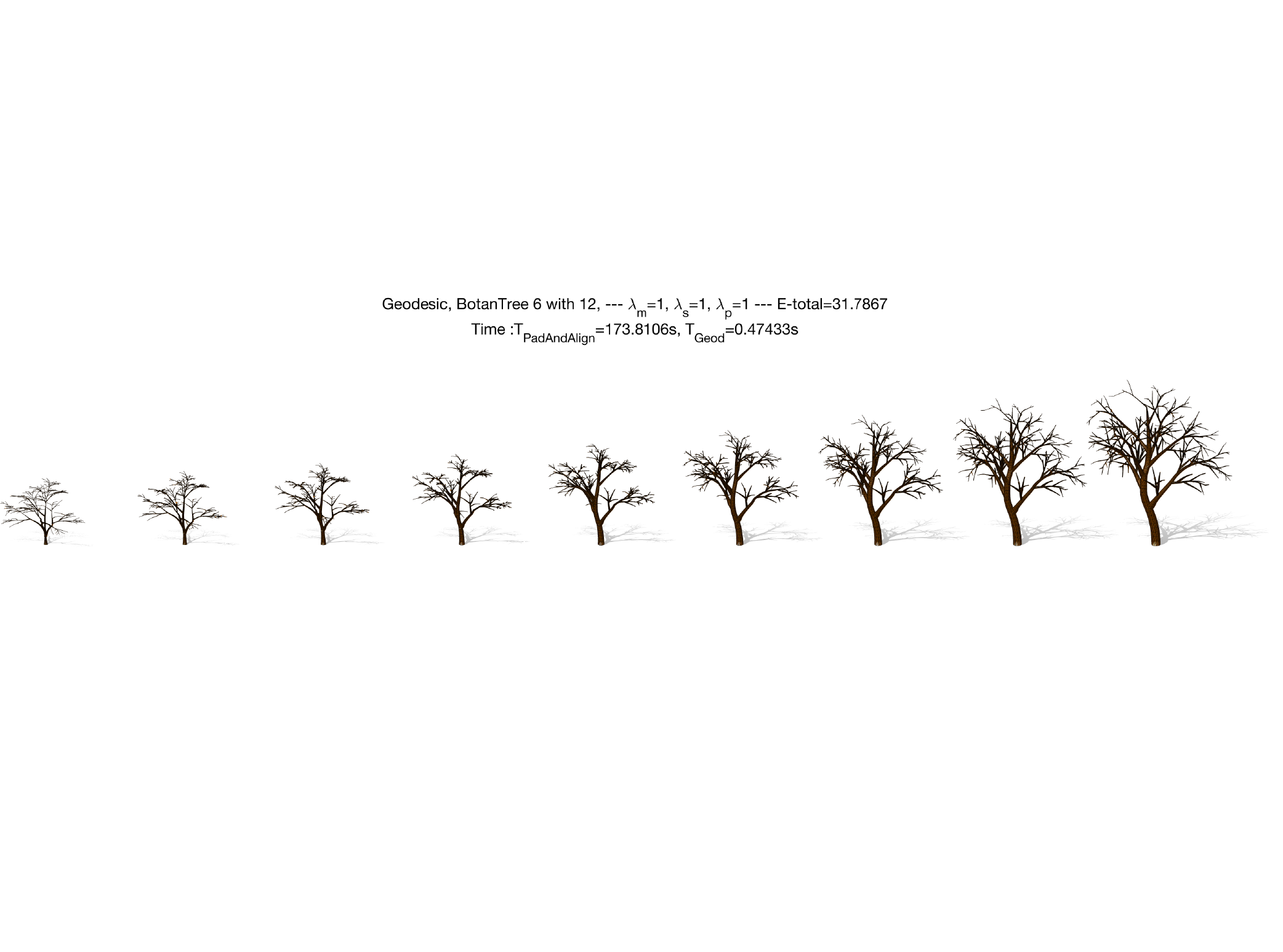}\\
\small{(b) Geodesic length $d_s = 31.7$. \guan{Computation time: $173.8$s for computing the correspondences and  $0.47$s for the geodesics.}}\\
\includegraphics[width=\textwidth, trim={0 10cm  0 10cm},clip]{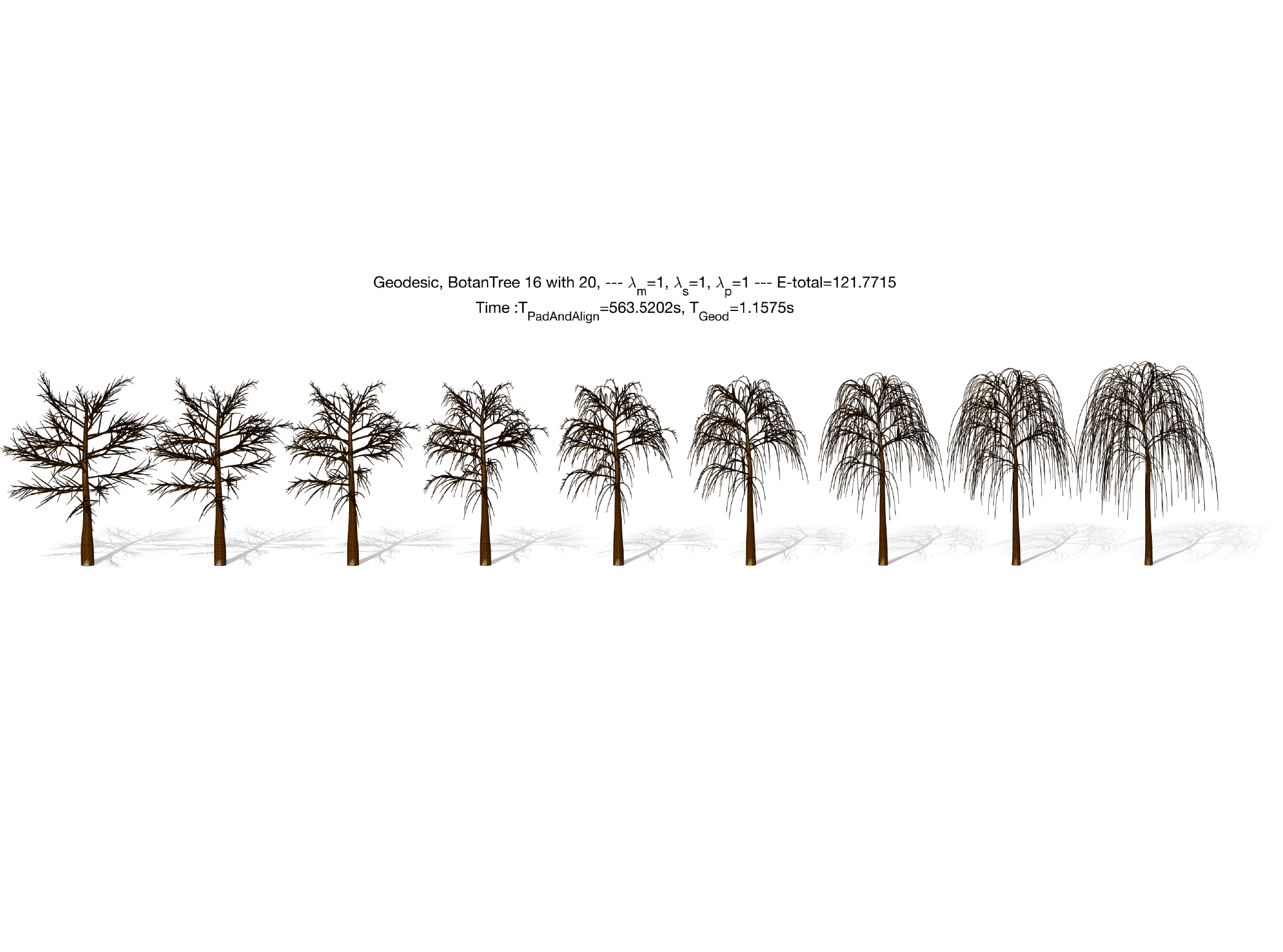}\\
\small{(c) Geodesic length $d_s= 121.8$. \guan{Computation time: $563.5$s for computing the correspondences and  $1.1$s for the geodesics.}}
\caption{Geodesic deformations between the most left and the most right 3D botanical trees in each row. In this experiment, we use $\lambda_m=\lambda_s=\lambda_p=1.0$.  \guanwangRRR{We refer the reader to Fig.~14 in the Supplementary Material, which shows the color-coded subtree-wise correspondences.} The Supplementary Material also includes more examples.}
\label{fig:geodesic_trees}
\end{figure*}

\section{Results and discussion}
\label{sec:results}
In this section, we demonstrate the performance of the proposed framework in finding correspondences and computing geodesics between pairs of tree-like 3D objects (Section~\ref{sec:results_corrs_geodesics}), the analysis of the symmetry  and symmetrization of such shapes (Section~\ref{sec:results_symmetry}), computing summary statistics such as means and modes of variation (Section~\ref{sec:results_statistics}), and finally synthesizing novel tree-shaped 3D objects 
(Section~\ref{sec:results_synthesis}). \hamid{We use two types of datasets, namely 3D botanical trees obtained from \href{https://www.xfrog.net/}{xfrog.net}\footnote{\url{https://www.xfrog.net/xfrog-software/}}, and 3D neuronal structures from~\cite{ascoli2007neuromorpho,suo2012protocadherin,chen2014morphological}.
}
In both cases, the models have complex branching structures composed of multiple layers, unlike Duncan \etal~\cite{duncan2018statistical}, which use simple trees of level one. The models used in this section also exhibit large geometric and topological variability, making it challenging to find one-to-one correspondences between such tree-like 3D objects. \hamid{In all our experiments, we assumed that the main branch (trunk) of the botanical trees is the one that is less bent   compared to the other branches. In the case of neuronal structures, the longest branch is selected as the main one. 
}Also, we omit the branch  thickness when computing correspondences. It is, however, included when computing geodesics and summary statistics.

The Supplementary Material includes more results as well as videos of all the examples shown in this paper. 



\subsection{Correspondence and geodesics}
\label{sec:results_corrs_geodesics}

\begin{figure}[!ht]
\center
\includegraphics[width=0.5\textwidth]{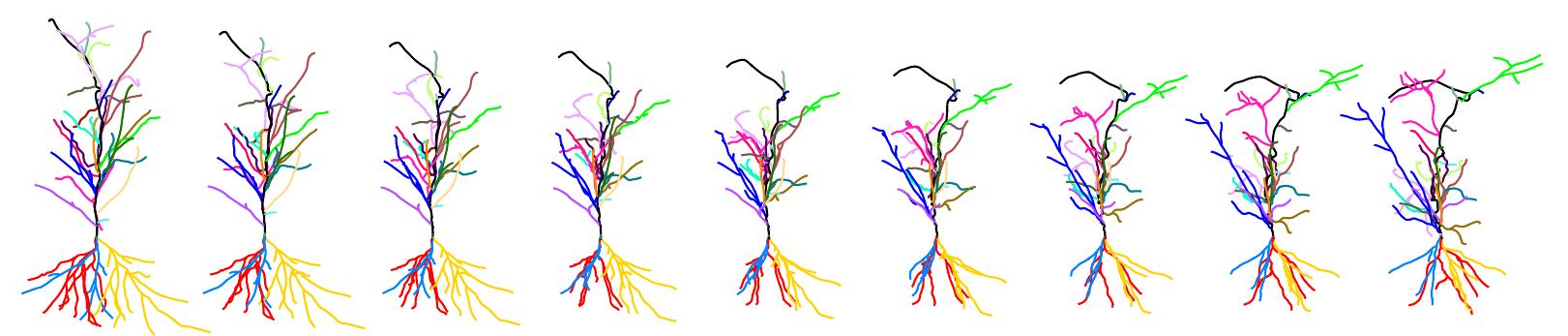}\\
\small{(a) Geodesic length $d_s = 156.7$. \guan{Computation time: $22.4$s for computing the correspondences and $0.21$s for the geodesics.}} \\
\includegraphics[width=0.5\textwidth]{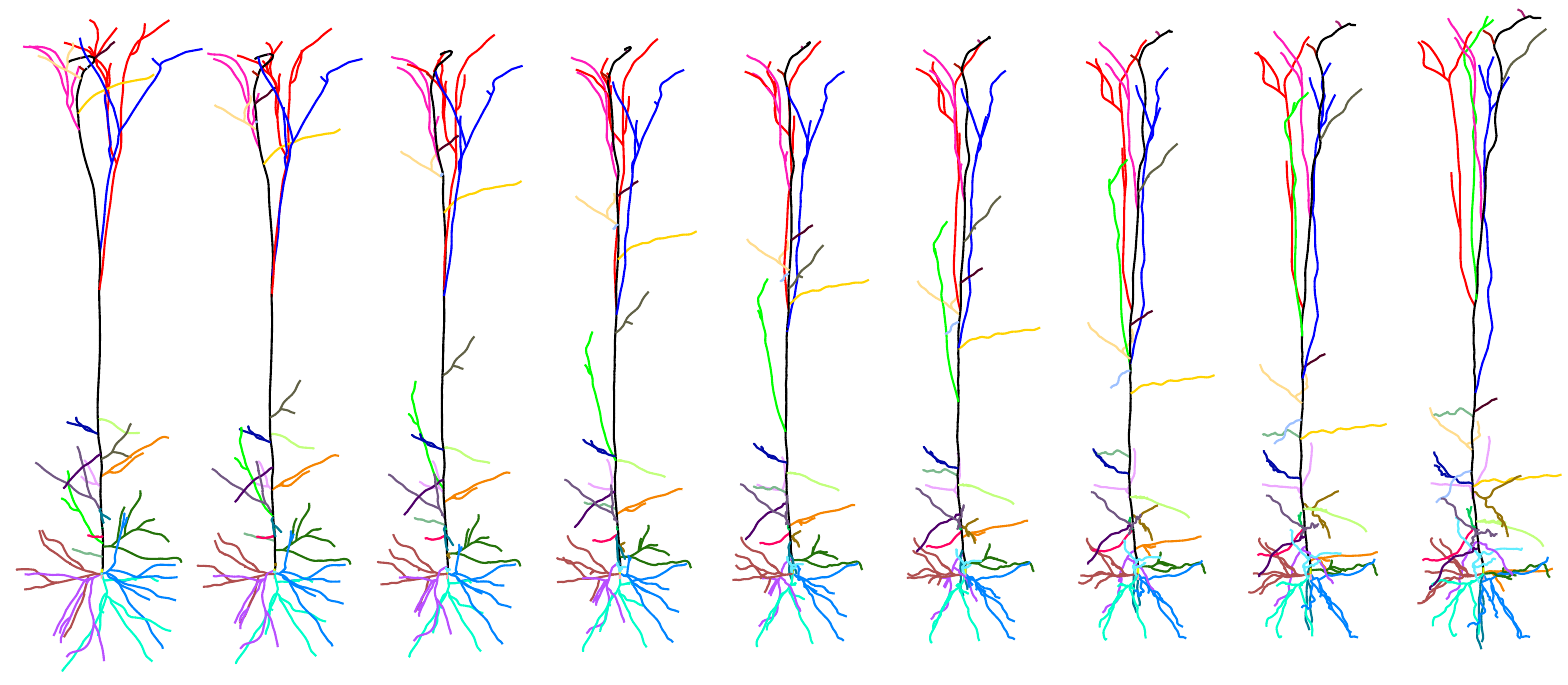}\\
\small{(b) Geodesic length $d_s = 165.2$. \guan{Computation time:  $29.6$s for computing the correspondences and $0.25$s for the geodesics.}}\\
\includegraphics[width=0.5\textwidth]{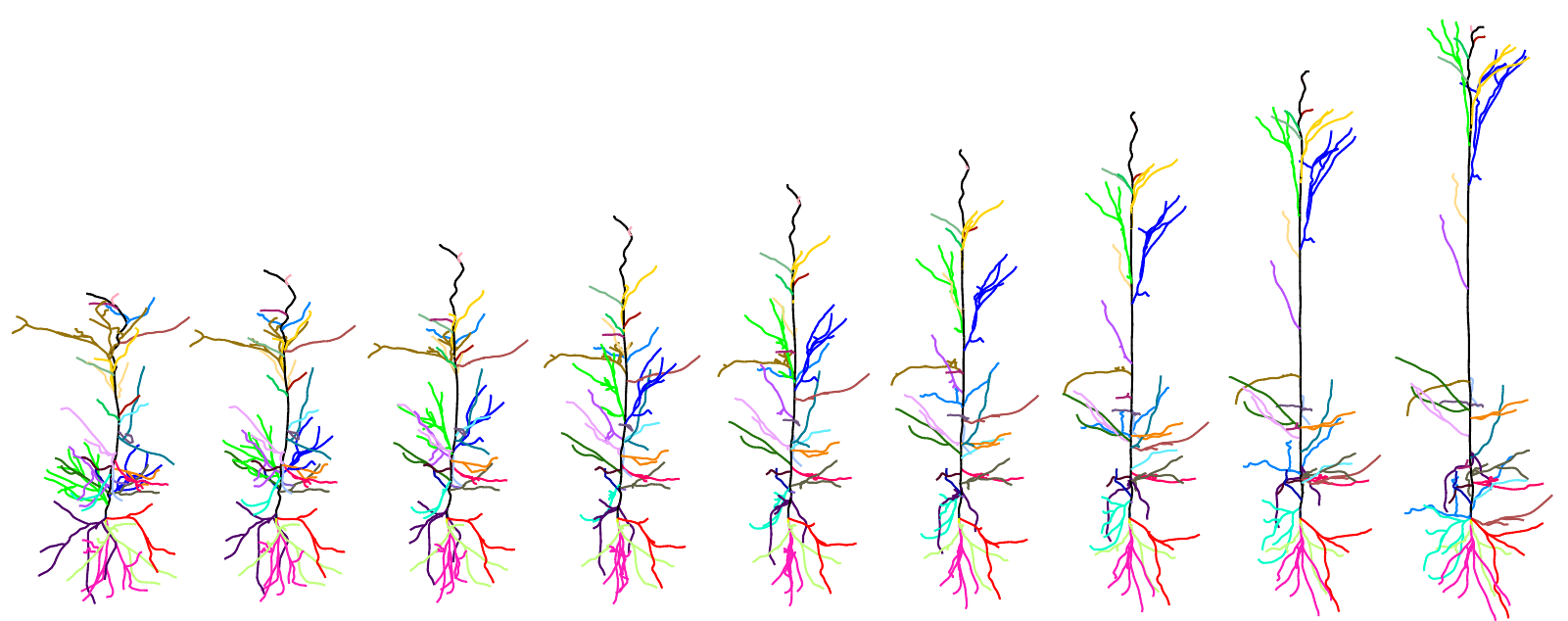}\\
\small{(c) Geodesic length $d_s = 181.8$. \guan{Computation time: $30.2$s for computing the correspondences and  $0.22$s for the geodesics.}}
\caption{Geodesic deformations between the most left and the most right 3D neuronal trees in each row. In this experiment, we use $\lambda_m=0.2, \lambda_s= 1.0, \lambda_p=0.2$. \guanwangRRR{In each row, we show the geodesic between the most left and the most right trees as well as the color-coded subtree-wise correspondences. }}
\label{fig:geodesic_neurons}
\end{figure}

\begin{figure}[t]
\center
\includegraphics[width=.5\textwidth, trim={0 8cm  3cm 8cm},clip]{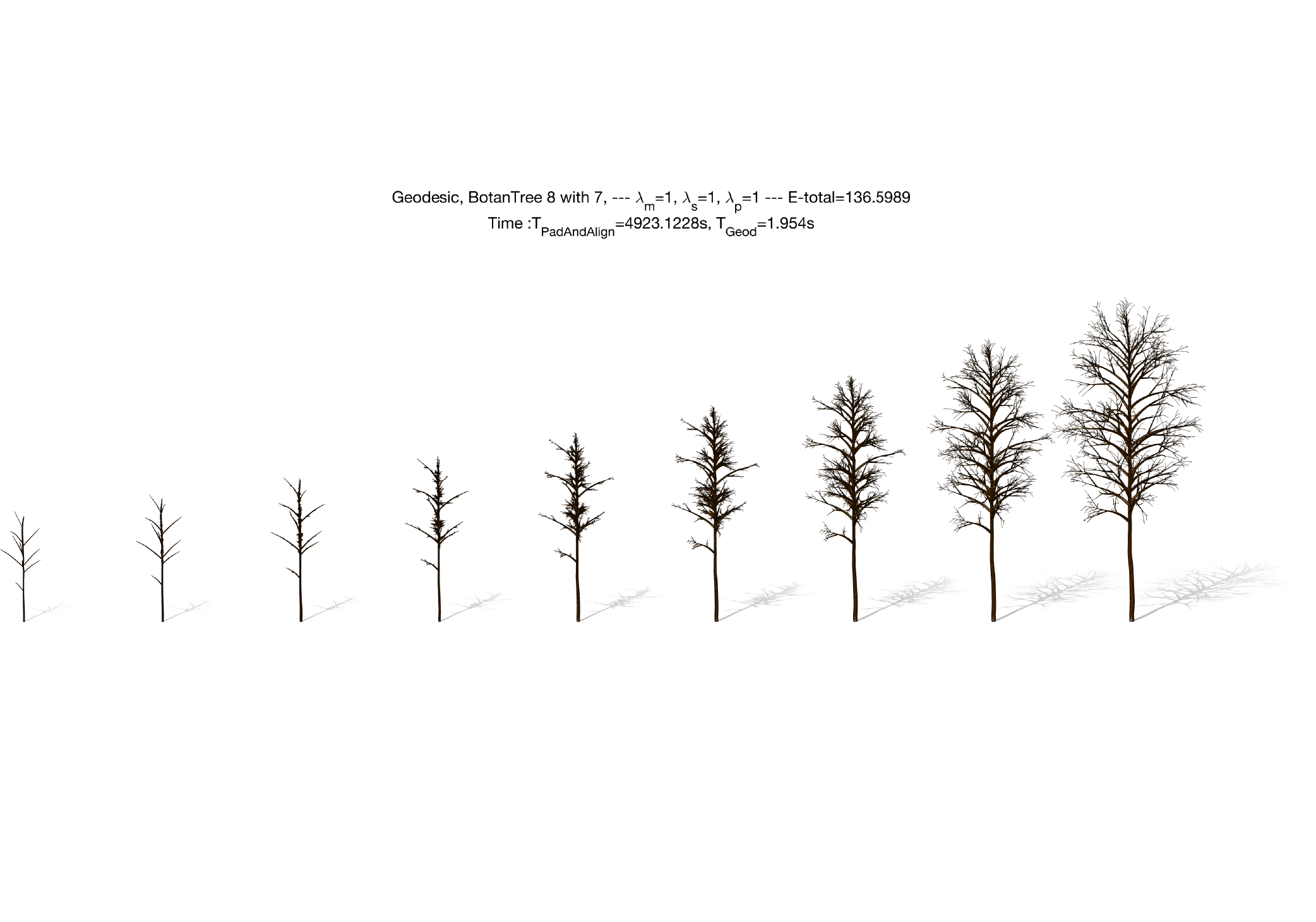}
\caption{A geodesic path between a small tree and a large botanical trees. \guanwangRRR{We refer the reader to Fig.~15 in the Supplementary Material, which shows the color-coded subtree-wise correspondences}}
\label{fig:growth}
\end{figure}

Figs.~\ref{fig:geodesic_trees} and~\ref{fig:geodesic_neurons} show examples of geodesic paths between 3D botanical trees and 3D neuronal structures, respectively. For each example, we show the geodesic between the leftmost (source) and the rightmost (target) trees. The botanical trees exhibit very complex 3D geometric and topological structures. Nevertheless, the approach can find one-to-one correspondences and a natural deformation between the source and target trees. The neuronal trees of Fig.~\ref{fig:geodesic_neurons} have a relatively simple structure since they are composed of a main branch and a number of lateral branches. They, however, have a very complex geometry since their branches significantly bend and stretch. Finally, Fig.~\ref{fig:growth} shows the geodesic, computed with the proposed approach, between a small tree and a large tree. 


\begin{figure*}[!t]
\center

\includegraphics[width=0.9\textwidth]{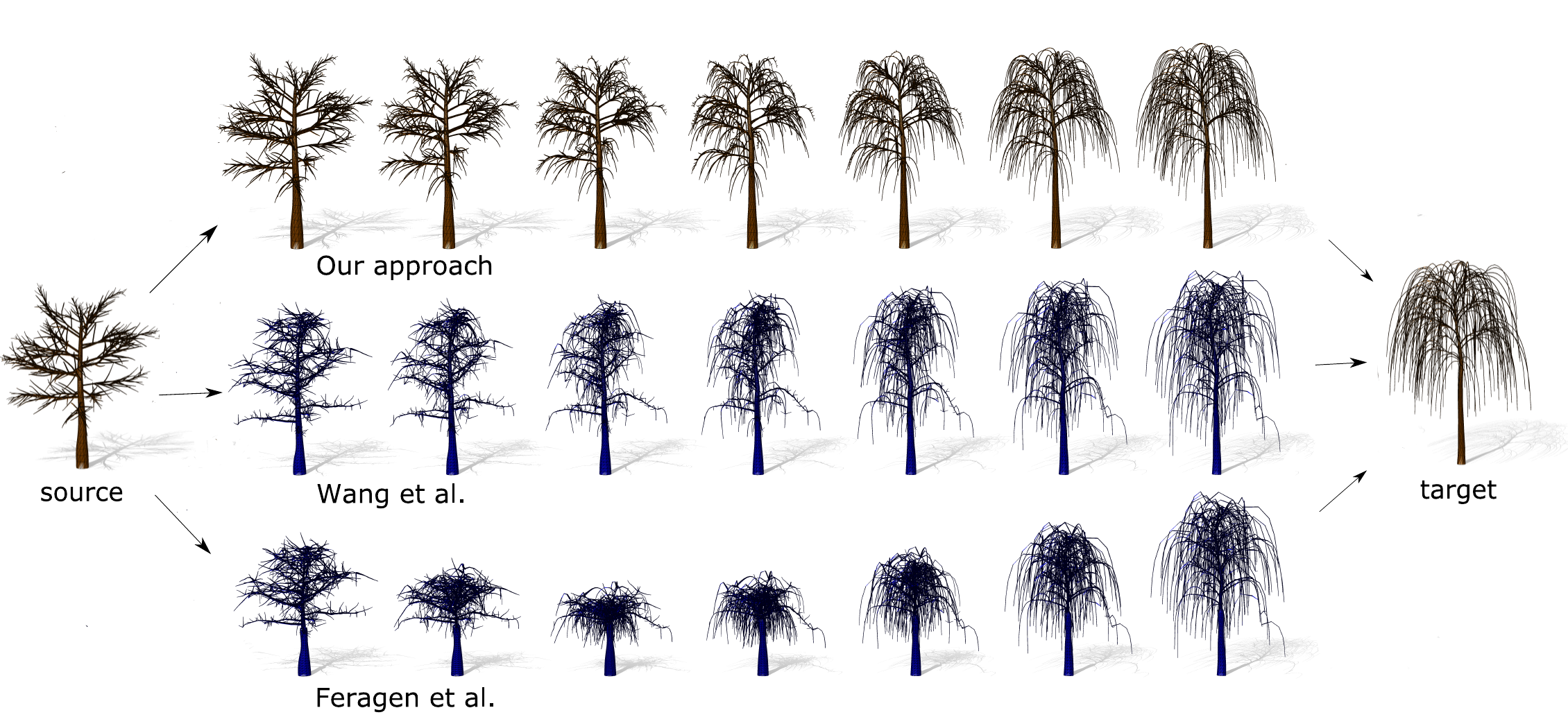} \\
\caption{Comparison of the quality of the geodesics obtained using our approach, the approach of Feragen \etal~\cite{feragen2013toward}, and the approach of \guanwangRRR{Wang} \etal~\cite{wang2018the}. The Supplementary Material includes more comparison examples.}
\label{fig:comparison_3botanGeods}
\end{figure*}

Fig.~\ref{fig:comparison_3botanGeods} compares the results of our approach with the approaches of Feragen \etal~\cite{feragen2013toward} and Wang \etal~\cite{wang2018the}. Conceptually, the approach of Feragen \etal~\cite{feragen2013toward}, which automatically finds correspondences and geodesics, does not consider the thickness of the branches that compose the tree shapes. It is also computationally costly, especially for complex trees. Wang \etal~\cite{wang2018the} extended the approach by pre-computing the correspondences in an ad-hoc manner. In both methods, the nodes represent bifurcation points, and the edges represent the branch segments between the bifurcation points. Topological changes are modeled using edge collapse and node split operations. As such, the intermediate shapes along a geodesic exhibit significant shrinkage \hamid{since the edge collapse operation reduces the size of the collapsed edge to zero, which shrinks the branch and subsequently the entire tree}; see the bottom two rows in Fig.~\ref{fig:comparison_3botanGeods}. \hamid{This will affect downstream applications such as average tree computation and tree synthesis from a collection of 3D tree models.} The approach proposed in this paper solves this problem by modeling topological changes as sliding branches. It also uses a full elastic metric, which explicitly quantifies the bending and stretching of the branches. Subsequently, the geodesics look more natural compared to the state-of-the-art; see the top row in Fig.~\ref{fig:comparison_3botanGeods}.

\vspace{6pt}
\noi\textbf{Computation time.}  
Our framework is implemented using Matlab and runs entirely on CPU, configured with 2.4GHz Intel Core i5, 8GB RAM. The most time-consuming part is the correspondence computation process, which takes $60.1$ to $4923.1$ seconds for the botanical trees and $22.4$ to $30.2$ seconds for the neuronal trees. The subsequent geodesic computation takes significantly less time ($0.4$ to $1.9$ seconds for the botanical \guan{trees} and $0.2$ seconds for the neuronal trees). \hamid{We refer the reader to the captions of the figures for the computation time of each example.} 


\subsection{Ablation study}
\label{sec:ablation}
\begin{figure}[t]
\center
\includegraphics[width=0.45\textwidth, trim={0 0cm  0 0.8cm},clip]{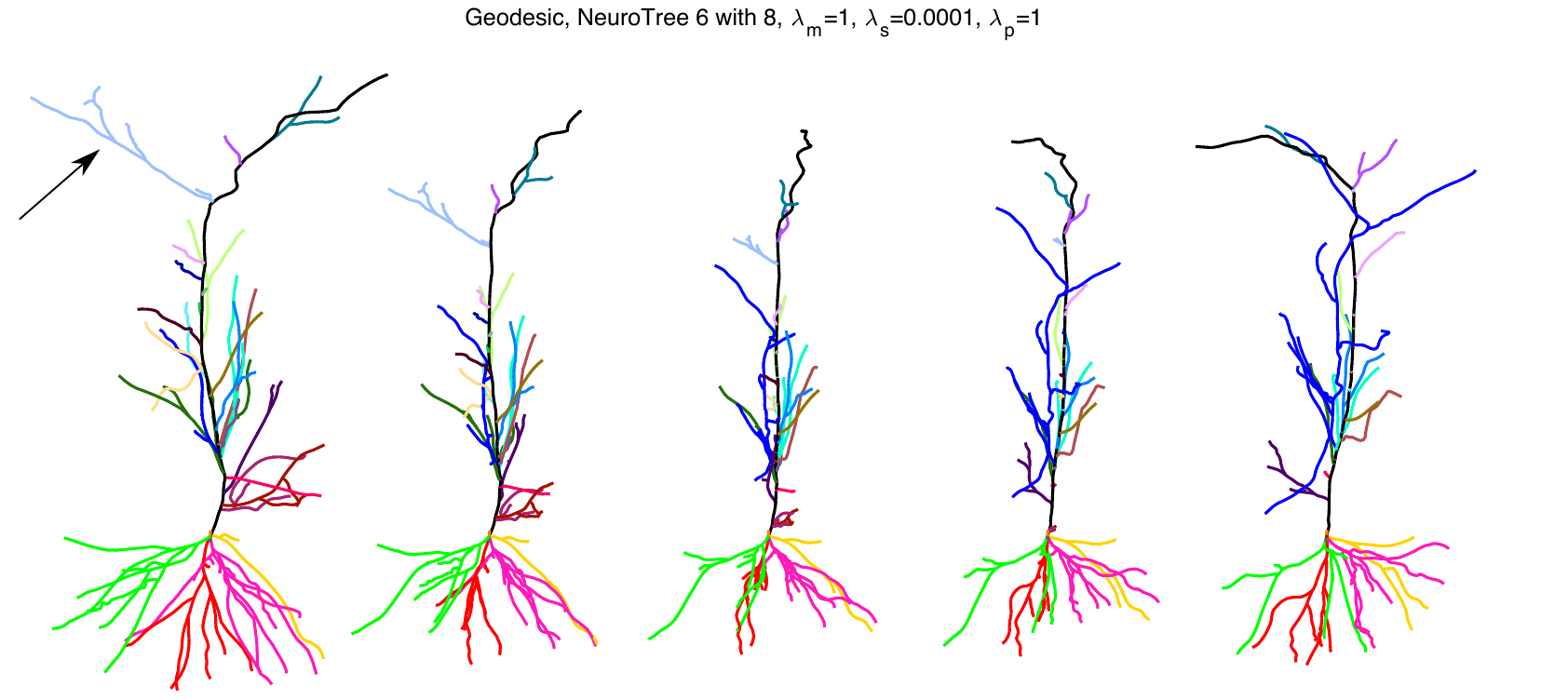} \\
\small{(a) $\lambda_m=1.0, \lambda_s=10^{-4}, \lambda_p=1.0$.}\\
\includegraphics[width=0.45\textwidth, trim={0 0cm  0 0.8cm},clip]{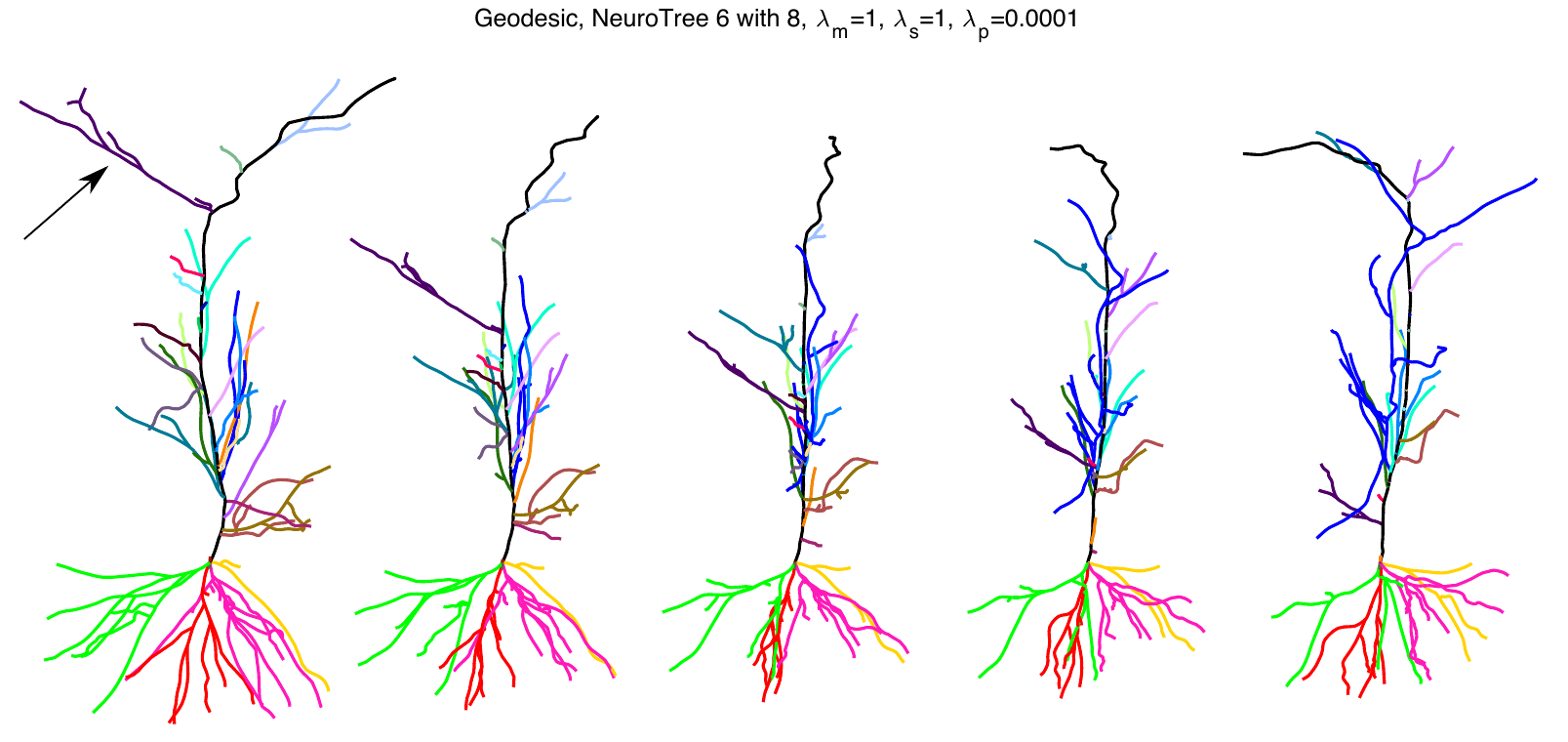} \\
\small{(b) $\lambda_m=1.0,\lambda_s=1.0, \lambda_p=10^{-4}$.}

\caption{The influence of weights $\lambda_m,\lambda_s$, and $\lambda_p$ of Eqn.~\ref{eq:init_distanceTwoTrees} on the geodesics computed with our approach. \guanwangRRR{The colors indicate subtree-wise correspondences.}}
\label{fig:weights_study}
\end{figure}

\vspace{6pt}
\noi\textbf{Effect of the weights \guan{$\lambda_m,\lambda_s$, and $\lambda_p$.} } These three parameters control the importance of each of the three terms in Eqn.~\eqref{eq:invariant_metric}. In the experiments of Fig.~\ref{fig:geodesic_trees}, these parameters were manually set to $\lambda_m=\lambda_s=\lambda_p=1.0$ while in Fig.~\ref{fig:geodesic_neurons}, they were set to $\lambda_m=0.2, \lambda_s= 1.0, \lambda_p=0.2$. Fig. ~\ref{fig:weights_study} shows how the correspondences and geodesics vary while varying these parameters. For the first example in Fig.~\ref{fig:weights_study} with $\lambda_m=1.0, \lambda_s=10^{-4}, \lambda_p=1.0$, the approach favors creating new virtual subtrees (see the subtree indicated by an arrow) rather than sliding the existing ones. We observe the opposite behavior in the  second example when we set $\lambda_m=1.0, \lambda_s=1.0, \lambda_p=10^{-4}$. This is reasonable since in the first example, the last term in Eqn.~\ref{eq:init_distanceTwoTrees}, which measures the distance between bifurcation points, is significantly higher than the second term, which measures shape dissimilarity between two subtrees.

\begin{figure}[!ht]
\center
\includegraphics[width=0.5\textwidth]{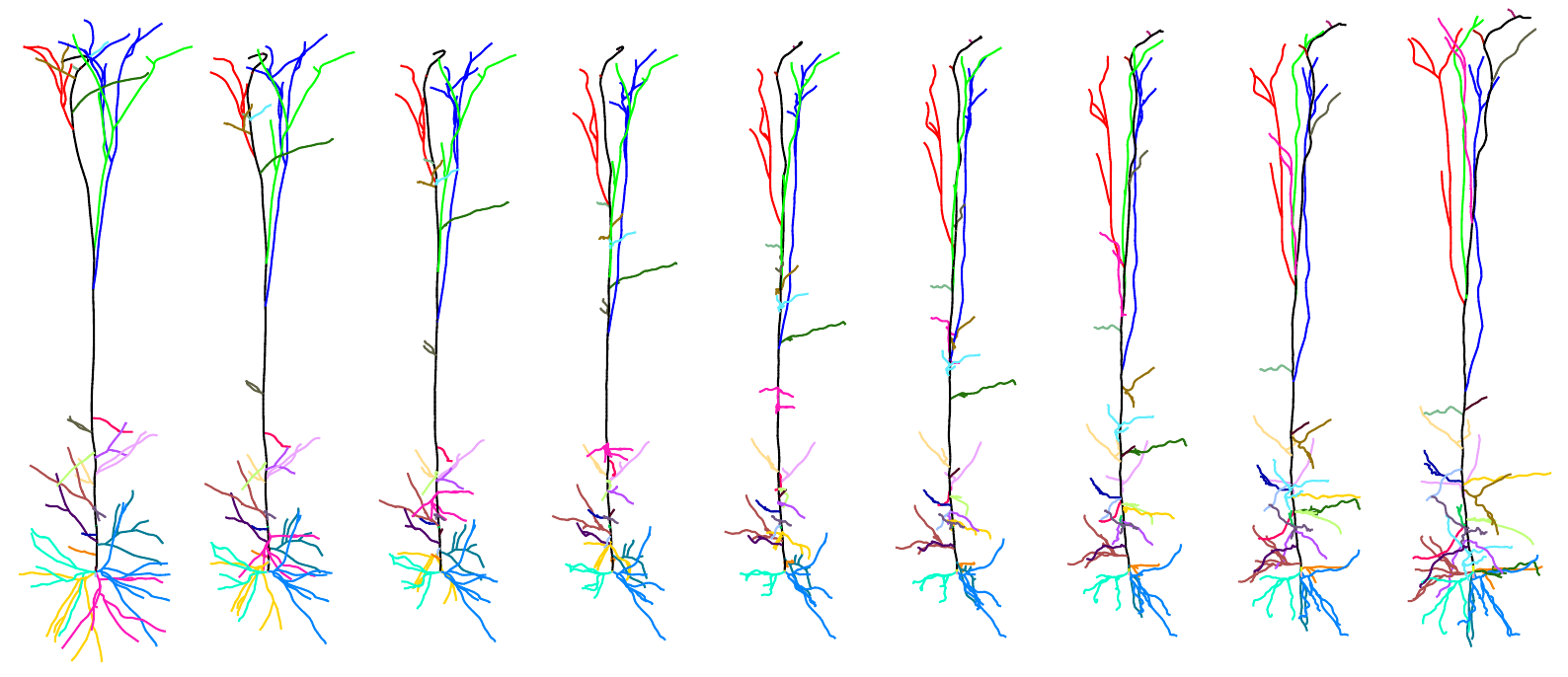}\\
\small{(a) Geodesic \textbf{without} shape matching.}\\ 
\includegraphics[width=0.5\textwidth]{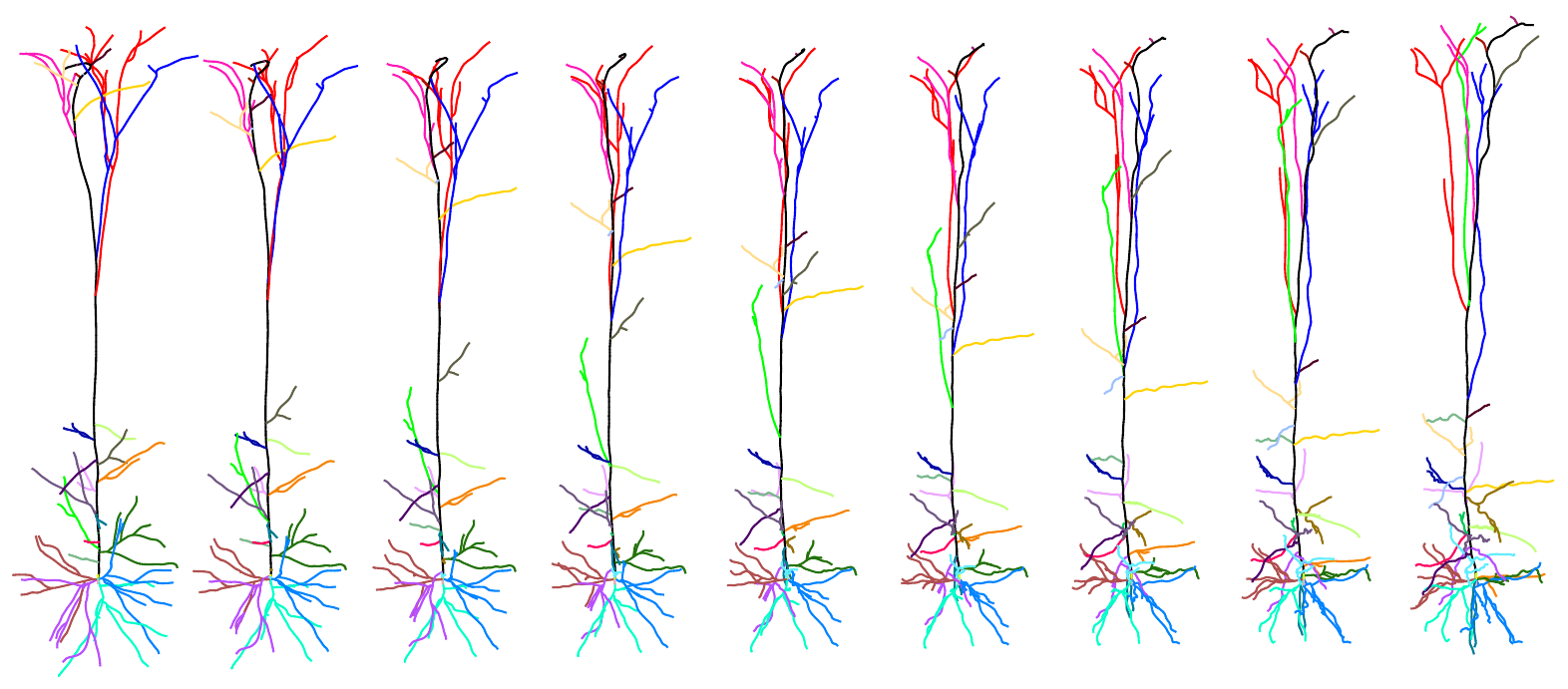}\\
\small{(b) Geodesic \textbf{with} shape matching.} 
\caption{Ablation study on neuronal tree geodesics. (a) Geodesic generated without graph matching, (b) geodesic generated with graph matching. \guanwangRRR{In each row, we show the geodesic between the most left and the most right trees as well as the color-coded subtree-wise correspondences. }}
\label{fig:abla_geodesic_neurons}
\end{figure}

\begin{figure}[t]
\center
\includegraphics[width=0.5\textwidth, trim={0 8cm  0 8cm},clip]{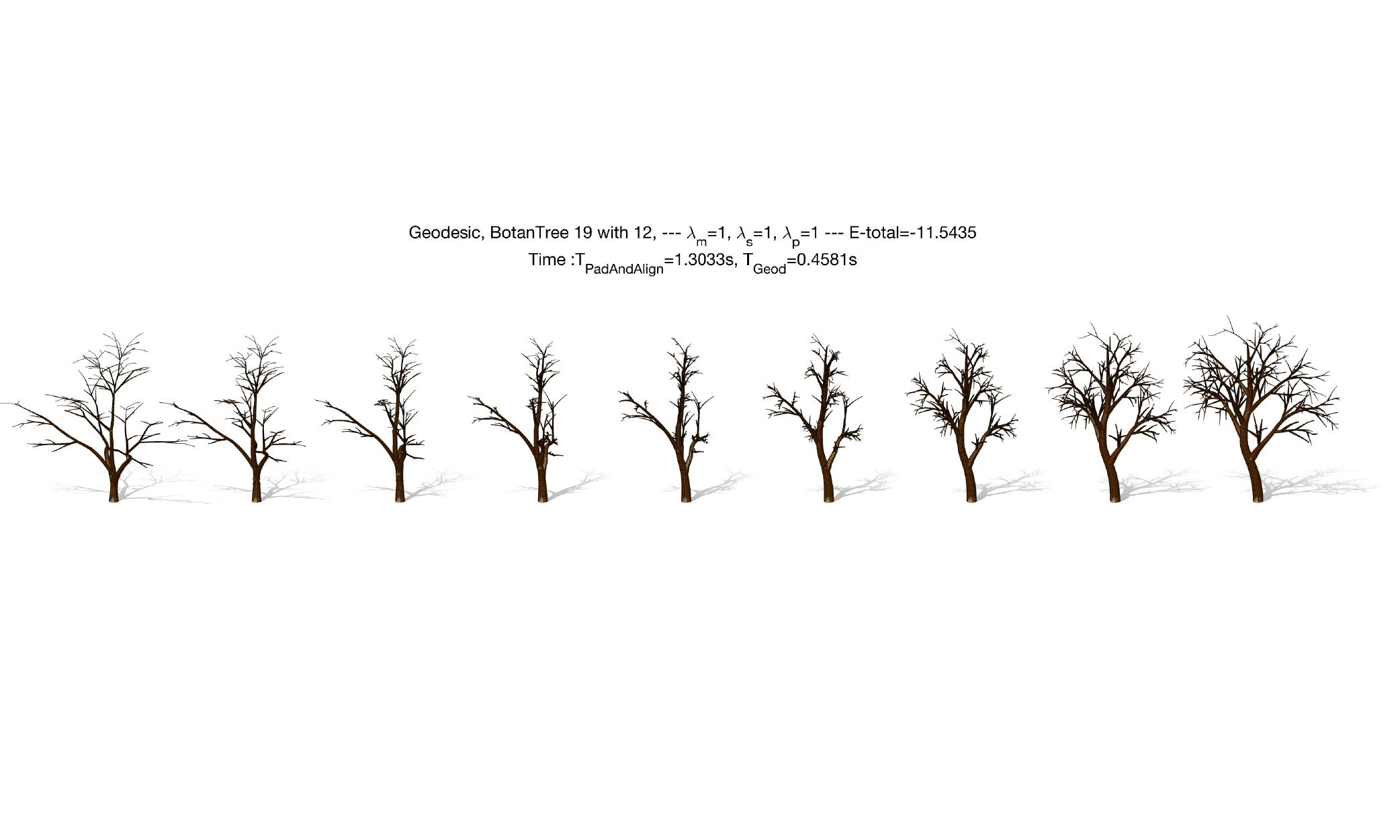}\\
\small{(a) Geodesic \textbf{without} shape matching.}\\
\includegraphics[width=0.5\textwidth, trim={0 8.5cm  0cm 8cm},clip]{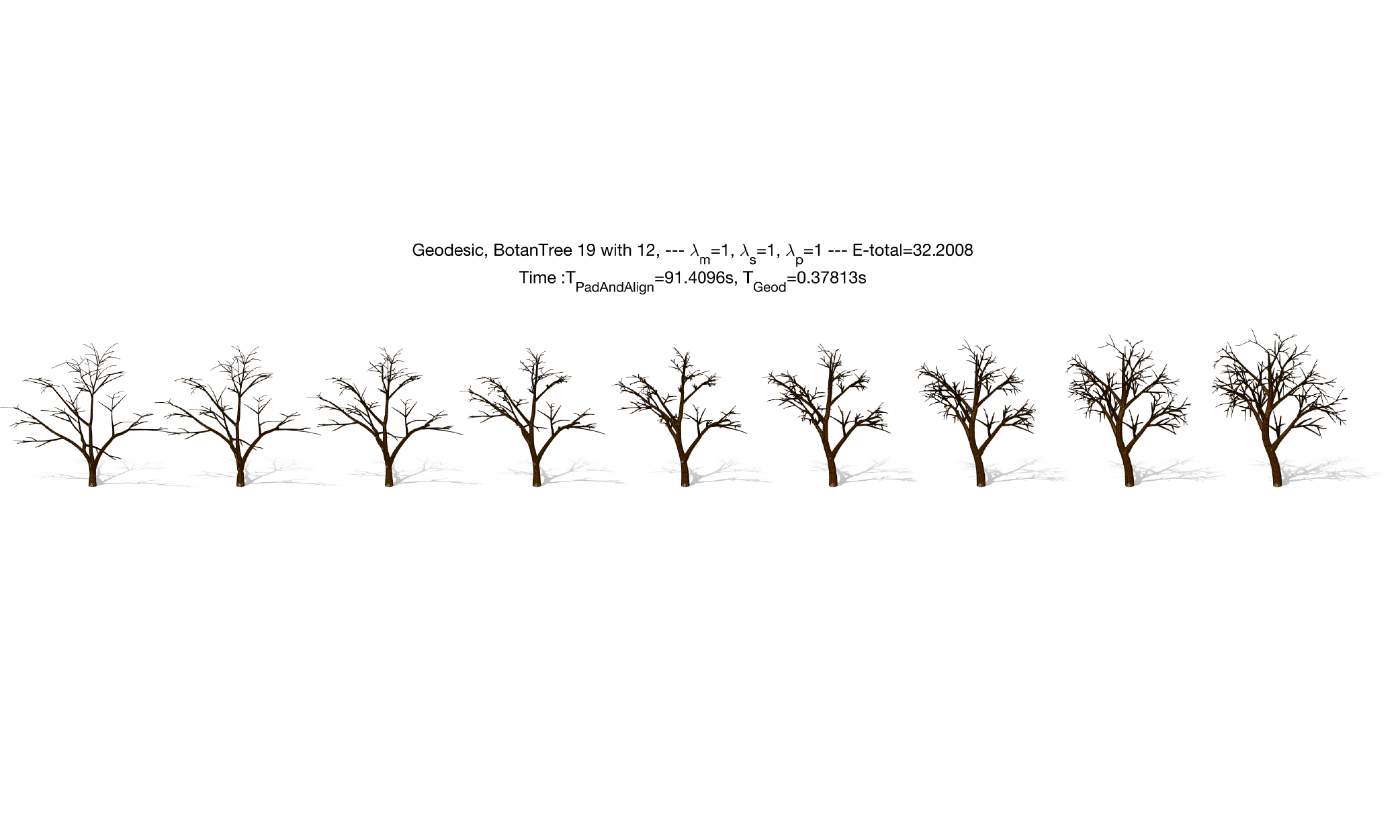}\\
\small{(b) Geodesic \textbf{with} shape matching.}
\caption{Ablation study on botanical tree geodesics. (a) Geodesic generated without graph matching, (b) geodesic generated with graph matching. \guanwangRRR{We refer the reader to Fig.~16 in the Supplementary Material, which shows the color-coded subtree-wise correspondences.}}
\label{fig:abla_geodesic_botanTrees}
\end{figure}

\vspace{6pt}
\noi\textbf{Effect of the matching step.} A core contribution of our framework is the shape matching between complex tree pairs. In this section, we investigate the effect of this step on the quality of the computed geodesics.  Figs.~\ref{fig:abla_geodesic_neurons} and~\ref{fig:abla_geodesic_botanTrees} compare the geodesics obtained using our approach with and without the shape matching component. We can see that the geodesics generated by the framework with shape matching look more natural than those without shape matching. These results demonstrate the contribution of shape matching to the quality of the geodesic generation. The framework enables the least morphing cost between tree pairs by shape matching, thereby resulting in more natural and smoother morphing sequences.


\subsection{Reflection symmetry analysis and symmetrization}
\label{sec:results_symmetry}

Symmetry is an important feature of artificial and biological objects and its analysis can benefit many applications.  Existing symmetry analysis techniques are mainly based on feature detection and matching; see, for example, the survey by Mitra \etal~\cite{mitra2013symmetry}. The proposed framework, which  provides a proper metric and a mechanism for computing geodesics, can be used to  analyze the reflection symmetry and to symmetrize tree-shaped 3D objects without extracting and computing shape descriptors. 

\begin{figure}[t]
\center
\includegraphics[width=0.5\textwidth, trim={0 8cm  0 8cm},clip]{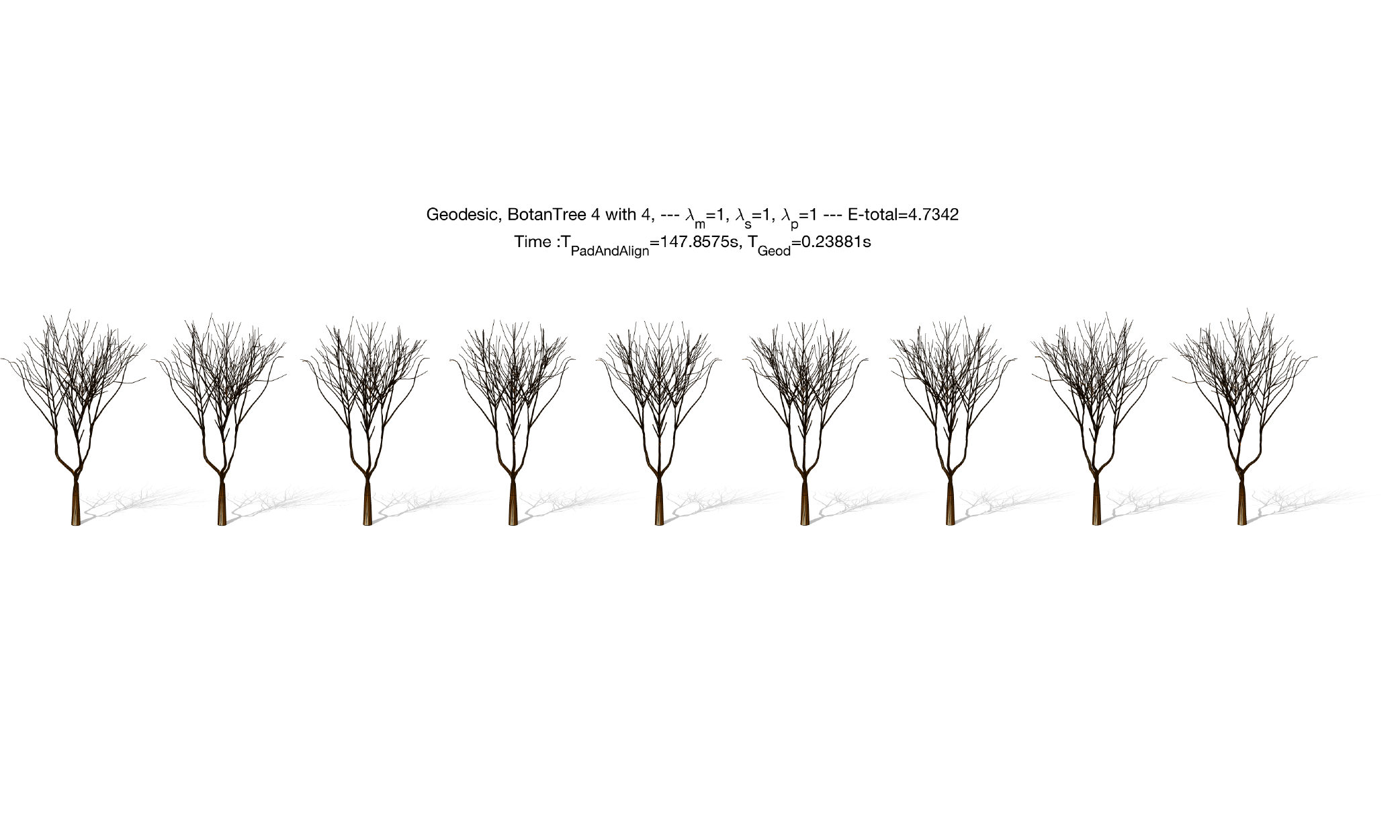}\\
\small{(a) Geodesic length $d_s = 4.7$.} \\
\includegraphics[width=0.5\textwidth, trim={0 8.5cm  0 8cm},clip]{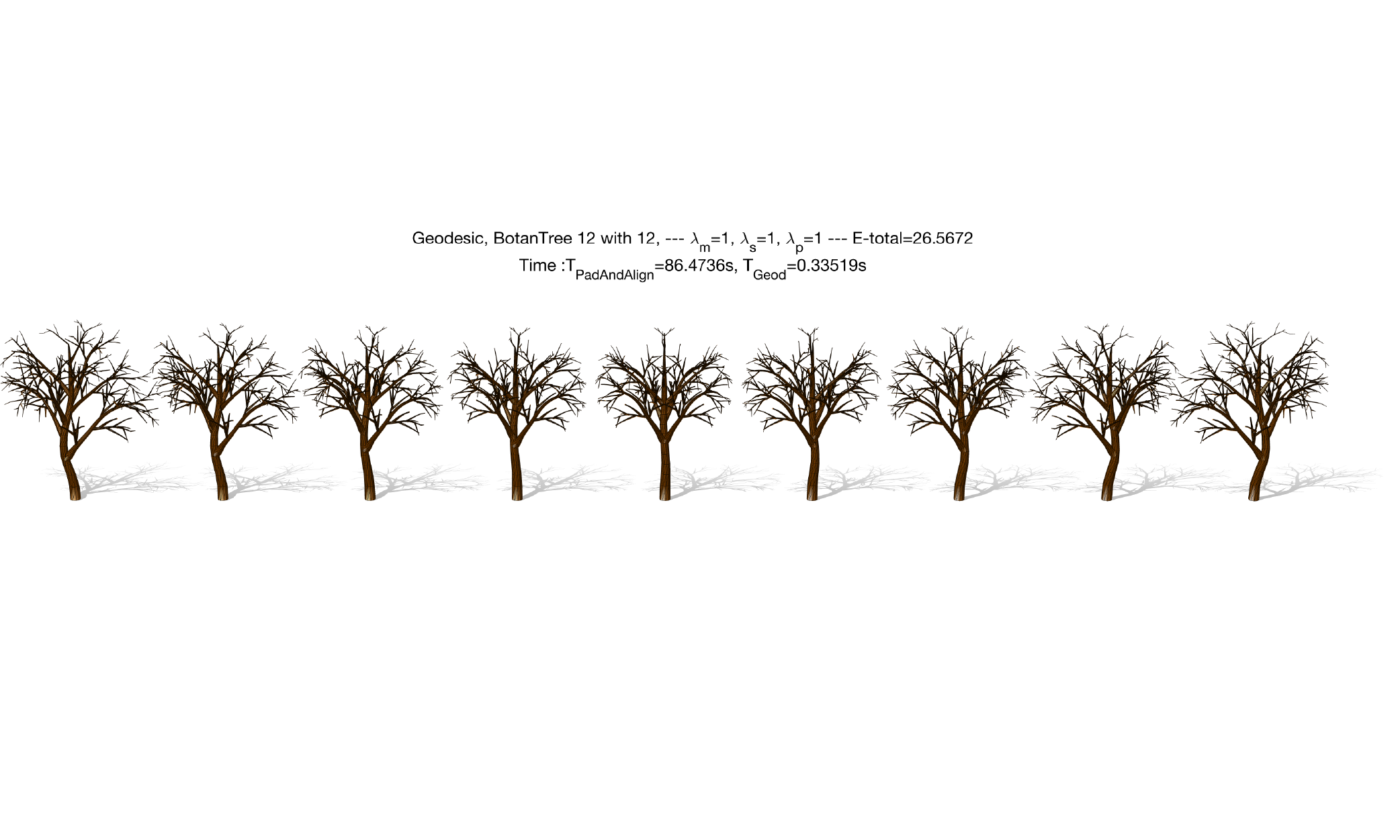}\\
\small{(b) Geodesic length $d_s = 26.6$.} 
\caption{Analysis of the symmetry and symmetrization of 3D botanical trees. The Supplementary Material includes more examples. \guanwangRRR{We refer the reader to Fig.~17 in the Supplementary Material, which shows the color-coded subtree-wise correspondences.}}
\label{fig:symmetry_trees}
\end{figure}

To analyze the level of asymmetry of a given tree-shaped object  $\proot$ using our  framework,  we first obtain its reflection across an arbitrary plane $\Delta$. Let $v \in  \real$ be a  normal vector to this plane. Then, the  reflection of $\proot$ with respect to $\Delta$ is given by
\begin{equation}
\tilde{\proot} =  \left(I - 2{vv^{\top} \over v^{\top} v}\right) \proot.
\end{equation}

\noi Here, $I$ is the $3\times3$ identity matrix and $v^{\top}$ refers to the transpose of the vector $v$. Next, we use the approach described in this paper to compute a geodesic path, $\treedeformationpath$, between $\proot$ and $\tilde{\proot}$.  $\treedeformationpath$ provides valuable information about the symmetry of $\proot$. First, its length gives a proper measure of the asymmetry of $\treedeformationpath$. Second, the halfway point along this geodesic, \ie $\treedeformationpath(0.5)$, is symmetric. Moreover, amongst all symmetric trees, $\treedeformationpath(0.5)$ is the nearest to $\proot$. Thus, the process of computing a geodesic between a tree-shaped object and its reflection is equivalent to symmetrizing the object.

\begin{figure}[t]
\center
\includegraphics[width=0.5\textwidth]{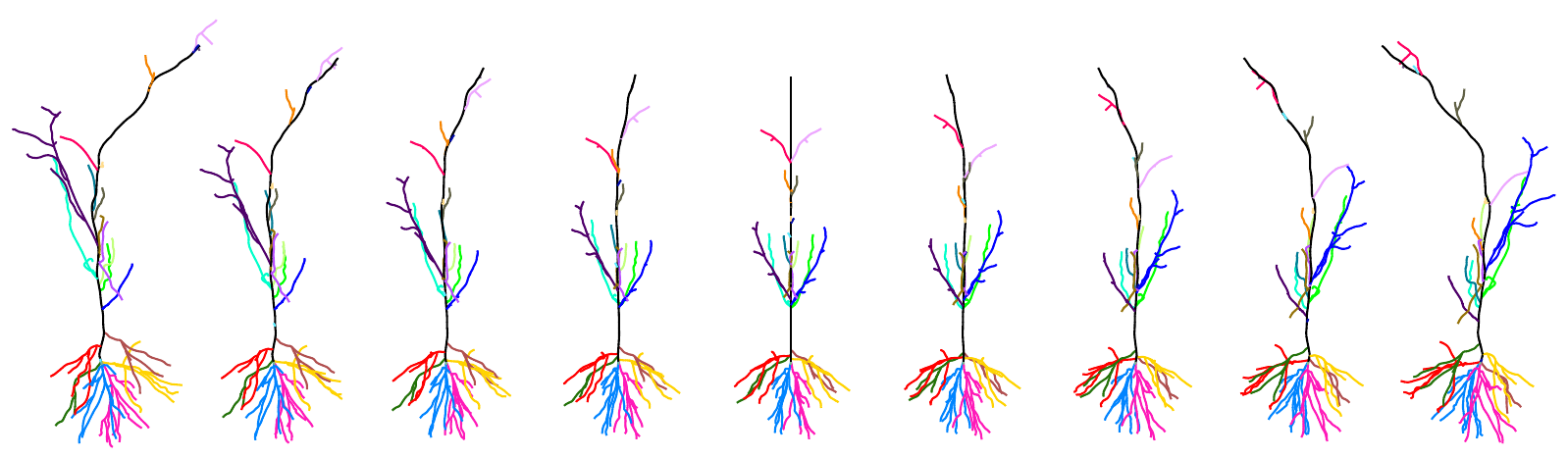}\\
\small{ Geodesic length $d_s = 639.7$.}
\caption{Symmetry and symmetrization of 3D neuronal structures. \guanwangRRR{We show the geodesic between the most left and the most right trees as well as the color-coded subtree-wise correspondences.} The Supplementary Material includes more examples.}
\label{fig:symmetry_neurones}
\end{figure}

Fig.~\ref{fig:symmetry_trees} shows two examples of \guan{complex} botanical trees, and Fig.~\ref{fig:symmetry_neurones} shows one example of neuronal structures symmetrized using the proposed approach. Observe that the midpoint of each geodesic is symmetric; thus, the geodesic path provides a natural symmetrization of the given tree-shaped 3D object.

\subsection{Summary statistics}
\label{sec:results_statistics}

\begin{figure}[t]
\center
\includegraphics[width=0.4\textwidth, trim={0 4cm  0 8cm},clip]{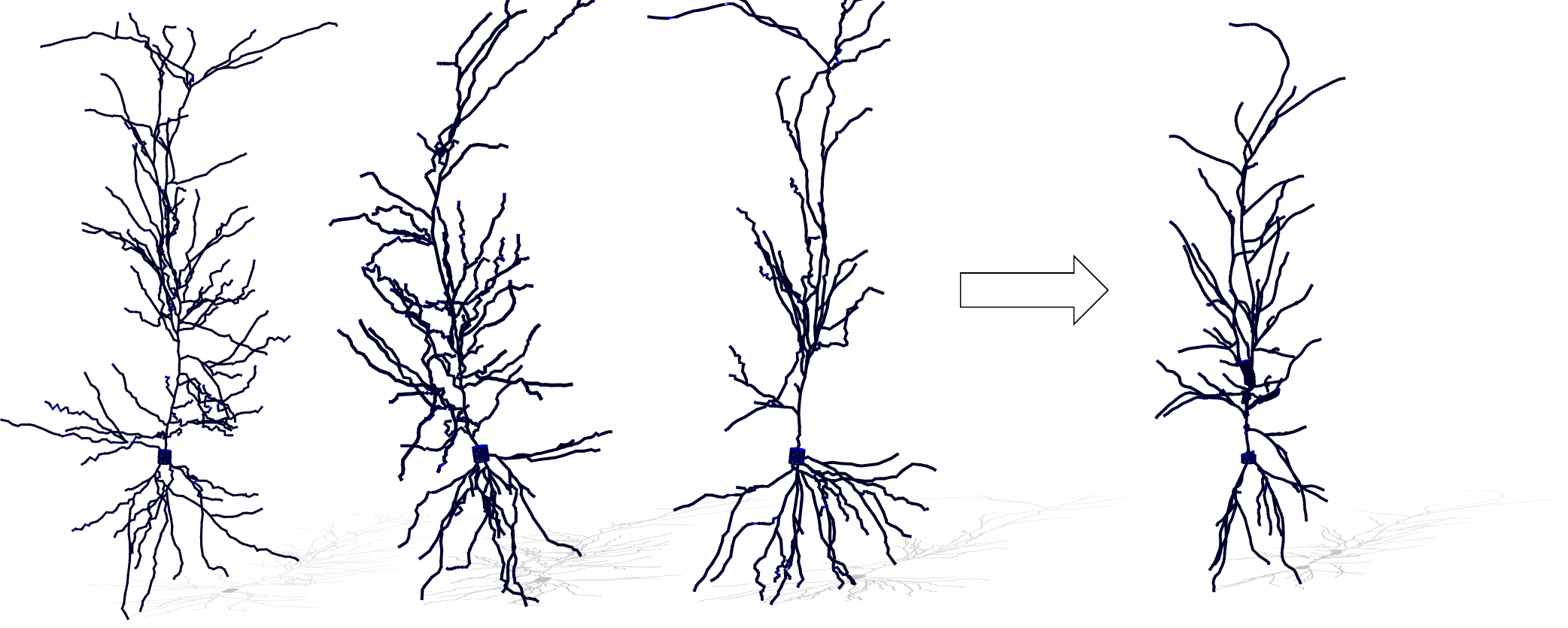}\\
\small{(a) mean shape of three neuronal trees.} 
\caption{Mean shape of 3D neuronal trees. The Supplementary Material includes more examples.}
\label{fig:mean_neuronTrees}
\end{figure}

Fig.~\ref{fig:mean_neuronTrees} shows one example of mean shapes computed from three 3D neuronal trees. The Supplementary Material provides more examples of mean shapes of 3D neuronal trees and  for 3D botanical trees. From these results, we can see that the mean shapes computed with the proposed framework capture the overall shape information of the input samples.  

\begin{figure}[t]
\center
\includegraphics[width=.5\textwidth, trim={0cm 5cm  0 4cm},clip]{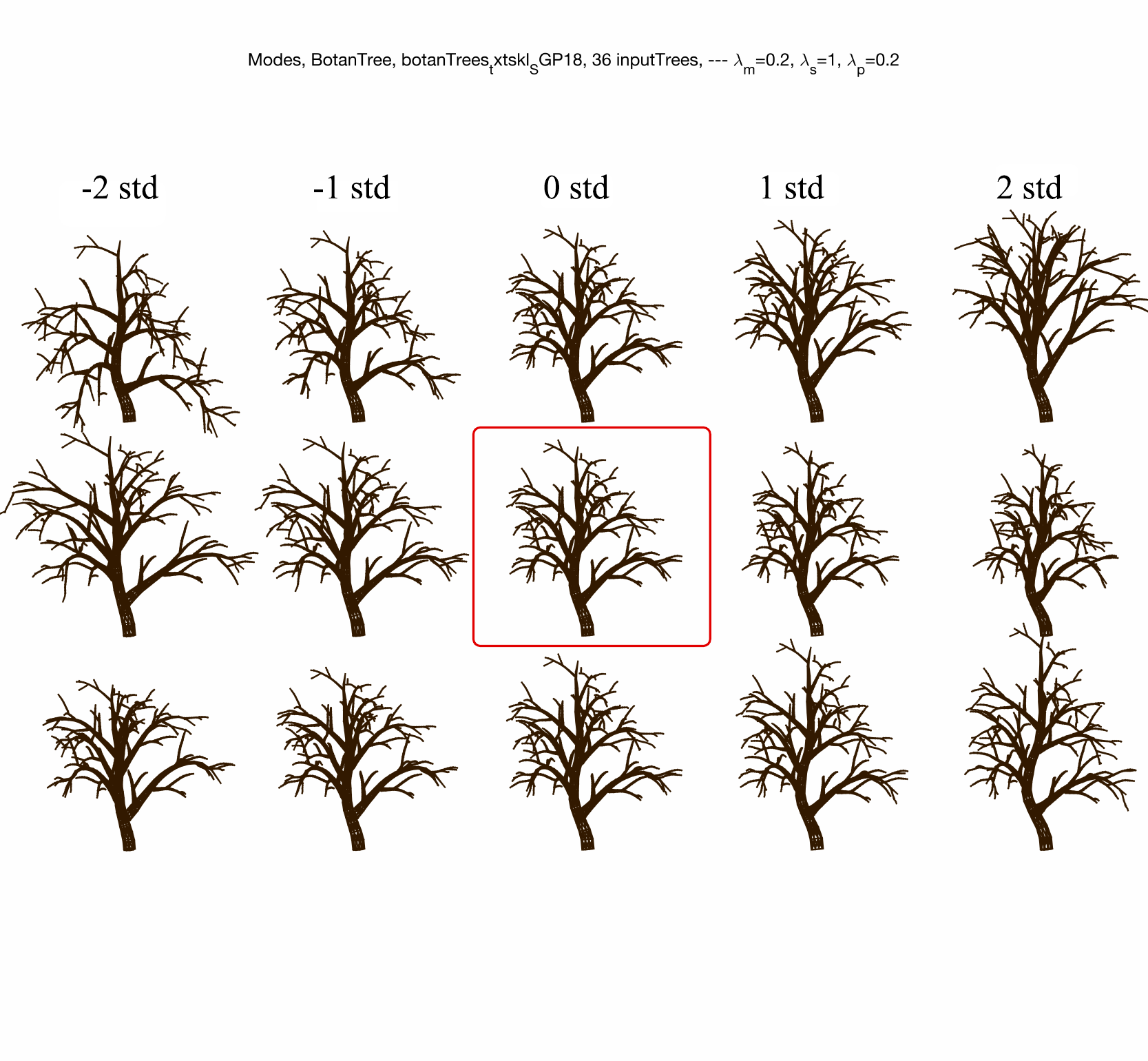}
\caption{Mean tree (highlighted in the middle) and the first three principal modes of variation (one per row) for $36$ botanical trees. The input  botanical trees are shown in the supplementary material. }
\label{fig:modes_botanTrees}
\end{figure}

Next, we perform a covariance analysis of a collection of $36$ models of botanical trees. \guan{Fig.}~\ref{fig:modes_botanTrees} shows the first three leading modes of variation (one per row). The middle shape in each row (at zero standard deviation) is the mean shape of the input collection. We also perform a similar experiment on $51$ neuronal structures; see Fig.~9 in the Supplementary Material. \guan{These two experiments show} that the leading modes of variation capture the main geometric and structural variation of the input 3D tree shape collections.

\subsection{3D Tree-shape generation}
\label{sec:results_synthesis}

\begin{figure}[!ht]
\center
\includegraphics[width=0.5\textwidth, trim={0 9.5cm  400 7cm},clip]{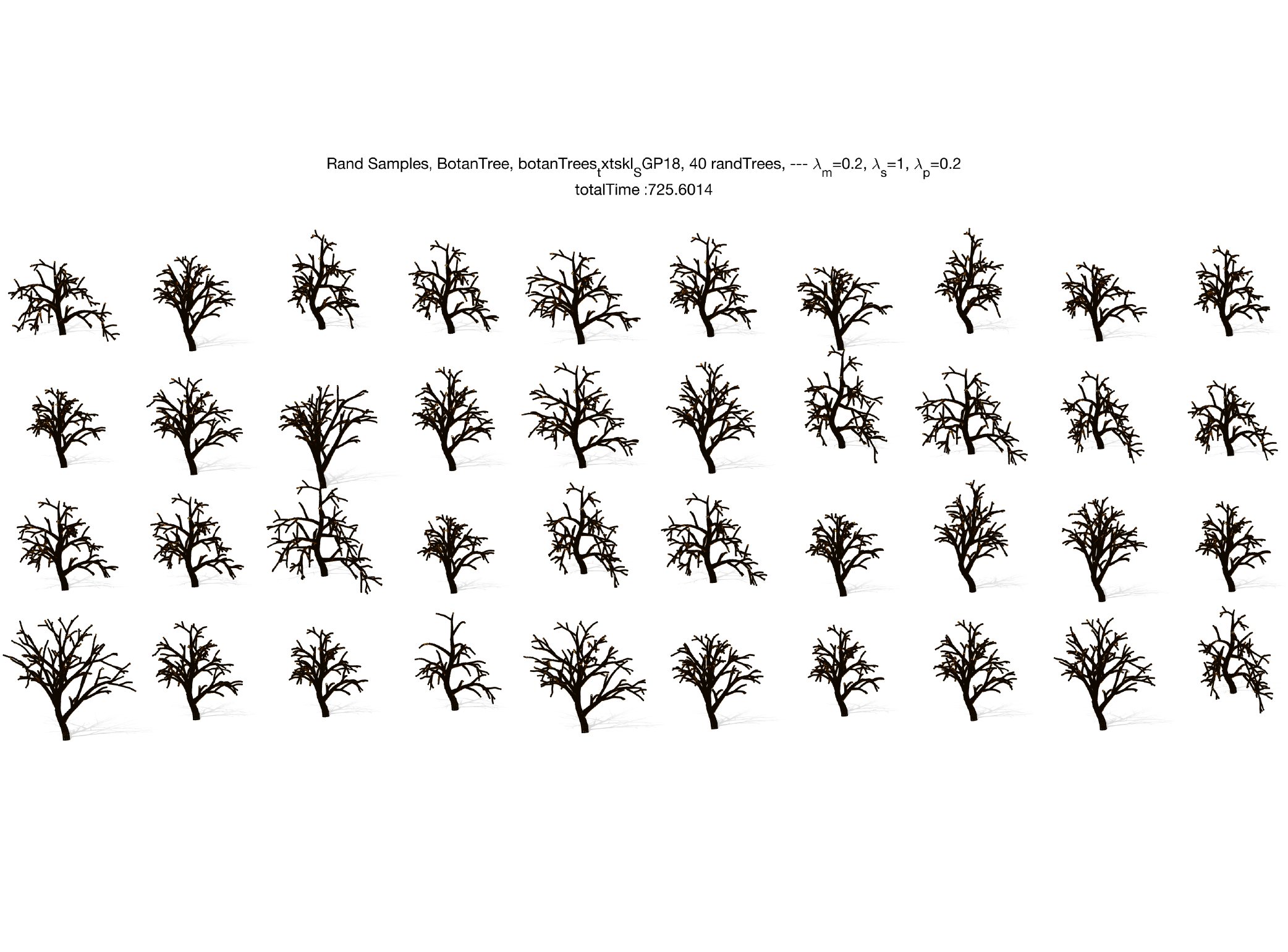}
\caption{Randomly synthesized botanical trees. The Supplementary Material includes more examples.}
\label{fig:randSamples_botanTrees}
\end{figure}

Finally, once a statistical model is fit  to a collection of tree-shaped 3D objects, one can synthesize new 3D shape instances via random sampling from the statistical model. Fig.~\ref{fig:randSamples_botanTrees} shows examples of randomly synthesized samples. These tree shapes have been produced without any professional knowledge. We can clearly see that they share some shape similarities with the input trees (see Fig.~7 in the Supplementary Material), but are not identical. 


\subsection{Quantitative analysis on model performance}


We quantitatively evaluate the performance of the proposed framework and compare it to Wang \etal~\cite{wang2018the} using the Minimum Description Length (MDL) criteria. Specifically, we use a  botanical tree collection composed of \textcolor{black}{6 examples} and compute the eigenvectors $\{\lambda_i\}_{i=1}^m$ and the eigenvalues $\{\Lambda_i\}_{i=1}^m$ of their covariance matrix $\covMatrix$ in their shape space. We record the captured information of the top-$k$ eigenvalues by computing the ratio $\frac{\sum_{i=1}^{k}  \lambda_i}{\sum_{i=1}^{m} \lambda_i}, k \le m$. A good model is the one that captures maximum information with a minimum number $k$ of eigenvalues.  Table~\ref{tab:MDLresults} reports the ratios for top-$1$ to top-$5$ eigenvalues and their corresponding shape space dimensions. As one can see, using the same number of eigenvalues and eigenvectors, our framework captures more information than the approach of Wang \etal~\cite{wang2018the}, even when using the same number of eigenvalues/eigenvectors.

\begin{table}[tb]
\center{
	\caption{\label{tab:MDLresults} Comparison of the performance of the proposed framework with Wang \etal~\cite{wang2018the}. We report the amount of information represented as a function of the number of eigenvalues/eigenvectors.}
	\resizebox{0.5\textwidth}{!}{
		\begin{tabular}{@{}llllllc@{}}
			\toprule
			& \multicolumn{5}{c}{Number of eigenvalues}  & Space dim. \\
			\cmidrule{2-6}
			& $1$ & $2$  & $3$ & $4$ & $5$ & $m$ \\
			\midrule
			Wang \etal~\cite{wang2018the} & $0.54$ & $0.79$ & $0.92$ & $0.97$ & $\textbf{1.00}$  & $ 2541  $ \\
			\cmidrule{1-7}
			Our method & $\textbf{0.60}$ & $\textbf{0.81}$ & $\textbf{0.94}$ & $\textbf{0.99}$ & $\textbf{1.00}$ & $\textbf{4910}$ \\
			
			\bottomrule
	\end{tabular}}
}
\end{table}

			

\section{Conclusion}
\label{sec:conclusion}
In this paper, we developed a comprehensive framework for quantifying the shape of biological 3D objects that deform both in geometry and topology. These objects are characterized by variable branching structures, sizes, and shapes. This framework provides tools for quantifying shape differences, computing geodesic deformations, and modeling shape variability. A key idea here is to match common substructures across objects to result in more natural deformations and shape summaries. The resulting framework provides a powerful setting for quantifying complex structural variability in tree-like 3D objects such as neuronal structures and botanical tree models.

Although effective as evidenced by the results presented in this paper, there are several improvements that can be considered for future work. \guanwang{First, the geodesic computation algorithm is based on a gradient-descent approach and thus converges to a local minimum. In practice, many heuristics can be used to approach the global minimum, \eg by running the algorithm multiple times, each time with a random initialization, and then choosing the optimal solution. While effective, this approach can result in an increase in computation time. Thus, we plan in the future to explore better optimization approaches.} Second, the geodesic quality depends on the weights of the three terms of the distance metric defined in Eqn.~\ref{eq:init_distanceTwoTrees}. In this work, we manually set those weights, while in practice they depend on specific medical knowledge, \eg plant biology and medicine. Thus, one can consider learning these parameters directly from data labeled with domain-specific knowledge.
Third, currently, we have only considered the synthesis by random sampling and not taking into account biological factors such as light intensity,  aging processes, disease, etc., which affect the 3D shape of tree-like structures such as botanical trees and neurons. One potential direction for future work is to incorporate these factors into the modeling process, \eg via regression. 
\hamid{Fourth, we plan to explore in the future how to model deformations and growth patterns of tree-shaped 3D objects as trajectories in the proposed SRVFT space, similar to~\cite{laga20224d}.}
\guan{Fifth, we have not considered the functional correspondence and functionality when we align two trees in this paper. So in the future, we plan to incorporate the part functionality to achieve function-consistent geodesic.}
Finally, the proposed framework can be used to enrich data collections, and thus can benefit data-driven applications such as deep learning-based 3D reconstruction~\cite{han2019image}.

\vspace{6pt}
\noi\textbf{Acknowledgement.} We would like to thank (1) the reviewers and associate editor for their suggestions and feedback during the review process, (2) Aasa Feragen for making her code publicly available and (3) Jing Ren for making her MATLAB rendering toolbox publicly available. This work is supported by the Australian Research Council through the Discovery Grant no. DP22010219, and Scientific Research Foundation for Yangtze Delta Region Institute of University of Electronic Science and Technology of China (Huzhou) no. U032200114. This research was also supported in part by NSF grants IIS 1955154 and DMS 1953087 to Anuj Srivastava.

\bibliographystyle{IEEEtran}
\bibliography{references}

\setcounter{figure}{0}
\setcounter{table}{0}
\setcounter{section}{0}
\vspace{8pt}
\centerline{\bf{\Large{Supplementary Material}}}
\vspace{5pt}

In this Supplementary Material, we provide additional results that could not fit into the main manuscript due to the page limit.   

\section{Geodesics and symmetry analysis}

\begin{figure*}[!ht]
	\center
	\includegraphics[width=\textwidth, trim={0 0cm  0 9cm},clip]{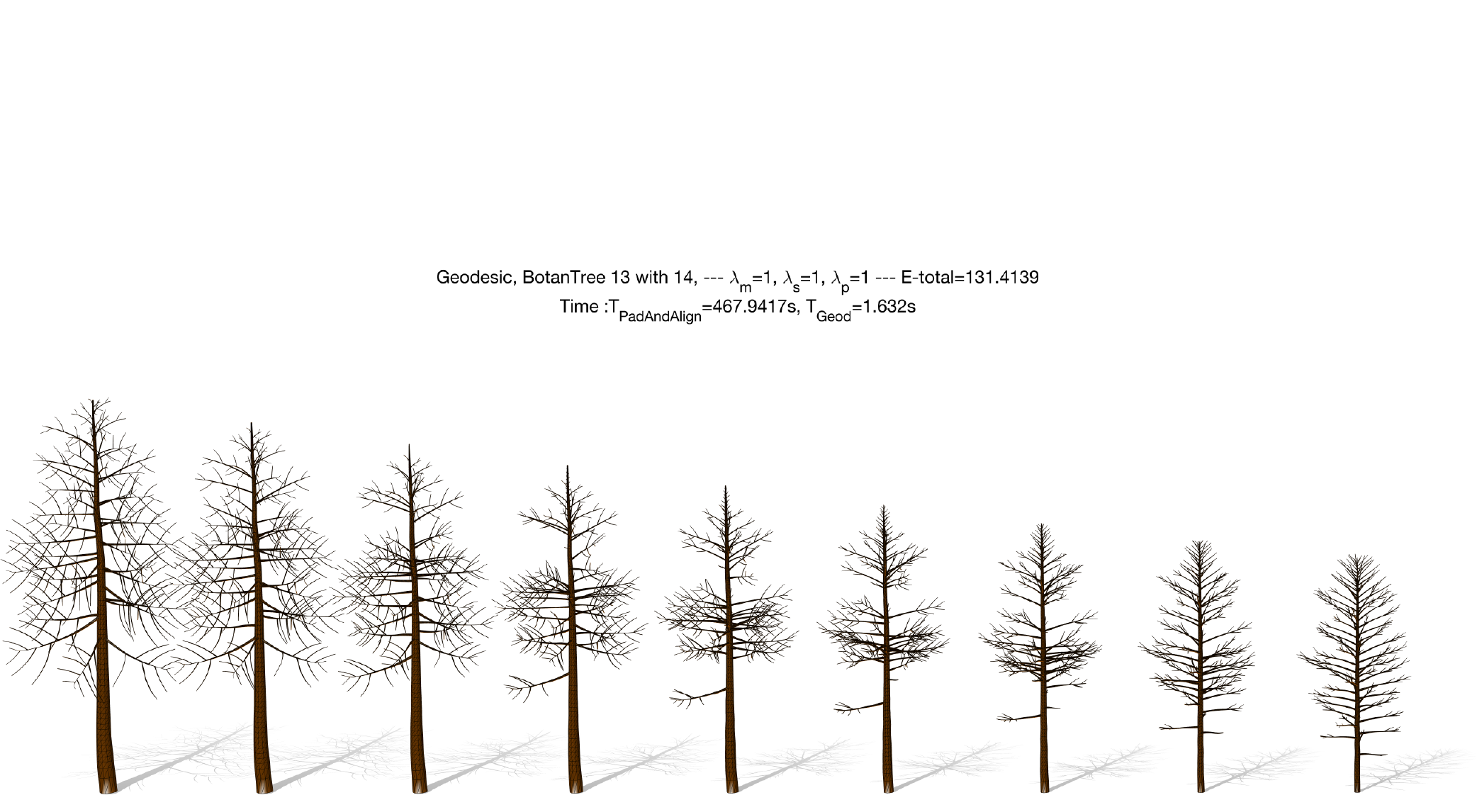}\\
	\small{ Geodesic length $d_s = 126.8$.}
	
	\caption{Geodesic deformations between the most left and the most right 3D botanical trees in each row. In this experiment, we use $\lambda_m=\lambda_s=\lambda_p=1.0$.}
	\label{fig:geodesic_trees_Example2}
\end{figure*}

Fig.~\ref{fig:geodesic_trees_Example2} shows an example of a geodesic  between two botanical trees. Fig.~\ref{fig:comparison_3botanGeods} compares geodesics computed with our approach, the approach of Feragen \etal~\cite{feragen2013toward}, and the approach of Guan \etal~\cite{wang2018the}. Fig.~\ref{fig:symmetry_trees_Examples2} shows two additional examples of symmetry analysis and  symmetrization of 3D botanical and neuronal trees. 
Tab.~\ref{tab:boatanicalTree_geodesics} and Tab.~\ref{tab:neuronalTree_geodesics} summarize the complexity of the 3D trees used in Fig. 3 and Fig. 4 in the main manuscript
.

\begin{table*}[!t]
    \center{
    \caption{\label{tab:boatanicalTree_geodesics} {Complexity of the botanical trees used in Fig. 3 in the main manuscript.}}
    \resizebox{0.8\textwidth}{!}{
    \begin{tabular}{@{}lcccccc@{}}
    \toprule
          & Fig.3(a) 	& Fig.3(a)	& Fig.3(b) & Fig.3(b) & Fig.3(c) & Fig.3(c) \\
          & (source) & (target) & (source) & (target) & (source) & (target)\\
    \midrule
    Number of layers &  $ 3$ & $ 4$ &$4 $ & $4 $ & $ 4$ & $ 3$ \\
    Number of branches & $ 14$ & $ 251$ &  $192 $ & $251 $ & $485 $ & $215 $ \\
    
    \bottomrule
    \end{tabular}}
    }
\end{table*}

\begin{table*}[t!]
    \center{
    \caption{\label{tab:neuronalTree_geodesics} {Complexity of the neuronal  trees used in  Fig. 4 in the main manuscript.}}
    \resizebox{0.8\textwidth}{!}{
    \begin{tabular}{@{}lcccccc@{}}
    \toprule
          & Fig.4(a) 	& Fig.4(a)	& Fig.4(b) & Fig.4(b) & Fig.4(c) & Fig.4(c) \\
          & (source) & (target) & (source) & (target) & (source) & (target) \\
    \midrule
    Number of layers &  $4 $ & $4 $ & $ 4$ & $ 4$ & $ 4$ & $ 4$\\
    Number of branches &  $68 $ & $60 $ & $70 $ & $71 $ & $ 88$ & $ 73$\\
    \bottomrule
    \end{tabular} }
    }
\end{table*}

\begin{figure*}[t]
	\center
	\includegraphics[width=\textwidth]{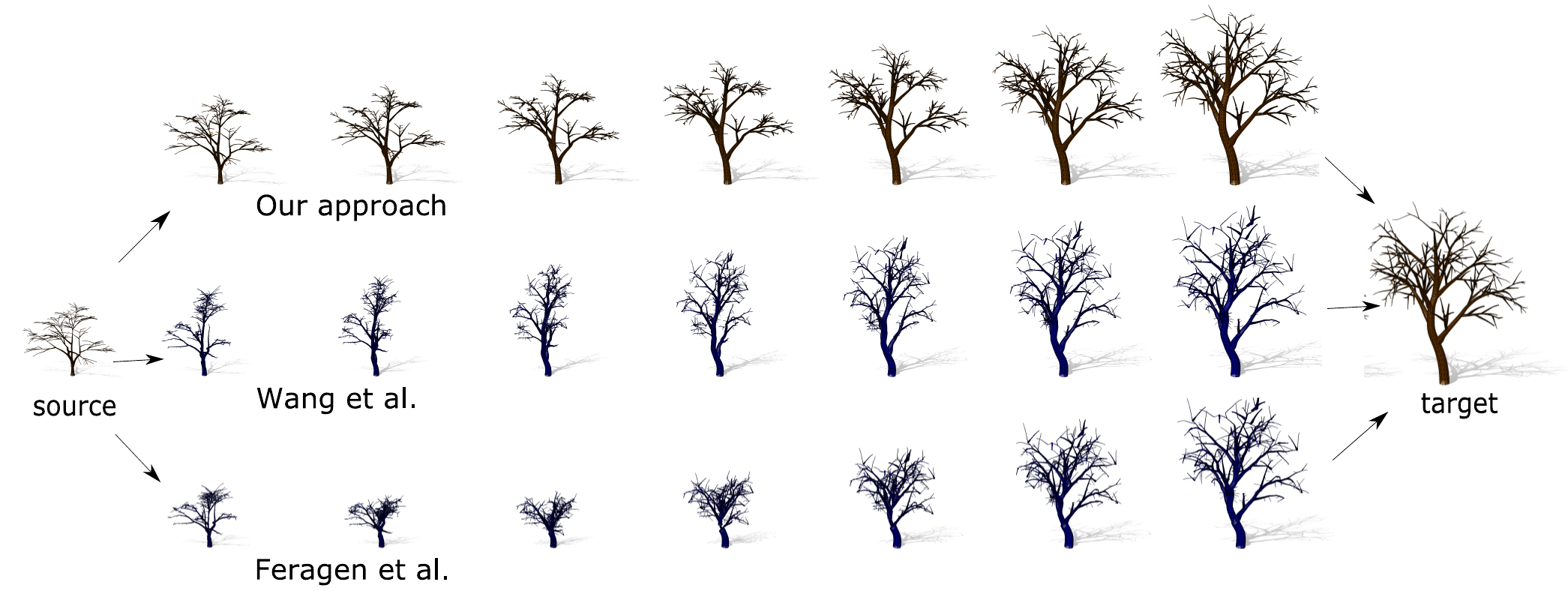} \\
	\caption{Comparison of the quality of the geodesics obtained using our approach, the approach of Feragen \etal~\cite{feragen2013toward}, and the approach of Wang \etal~\cite{wang2018the}.}
	\label{fig:comparison_3botanGeods}
\end{figure*}

\begin{figure*}[t]
	\center 
		\includegraphics[width=0.9\textwidth, trim={0 8.7cm  0 8cm},clip]{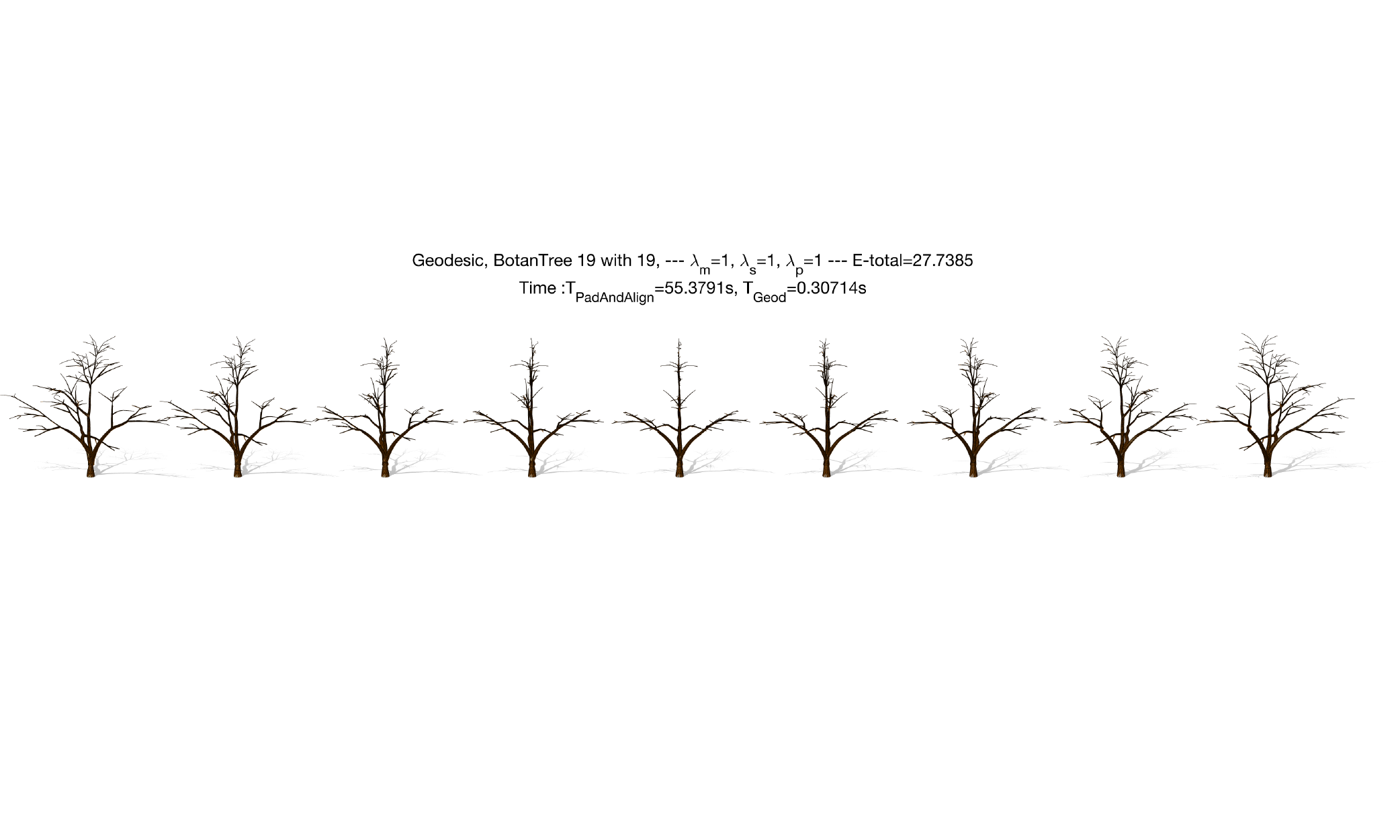}\\
		\small{(a) Geodesic length $d_s = 27.7$.}
		\includegraphics[width=0.9\textwidth, trim={0 6cm  4cm 8cm},clip]{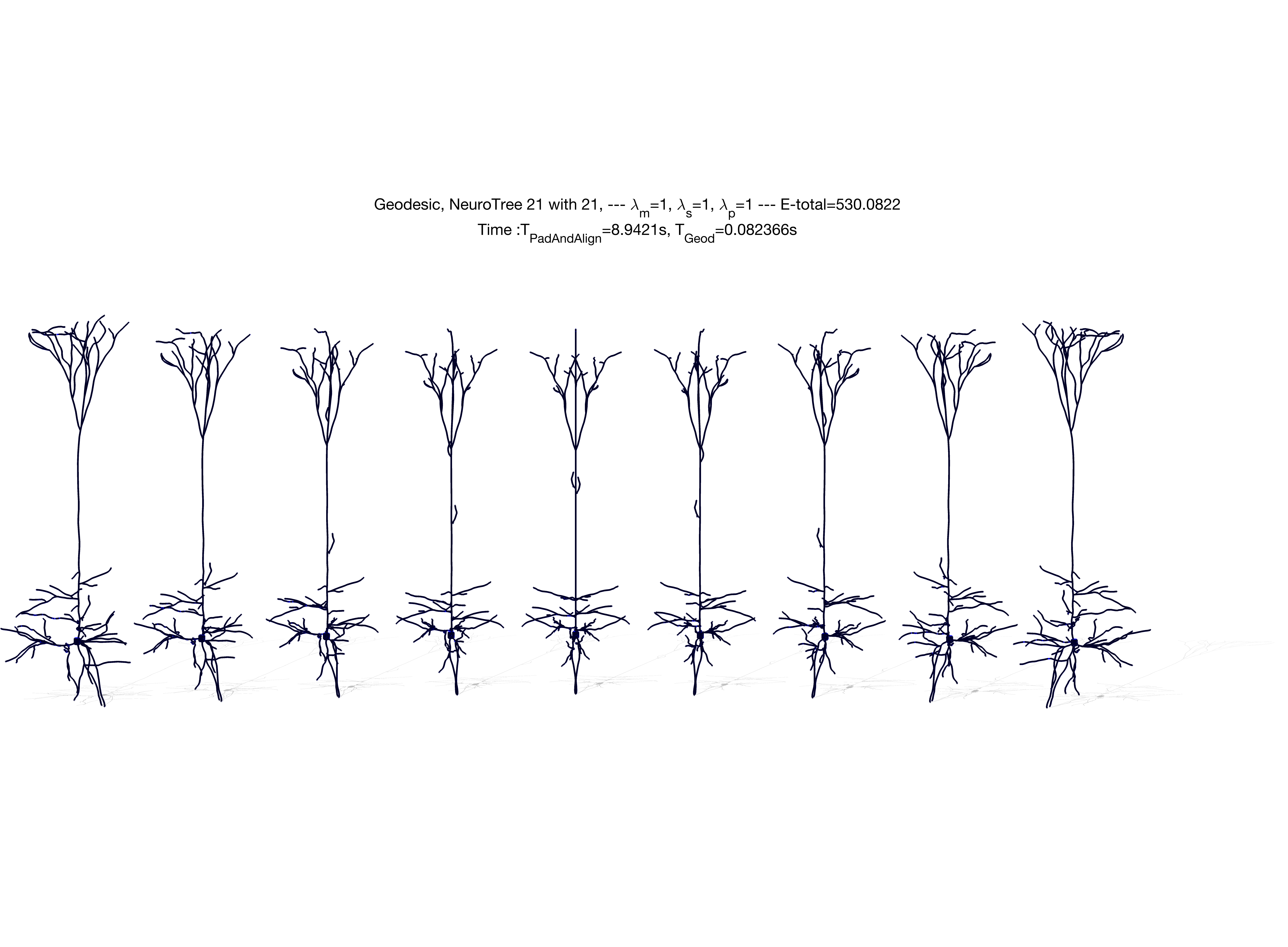}\\
		\small{(b) Geodesic length $d_s = 530.0$. }
		
	\caption{Analysis of the symmetry and symmetrization of (a) 3D botanical trees and (b) 3D neuronal structures. }
	\label{fig:symmetry_trees_Examples2}
\end{figure*}

%
%

\begin{figure*}[t]
	\center
	\includegraphics[width=0.75\textwidth, trim={0 0cm  0 0cm},clip]{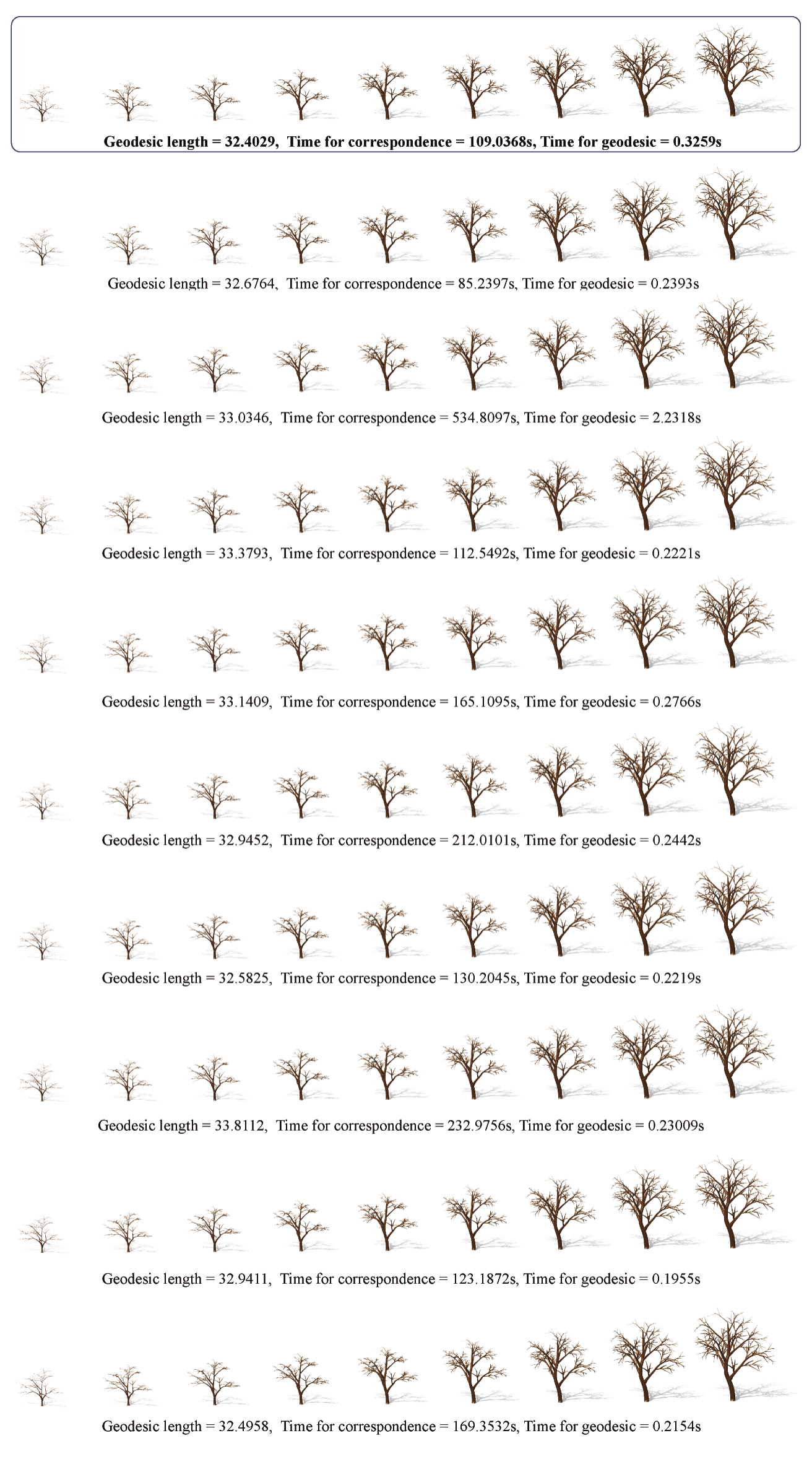}
	\caption{Geodesic deformations between the same left and most right 3D botanical trees with $10$ different  random initializations of the correspondence. Observe that the geodesic length for each initialization varies between $32.4029$ and $33.8112$.}
	\label{fig:tenTimesRunExamples}
\end{figure*}

\vspace{6pt}
\noi\textbf{Effect of the correspondence initialization on the quality of the geodesic.}
The process of computing correspondences and geodesics uses the gradient descent algorithm, which is known to converge to a local minimum. Also, the quality of the local minimum depends on the initialization. In this section, we analyze the effect of the correspondence initialization on the quality of the geodesics computed using the proposed framework. To do so, we consider a source and a target tree and run $10$ times the geodesic computation algorithm, each time with random initialization of the correspondences, and measure the length of the computed geodesic. Fig.~\ref{fig:tenTimesRunExamples} shows the results of the $10$ runs on an example of two 3D botanical tree models. Observe that the geodesic length for each initialization varies between $32.4029$ and $33.8112$. Despite this variation, we observe that the computed geodesics are plausible in all cases.

One common way to overcome this limitation of the gradient descent algorithm is to run it $N$ times, every time with a different random initialization, and then pick up the solution that has the minimum energy (minimum geodesic length in our case). This is illustrated in Fig.~\ref{fig:tenTimesRunExamples} where the optimal solution out of the ten is highlighted (first row).

\section{Summary statistics}

\begin{figure*}
    \center
    \begin{tabular}{cc}
	\includegraphics[width=0.5\textwidth, trim={0 2cm  0 4cm},clip]  {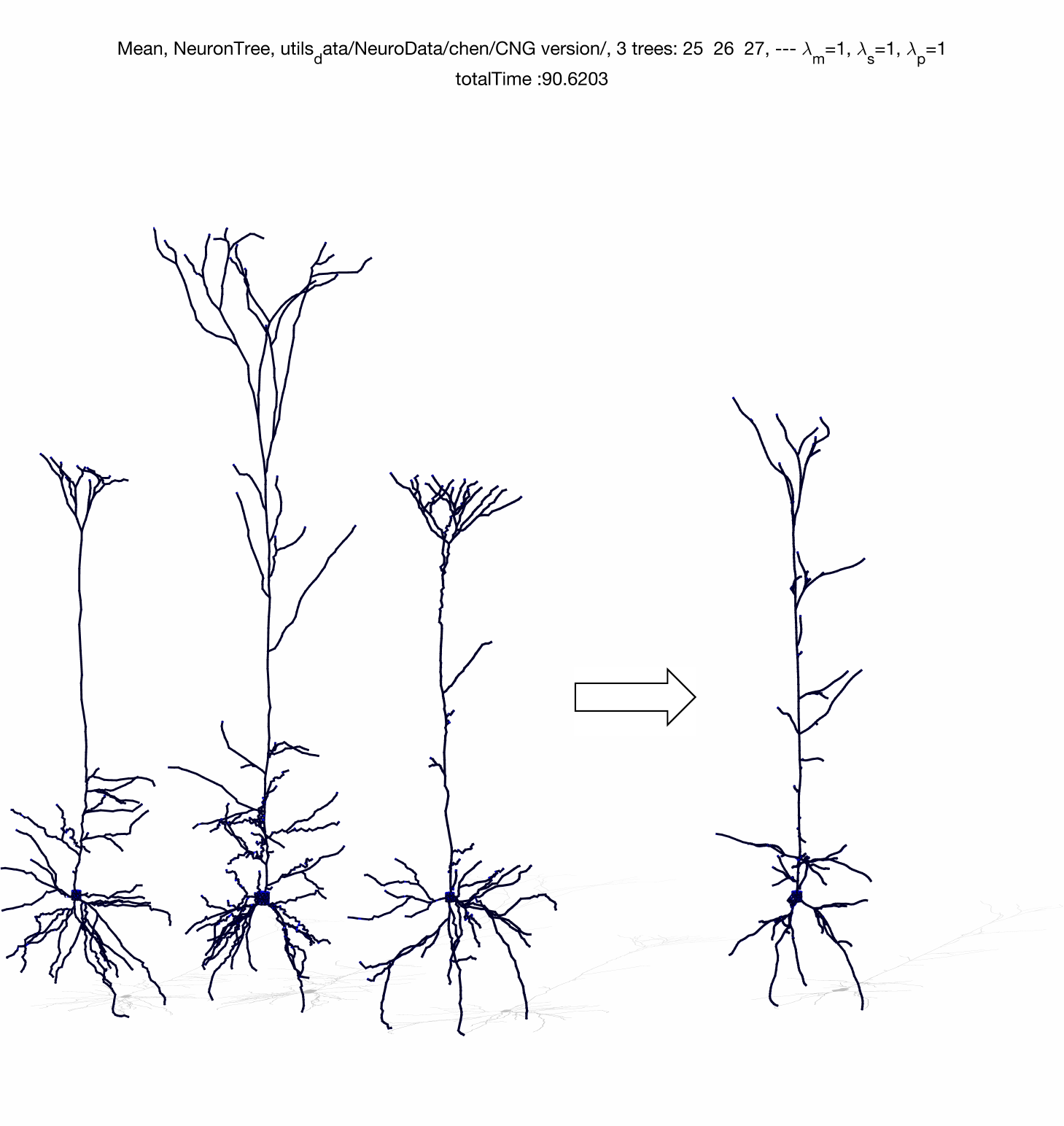} & 
        \includegraphics[width=0.5\textwidth, trim={0 2cm  0 4cm},clip]{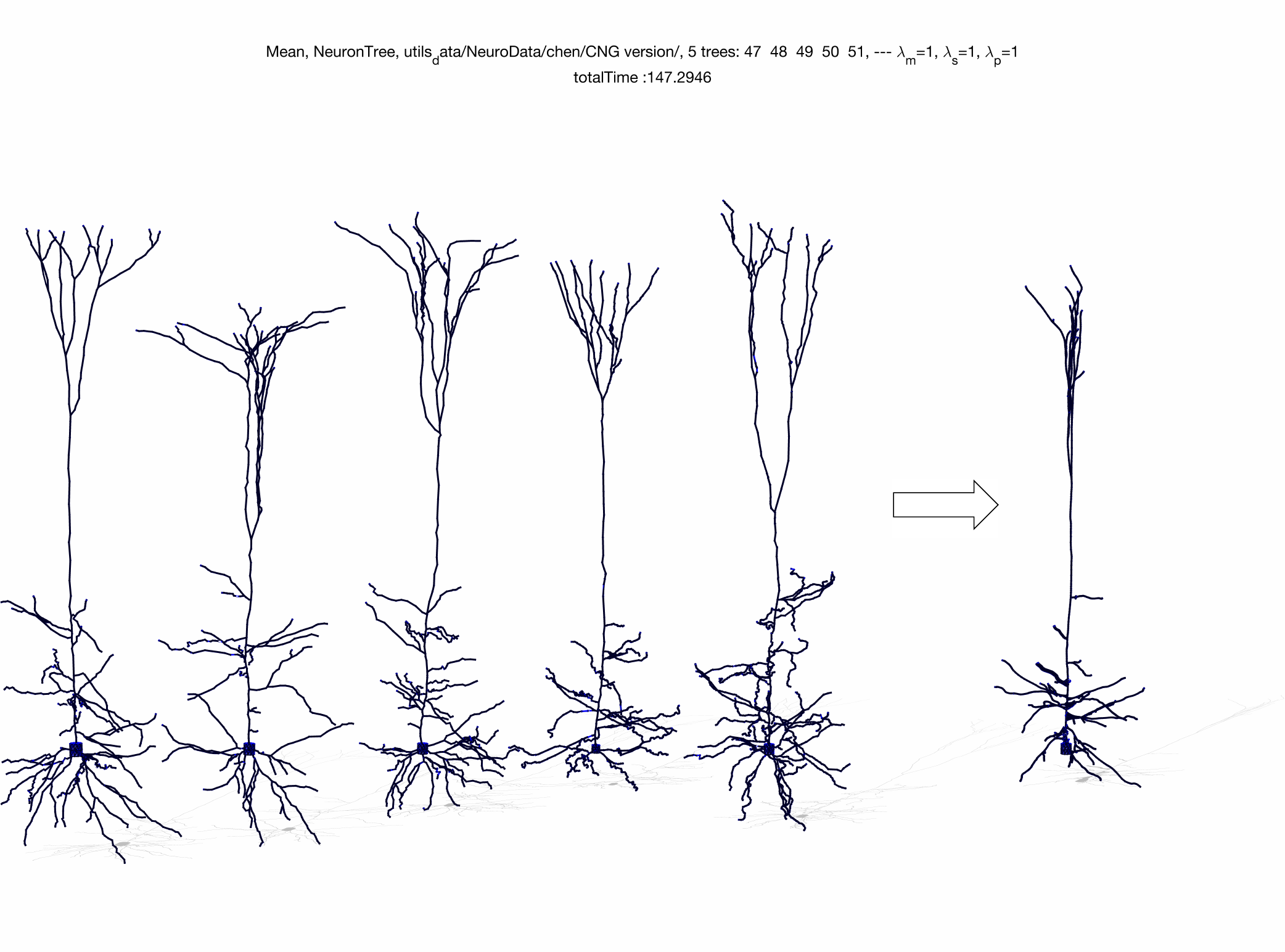}\\
 
	(a) Mean shape of three neuronal trees. & 
        (b) Mean shape of five neuronal trees. \\
        $\lambda_m =1.0, \lambda_s =1.0, \lambda_p =1.0$.&
	$\lambda_m =1.0, \lambda_s =1.0, \lambda_p =1.0$.
    \end{tabular}
	\caption{Mean shape of 3D neuronal trees.}
	\label{fig:mean_3neurones_2}
\end{figure*}

\begin{figure*}
	\center{
     \begin{tabular}{cc}
	\includegraphics[width=0.5\textwidth, trim={0 6cm  0 9cm},clip]{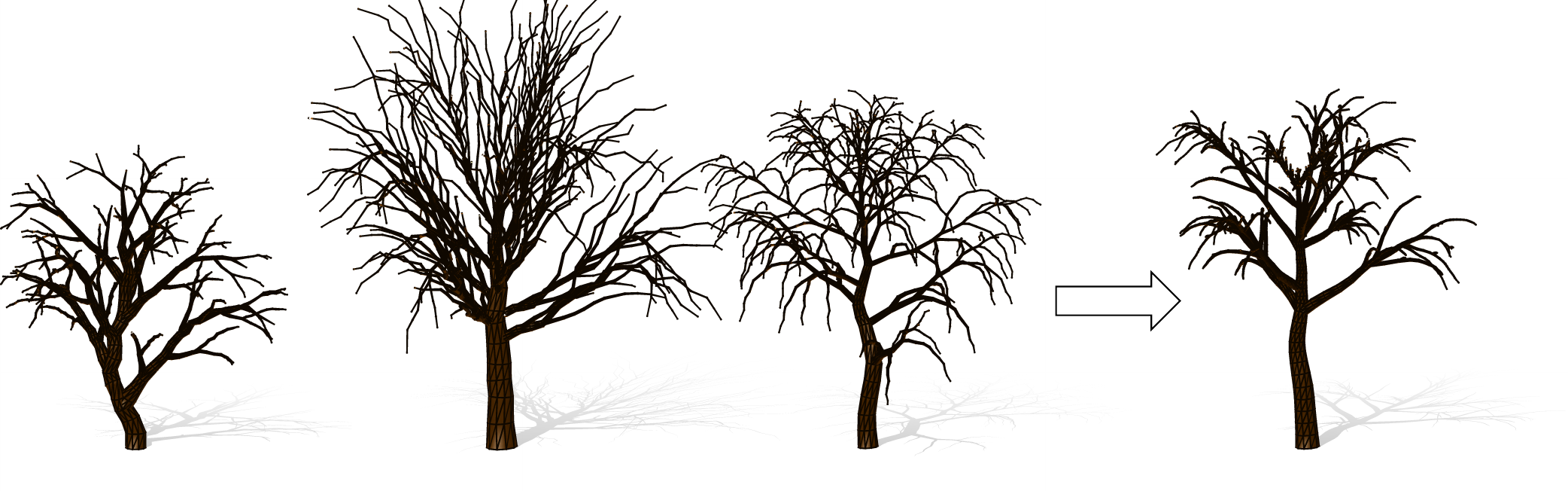} &  \includegraphics[width=0.5\textwidth, trim={0 7cm  0 8cm},clip]{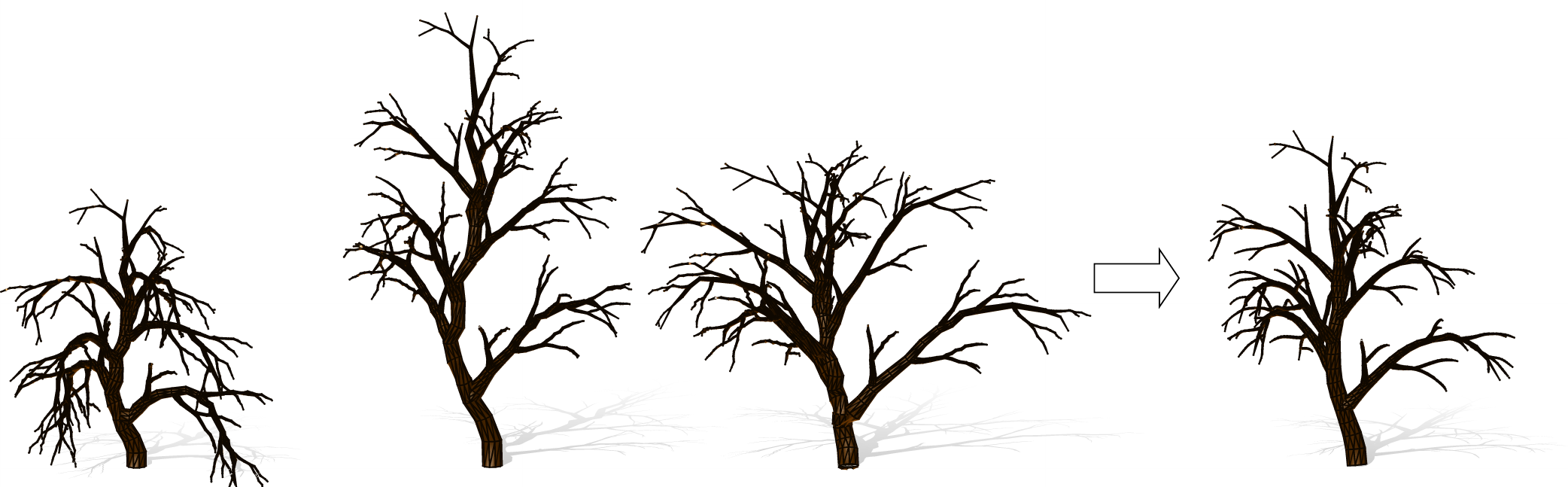}\\
	\small{(a) Mean shape of 3 botanical trees.}  & \small{(b) Mean shape of 3 botanical trees.}\\
        \small{$\lambda_m =1.0, \lambda_s =1.0, \lambda_p =1.0$.} &
        \small{$\lambda_m =1.0, \lambda_s =1.0, \lambda_p =1.0$.}\\

	\includegraphics[width=0.5\textwidth, trim={0 9cm  0 9cm},clip]        {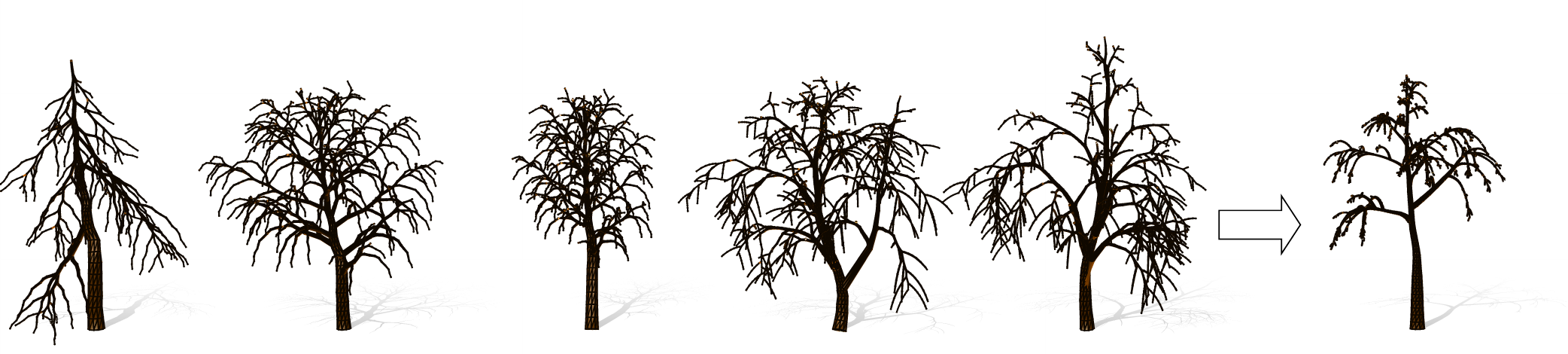} & 
        \includegraphics[width=0.5\textwidth, trim={0 8cm  0 9cm},clip]{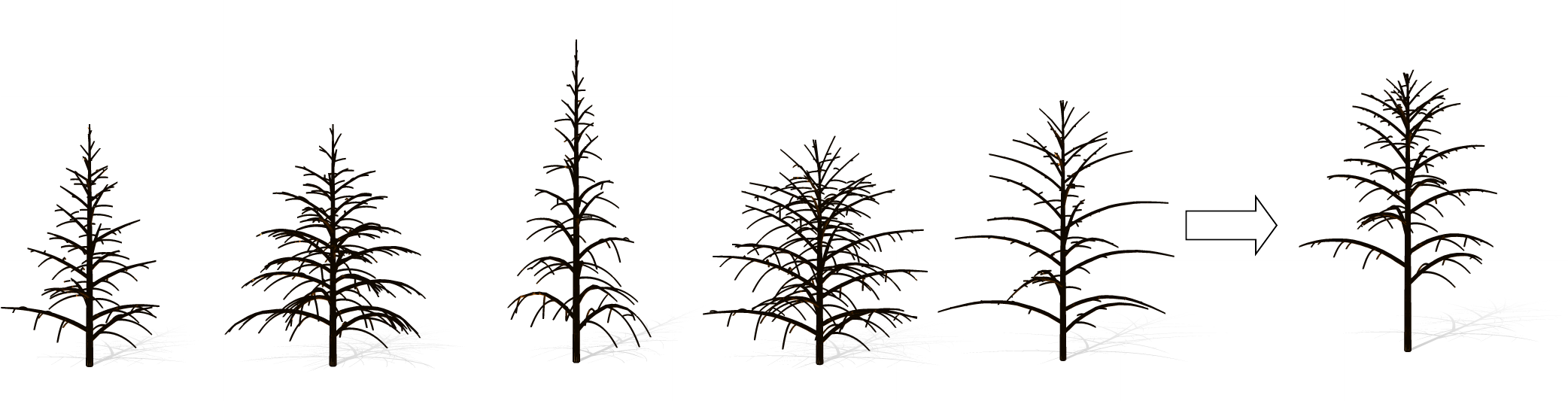}\\
	
        \small{(c) Mean shape of 5 botanical trees.} &
        \small{(d) Mean shape of 5 botanical trees.} \\
        
        \small{$\lambda_m =1.0, \lambda_s =1.0, \lambda_p =1.0$.} &	
	\small{$\lambda_m =0.1, \lambda_s =0.1, \lambda_p =1.0$}
    \end{tabular}
    
    \caption{Mean shape of 3D botanical trees. }
    \label{fig:mean_botanTrees_1}
 }
\end{figure*}

Fig.~\ref{fig:mean_3neurones_2} shows examples of mean shapes of three and five neuronal trees. Fig.~\ref{fig:mean_botanTrees_1}, on the other hand, shows additional examples of  mean shapes of three and five botanical trees.

\subsection{Inputs, modes and random samples}

\begin{figure*}[!h]
	\center
	\includegraphics[width=0.95\textwidth, trim={0 5cm  0 4cm},clip]{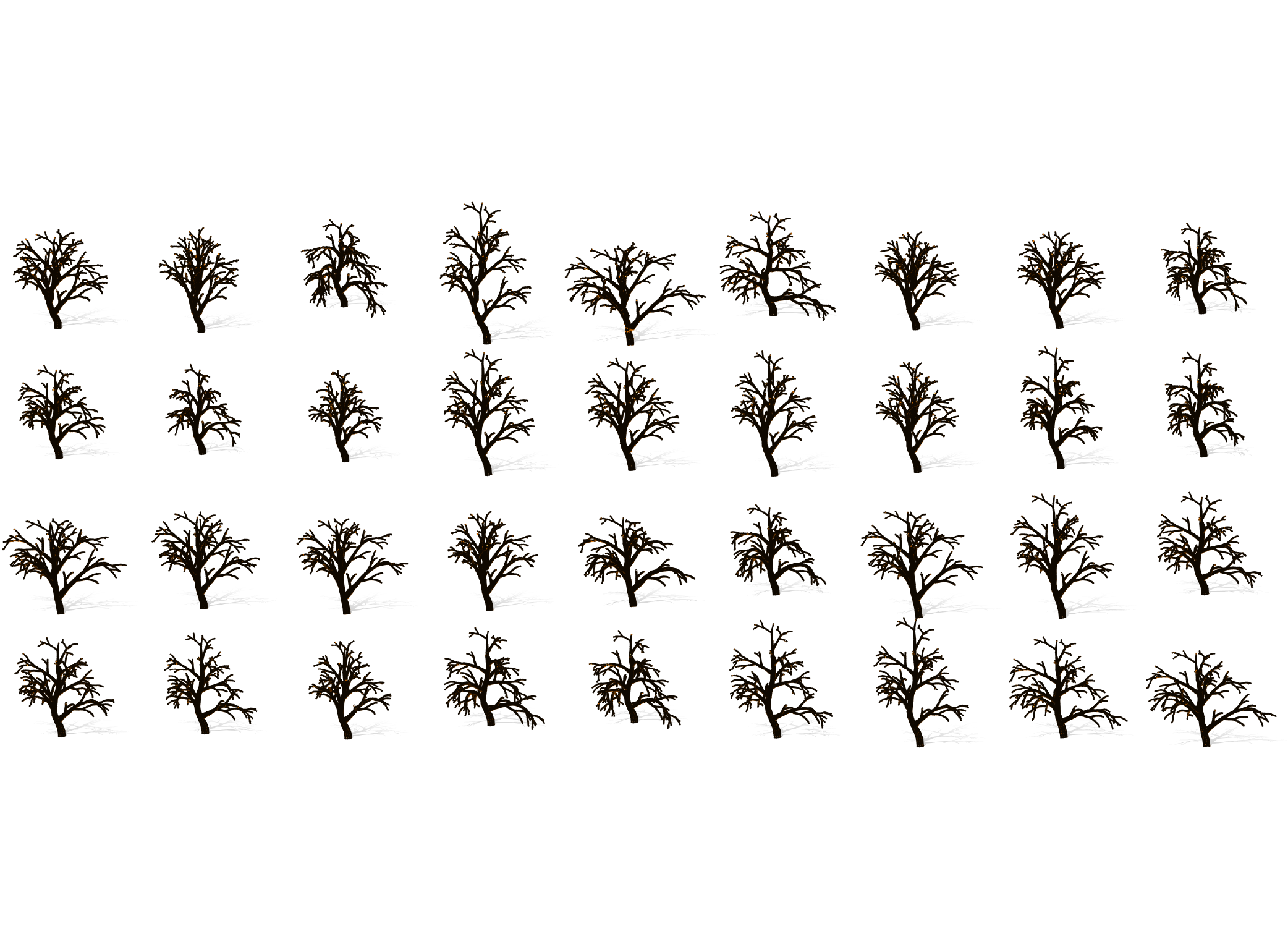}\\
	\caption{Input  3D tree shapes used to generate  the random samples of Fig. 14 in the main manuscript.}
	\label{fig:input_botanTrees_mainScript}
\end{figure*}

\begin{figure*}[!h]
	\center
	\includegraphics[width=0.95\textwidth, trim={0 5cm  0 5cm},clip]{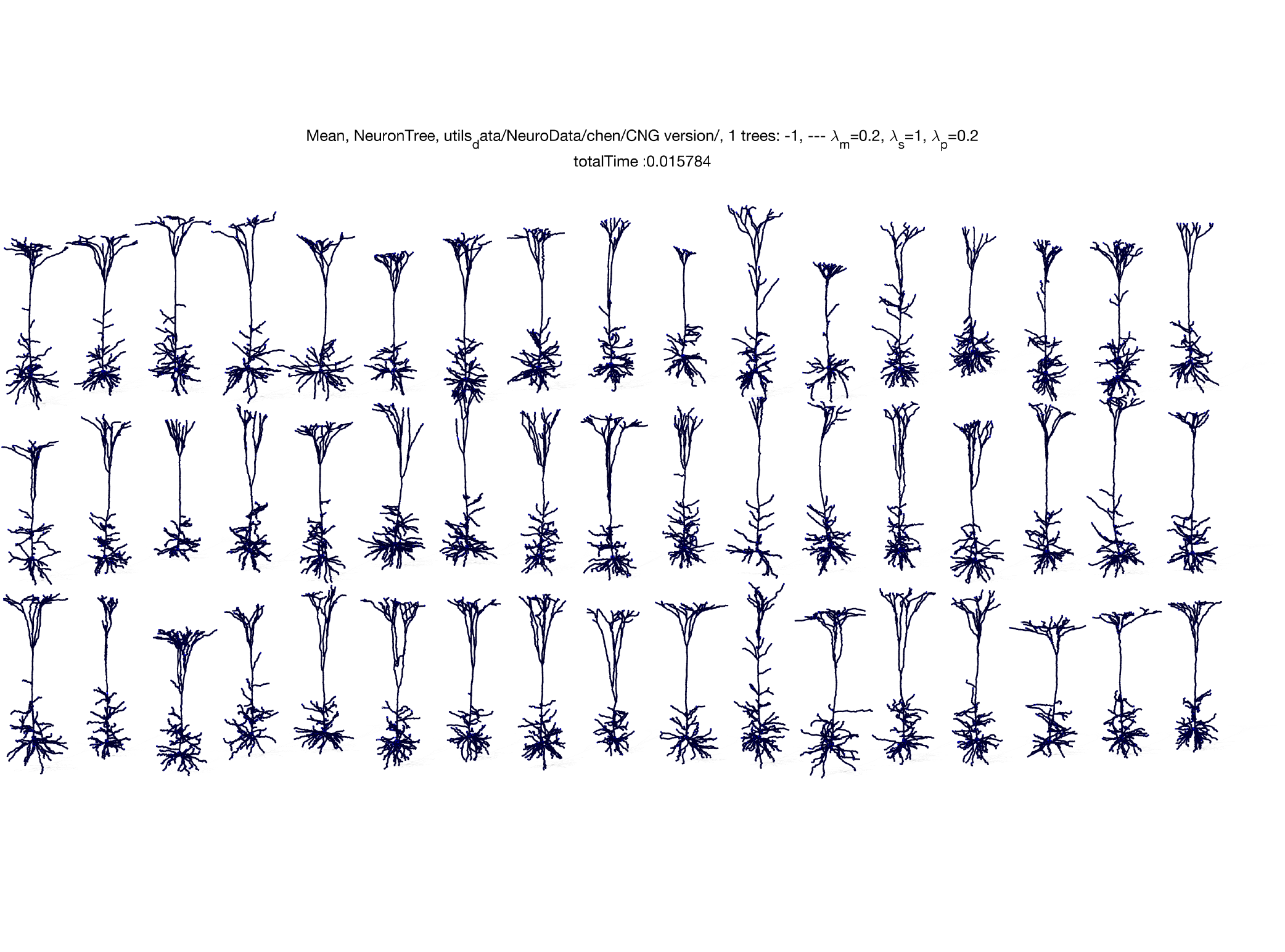}
	\caption{Input 3D neuronal trees for Figs.~\ref{fig:modes_neuronTrees} and ~\ref{fig:randSamples_neuronTrees}.}
	\label{fig:input_neuronTrees}
\end{figure*}

\begin{figure*}[t]
	\center
	\includegraphics[width=0.45\textwidth, trim={20cm  5cm  20cm 5cm},clip]{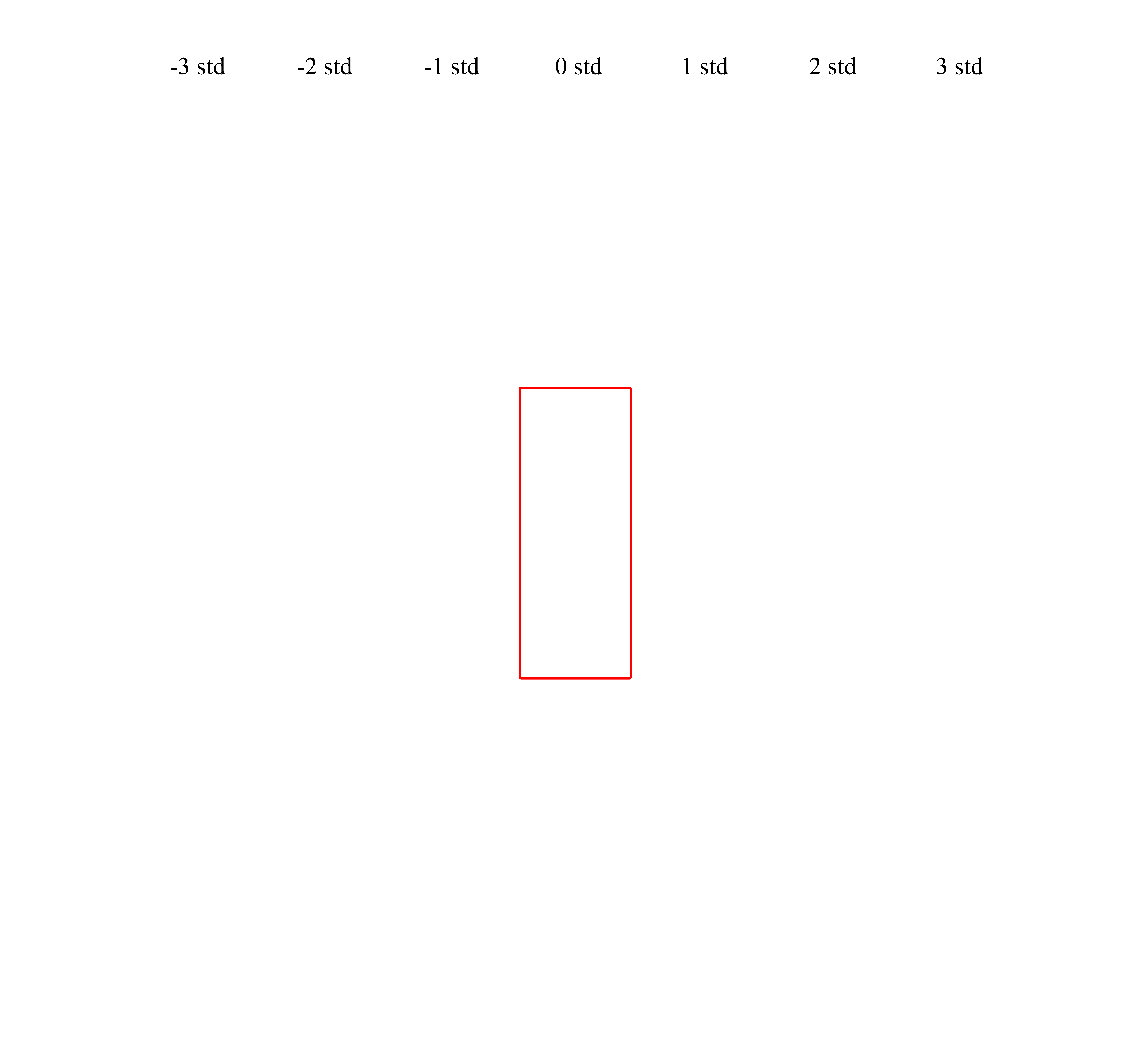}
	\caption{First three principal modes of variation (one per row)  for $51$ neuronal trees.  The mean 3D neuronal tree is highlighted in red.}
	\label{fig:modes_neuronTrees}
\end{figure*}

\begin{figure*}[t]
	\center
	\includegraphics[width=0.5\textwidth, trim={6cm 0cm  6cm 0cm},clip]{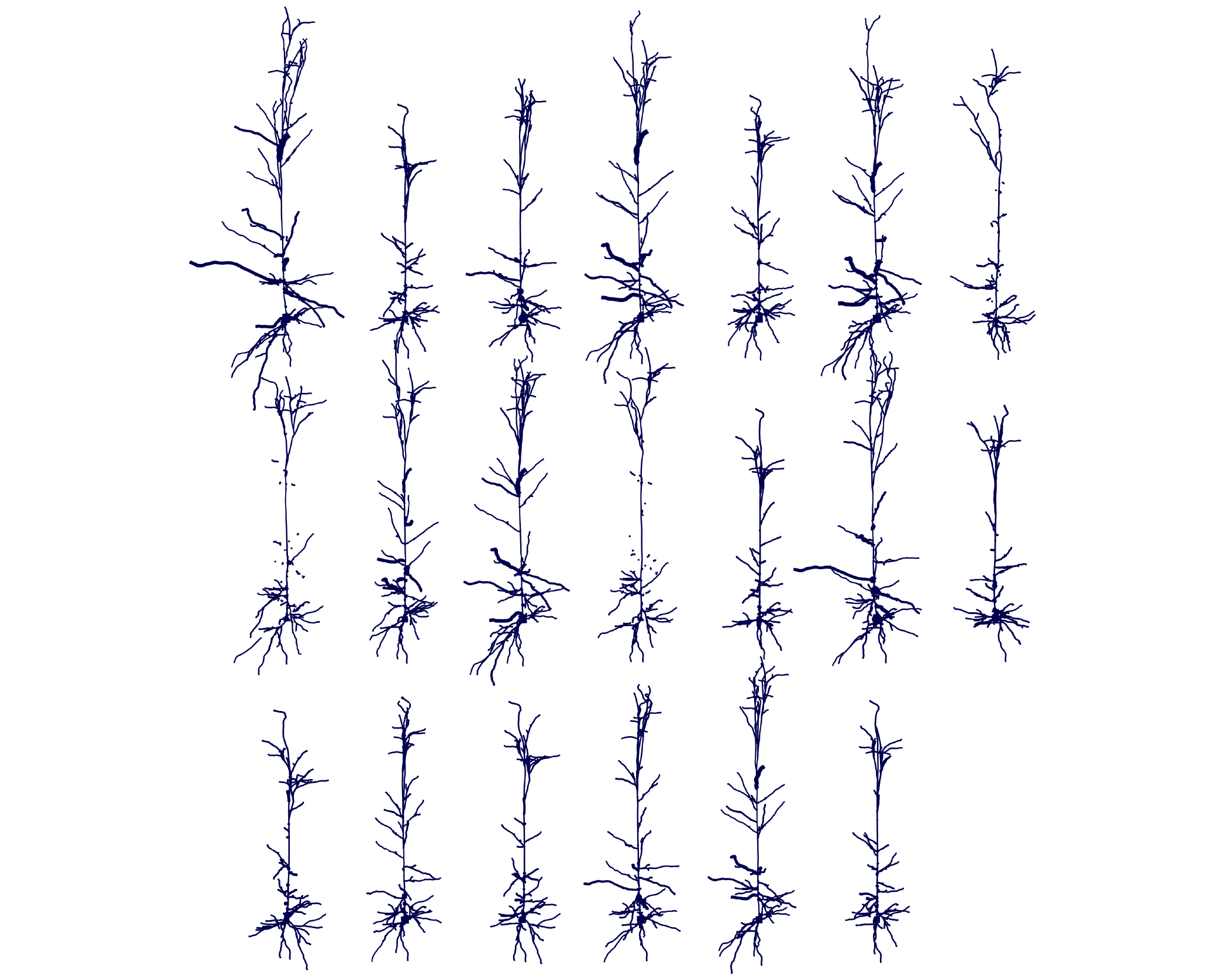}
	\caption{Randomly synthesized neuronal trees.}
	\label{fig:randSamples_neuronTrees}
\end{figure*}

Fig.~\ref{fig:input_botanTrees_mainScript} shows the 3D botanical trees used to generate the random samples of Fig. 14 in the main manuscript. Figs.~\ref{fig:modes_neuronTrees} and ~\ref{fig:randSamples_neuronTrees} are, respectively,  examples of principal modes of variation and randomly synthesized neuronal trees, computed from the collection of $51$ neuronal trees shown in Fig.~\ref{fig:input_neuronTrees}.

\begin{figure*}[!h]
	\center
	\includegraphics[width=\textwidth, trim={0 0cm  0 3cm}, clip]{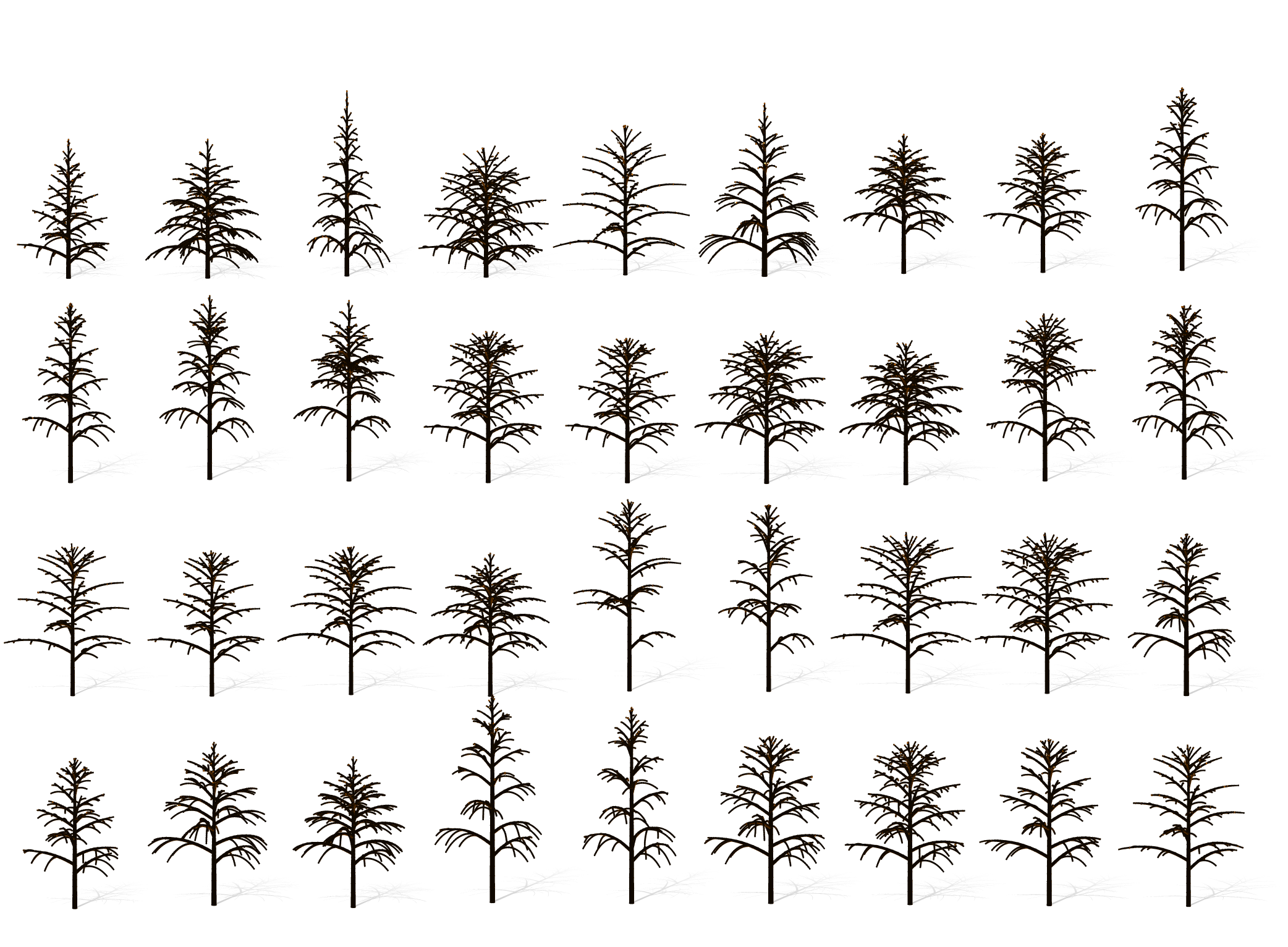}
	\caption{Input botanical trees for Figs.~\ref{fig:modes_botanTrees_case2} and ~\ref{fig:randSamples_botanTrees_case2}.}
	\label{fig:input_botanTrees_case2}
\end{figure*}

\begin{figure*}[!h]
	\center
	\includegraphics[width=\textwidth, trim={0 22cm  0.2cm 18cm}, clip]{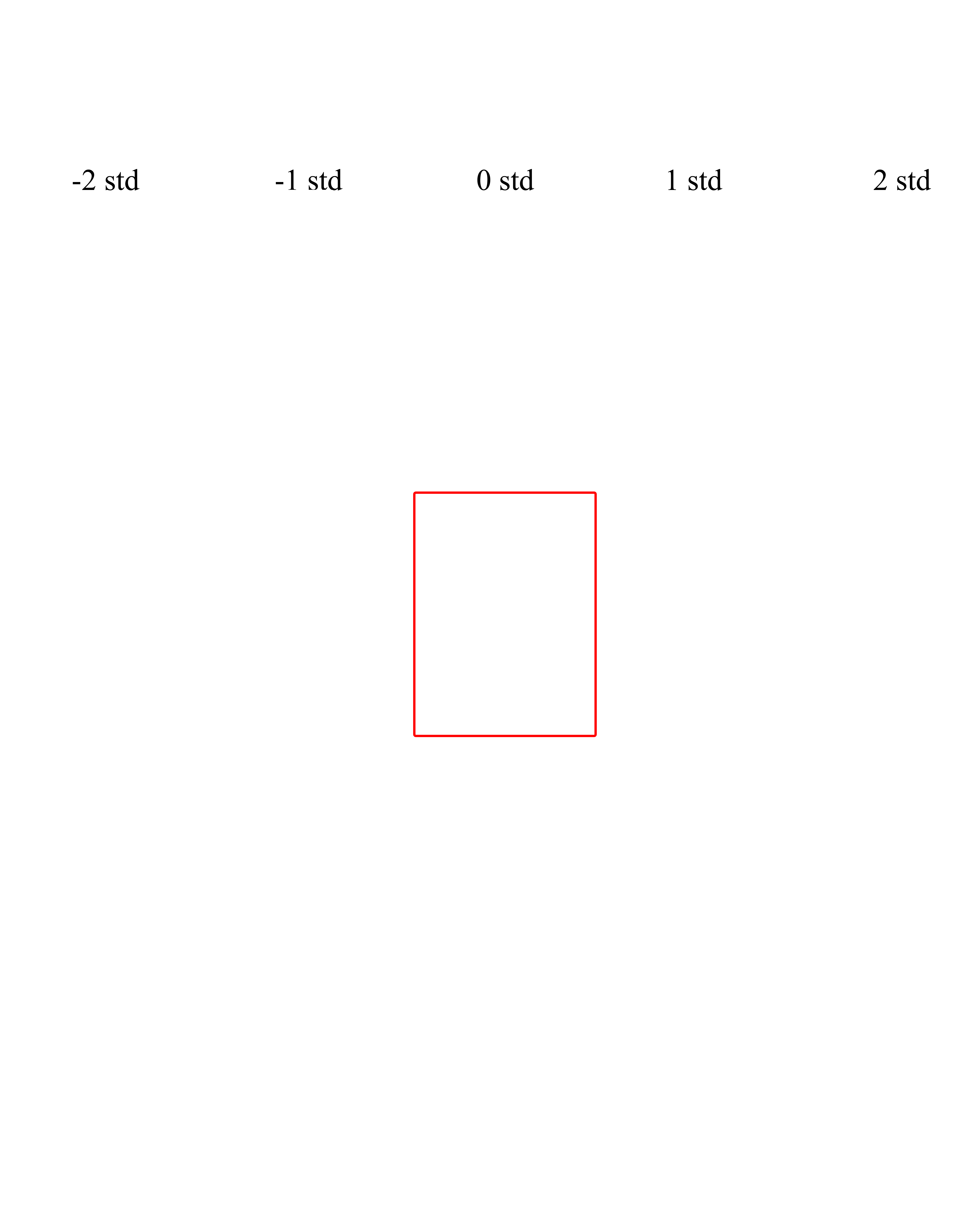}
	\caption{Principal modes of variation for 36 botanical trees. The mean 3D tree is highlighted in red.}
	\label{fig:modes_botanTrees_case2}
\end{figure*}

\begin{figure*}[h]
	\center
	\includegraphics[width=\textwidth, trim={0 6cm  0 6cm},clip]{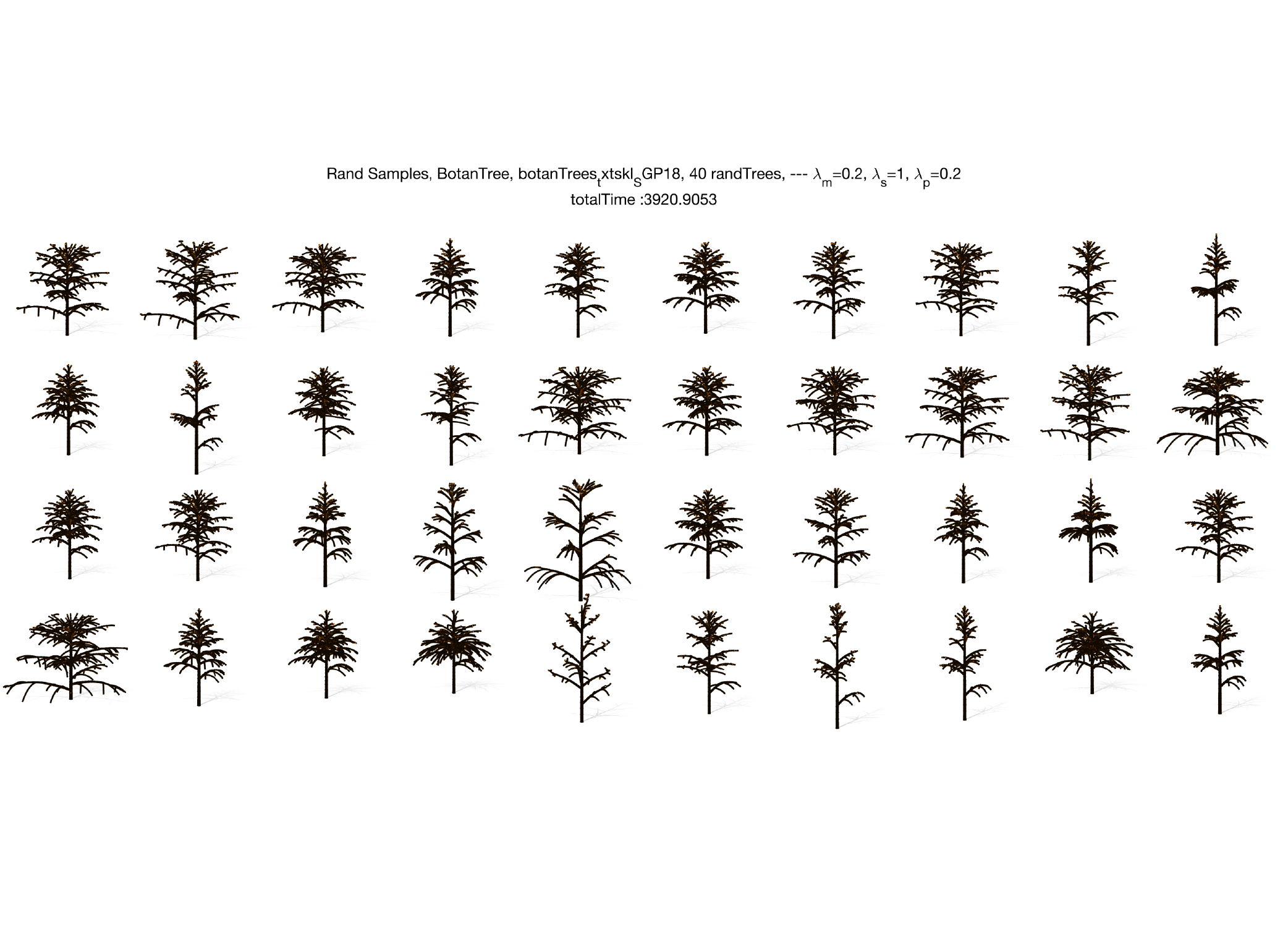}
	\caption{Randomly synthesized botanical trees}
	\label{fig:randSamples_botanTrees_case2}
\end{figure*}

Finally, Fig.~\ref{fig:modes_botanTrees_case2}  shows the leading  principal modes of variation computed from the  $36$ botanical tree models of  Fig.~\ref{fig:input_botanTrees_case2}. Fig.~\ref{fig:randSamples_botanTrees_case2}, on the other hand, shows 3D botanical trees randomly synthesized by sampling from the probability distribution fitted to the population of Fig.~\ref{fig:input_botanTrees_case2}.  We can clearly see that sampling from the probability distribution fitted to the input population allows the synthesis of rich 3D botanical tree models and neuronal tree shapes.

\section{Correspondence Map Visualization}

\begin{figure*}[!ht]
\center
\includegraphics[width=\textwidth]{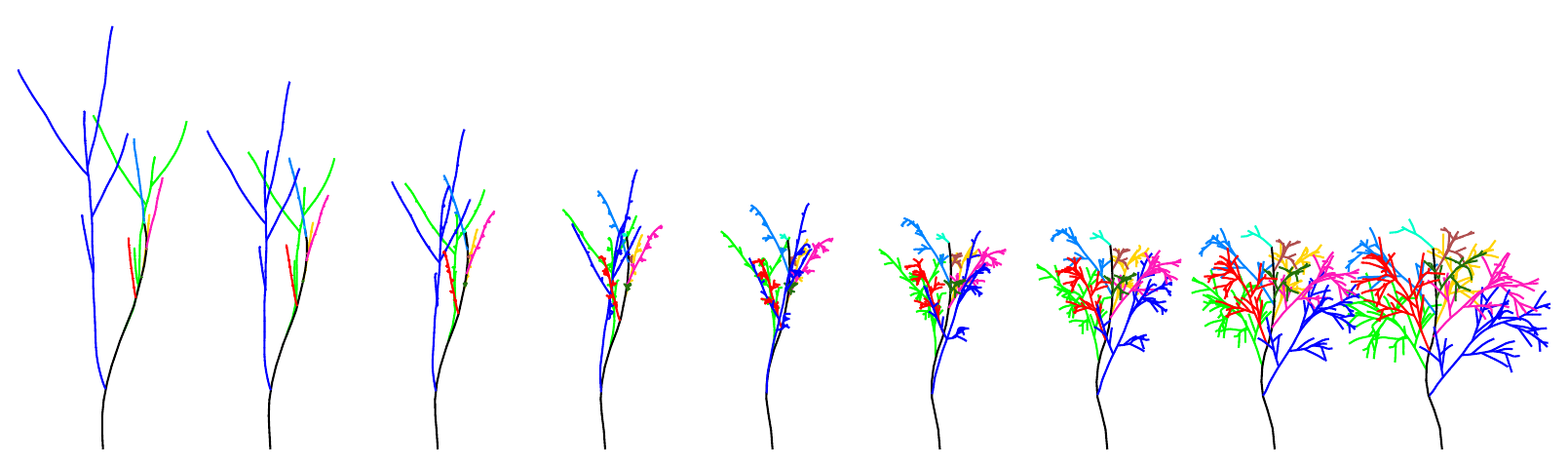}\\
\small{(a) }\\
\includegraphics[width=\textwidth]{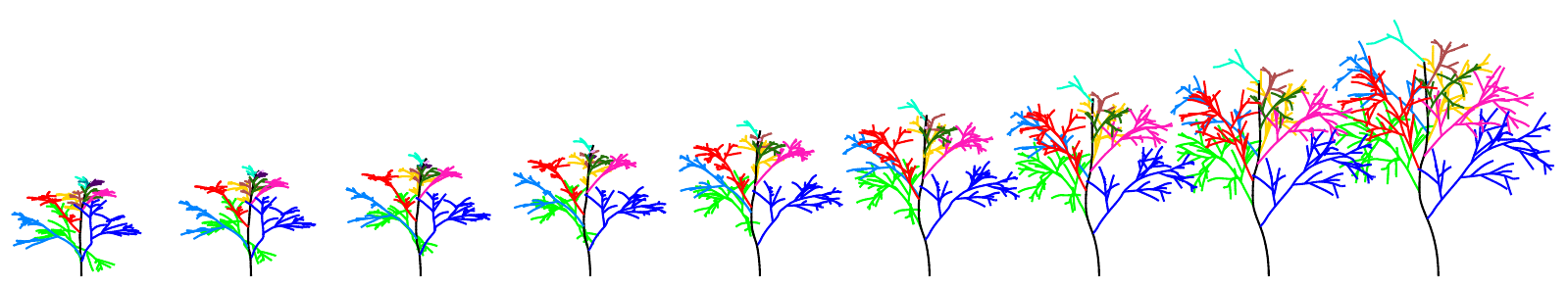}\\
\small{(b)}\\
\includegraphics[width=\textwidth]{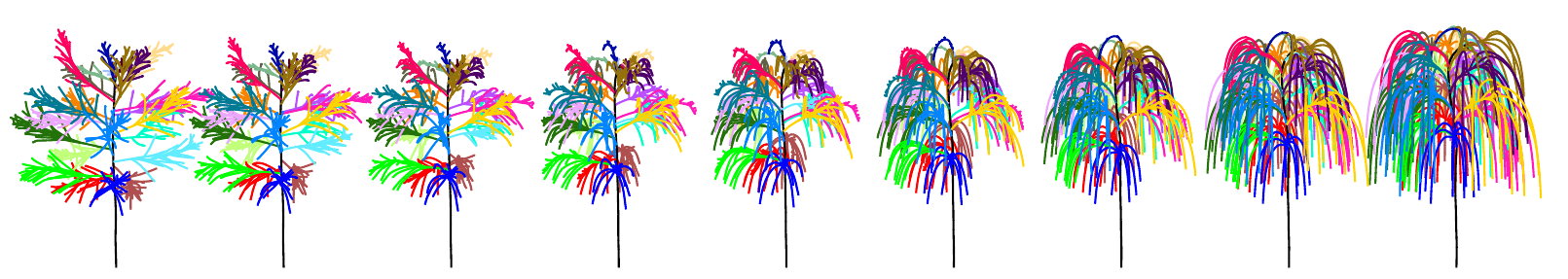}\\
\small{(c) }\\
\caption{Subtree correspondence map for Fig.~3 in the main manuscript. In each row, we show the geodesic between the most left and the most right trees as well as the color-coded subtree-wise correspondences. }
\label{fig:geodesic_corrMap_fig3}
\end{figure*}

\begin{figure*}[!ht]
\center
\includegraphics[width=\textwidth]{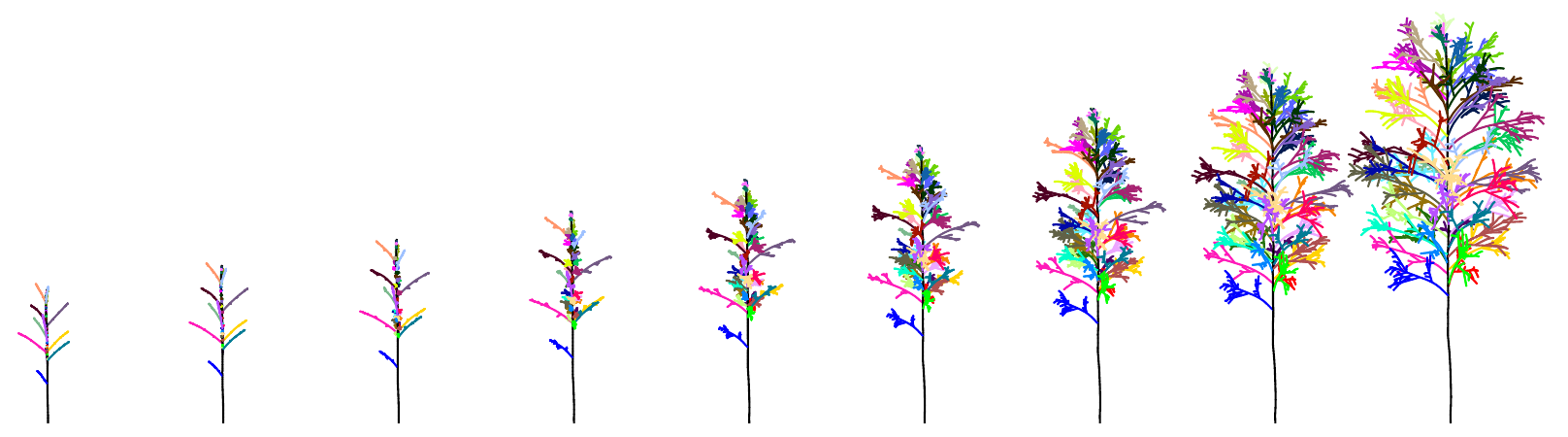}\\
\caption{Subtree correspondence map for Fig.~5 in the main manuscript. In each row, we show the geodesic between the most left and the most right trees as well as the color-coded subtree-wise correspondences. }
\label{fig:geodesic_corrMap_fig5}
\end{figure*}

\begin{figure*}[!ht]
\center
\includegraphics[width=\textwidth]{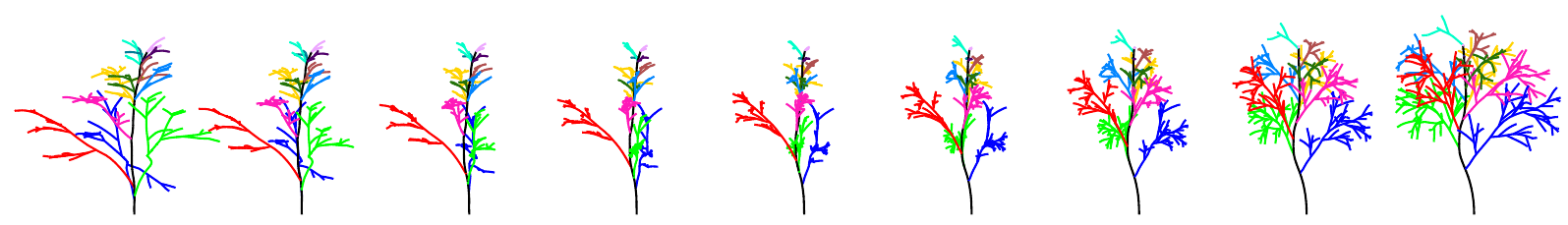}\\
\small{(a) }\\
\includegraphics[width=\textwidth]{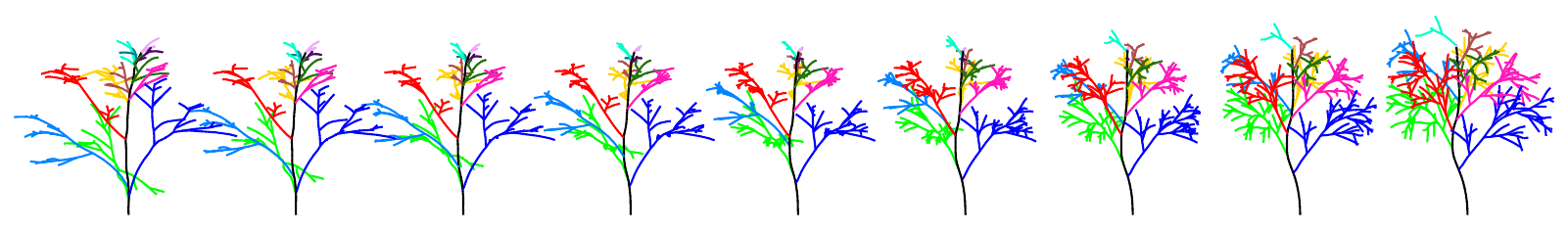}\\
\small{(b)}\\
\caption{Subtree correspondence map for Fig.~9 in the main manuscript. In each row, we show the geodesic between the most left and the most right trees as well as the color-coded subtree-wise correspondences. }
\label{fig:geodesic_corrMap_fig9}
\end{figure*}

\begin{figure*}[!ht]
\center
\includegraphics[width=\textwidth]{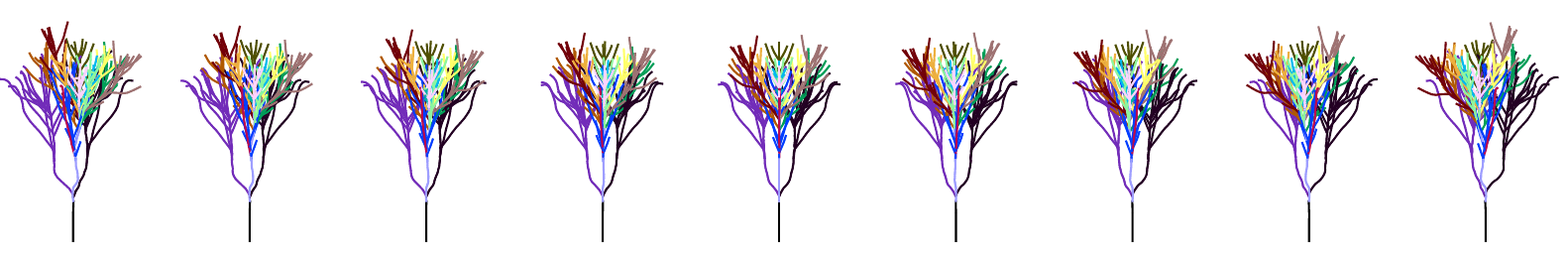}\\
\small{(a) }\\
\includegraphics[width=\textwidth]{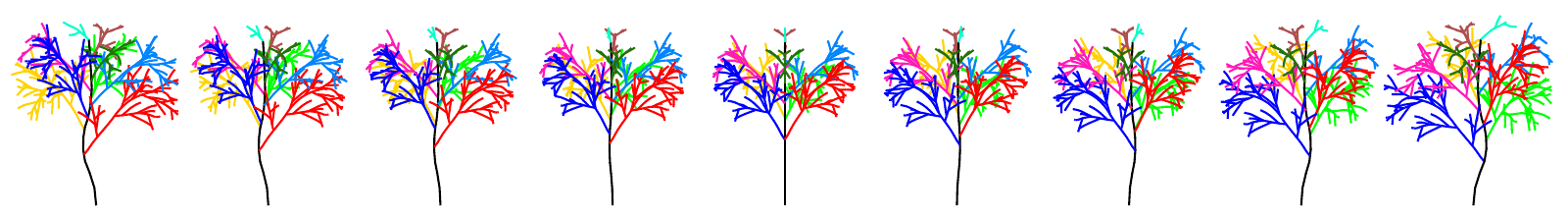}\\
\small{(b)}\\
\caption{Subtree correspondence map for Fig.~10 in the main manuscript. In each row, we show the geodesic between the most left and the most right trees as well as the color-coded subtree-wise correspondences. }
\label{fig:geodesic_corrMap_fig10}
\end{figure*}

Figs.~\ref{fig:geodesic_corrMap_fig3} to~\ref{fig:geodesic_corrMap_fig10} illustrate the computed correspondences in each of the geodesics reported in the main manuscript. For clarity, we only show subtree-wise correspondences using color maps. 

\end{document}